# Probabilistic Planning for Continuous Dynamic Systems under Bounded Risk

**Masahiro Ono**                                    ONO@APPI.KEIO.AC.JP
*Keio University*
*3-14-1 Hiyoshi, Kohoku-ku*
*Yokohama, Kanagawa, 223-8522 Japan*

**Brian C. Williams**                                    WILLIAMS@MIT.EDU
*Massachusetts Institute of Technology*
*77 Massachusetts Avenue*
*Cambridge, MA 02139 USA*

**Lars Blackmore**                                    LARS.BLACKMORE@SPACEX.COM
*SpaceX*
*1 Rocket Road*
*Hawthorne, CA 90250 USA*

## Abstract

This paper presents a model-based planner called the *Probabilistic Sulu Planner* or the *p-Sulu Planner*, which controls stochastic systems in a goal directed manner within user-specified risk bounds. The objective of the p-Sulu Planner is to allow users to command continuous, stochastic systems, such as unmanned aerial and space vehicles, in a manner that is both intuitive and safe. To this end, we first develop a new plan representation called a *chance-constrained qualitative state plan* (*CCQSP*), through which users can specify the desired evolution of the plant state as well as the acceptable level of risk. An example of a CCQSP statement is "go to A through B within 30 minutes, with less than 0.001% probability of failure." We then develop the p-Sulu Planner, which can tractably solve a CCQSP planning problem. In order to enable CCQSP planning, we develop the following two capabilities in this paper: 1) risk-sensitive planning with risk bounds, and 2) goal-directed planning in a continuous domain with temporal constraints. The first capability is to ensures that the probability of failure is bounded. The second capability is essential for the planner to solve problems with a continuous state space such as vehicle path planning. We demonstrate the capabilities of the p-Sulu Planner by simulations on two real-world scenarios: the path planning and scheduling of a personal aerial vehicle as well as the space rendezvous of an autonomous cargo spacecraft.

## 1. Introduction

There is an increasing need for risk-sensitive optimal planning in uncertain environments, while guaranteeing an acceptable probability of success. A motivating example for this article is the Boeing concept of a future aerial personal transportation system (PTS), as shown in Figure 1. The PTS consists of a fleet of small personal aerial vehicles (PAV) that enable the flexible point-to-point transportation of individuals and families.

In order to provide safety, PTS should be highly automated. In 2004, in the US, pilot error was listed as the primary cause of 75.5% of fatal general aviation accidents, according to the 2005 Joseph T. Nall Report (Aircraft Owners and Pilots Association Air Safety Foundation, 2005). Automated path planning, scheduling, collision avoidance, and traffic management will significantly improve the safety of PTS, as well as its efficiency. The challenges to operating such a system include adapting to uncertainties in the environment, such as storms and turbulence, while satisfying the complicated needs of users.

There is a substantial body of work on planning under uncertainty that is relevant. However, our approach is distinctive in three key respects. First, our planner, the p-Sulu Planner, allows users to explicitly limit the probability of constraint violation. This capability is particularly important for risk-sensitive missions where the impact of failure is significant. Second, the planner is goal-directed, by which we mean that it achieves time-evolved goals within user-specified temporal constraints. Third, the planner works in a continuous state space. A continuous state space representation fits naturally to many real-world applications, such as planning for aerial, space, and underwater vehicles. It is also important for problems with resources.

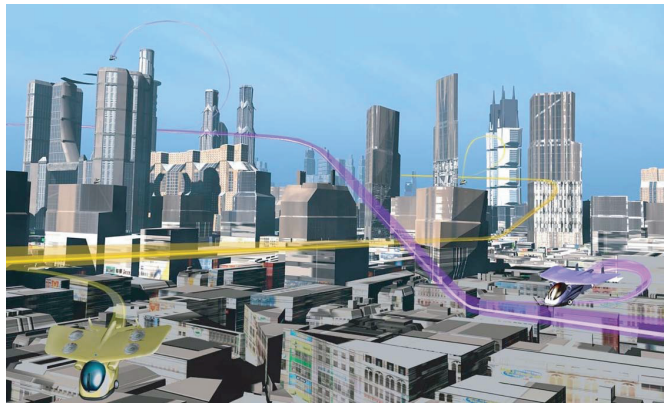

Figure 1: Personal Transportation System (PTS). (Courtesy of the Boeing Company)

Figure 2 shows a sample PTS scenario. A passenger of a PAV starts in Provincetown, MA and wants to go to Bedford within 30 minutes. The passenger also wants to go through a scenic area and remain there between 5 and 10 minutes during the flight. There is a no-fly zone (NFZ) and a storm that must be avoided. However, the storm's future location is uncertain; the vehicle's location is uncertain as well, due to control error and exogenous disturbances. Thus there is a risk of penetrating the NFZ or the storm. The passengers want to limit such risk to at most 0.001%.

In order to handle such a planning problem, we introduce a novel planner called the *Probabilistic Sulu Planner (p-Sulu Planner)*, building upon prior work on the model-based plan executive called Sulu (Léauté & Williams, 2005). The p-Sulu Planner provides the following three capabilities, in order to meet the needs described in the above scenario: 1) goal-directed planning in a continuous domain, 2) near-optimal planning, and 3) risk-sensitive planning with risk bounds.

- **Goal-directed planning in a continuous domain** The p-Sulu Planner must plan actions with continuous effects that achieve time evolved goals specified by users. In the case of the PTS scenario in Figure 2, the PAV must sequentially achieve two temporally extended goals, called episodes: going through the scenic area and then arriving at Bedford. There are additional temporal constraints on the goals that are inherent to the scenario; some temporal constraints come from physical limitations, such as fuel capacity, and others come from passenger requirements.

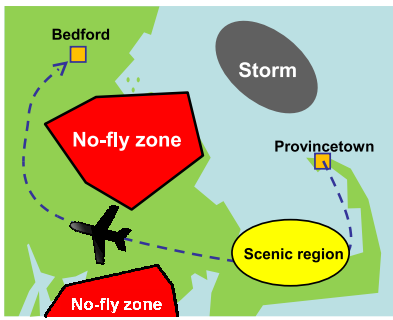

Figure 2: A sample plan for personal aerial vehicle (PAV)

- **Near-optimal stochastic planning** Cost reduction and performance improvement are important issues for any system. In the PTS scenario, passengers may want to minimize the trip time or fuel usage. The p-Sulu Planner finds a near-optimal control sequence according to the user-defined objective function, while satisfying given constraints.

- **Risk-sensitive planning with risk bounds** Real-world systems are subject to various uncertainties, such as state estimation error, modeling uncertainty, and exogenous disturbance. In the case of PAVs, the position and velocity of the vehicle estimated by the Kalman filter typically involve Gaussian-distributed uncertainties; the system model used for planning and control is not perfect; and the vehicles are subject to unpredictable disturbances such as turbulence. Under such uncertainty, the executed result of a plan inevitably deviates from the original plan and hence involves risk of constraint violation. Deterministic plan execution is particularly susceptible to risk when it is optimized in order to minimize a given cost function, since the optimal plan typically pushes against one or more constraint boundaries, and hence leaves no margin for error. For example, the shortest path in the PTS scenario shown in Figure 2 cuts in close to the NFZs and the storm, or more generally, constraint boundaries. Then, a tiny perturbation to the planned path may result in a penetration into the obstacles. Such risk can be reduced by setting a safety margin between the path and the obstacles, at a cost of longer path length. However, it is often impossible to guarantee zero risk, since there is typically a non-zero probability of having a disturbance that is large enough to push the vehicle out of the feasible region. Therefore, passengers of the vehicle must accept some risk, but at the same time they need to limit it to a certain level. More generally, users of an autonomous system under uncertainty should be able to specify their bounds on risk. The planner must guarantee that the system is able to operate within these bounds. Such constraints are called *chance constraints*.

## 1.1 Overview of the Planner

This section describes the inputs and outputs of the p-Sulu Planner informally. They are rigorously defined in Section 2.

### 1.1.1 INPUTS

**Initial Condition**   The p-Sulu Planner plans a control sequence starting from the current state, which is typically estimated from noisy sensor measurements. Therefore, the p-Sulu Planner takes the probability distribution, instead of the point estimate, of the current state as the initial condition.

**Stochastic Plant Model**   In the control community the planning problem is to generate a sequence of control inputs that actuate a physical system, called the plant. The action model for a plant is typically a system of real-valued equations over control, state and observable variables. The p-Sulu Planner takes as an input a linear stochastic plant model, which specifies probabilistic state transitions in a continuous domain. This is a stochastic extension of the continuous plant model used by Léauté and Williams (2005). In this paper we limit our focus to Gaussian-distributed uncertainty.

**Chance-constrained qualitative state plan (CCQSP)**   In order to provide users with an intuitive way to command stochastic systems, we develop a new plan representation called a *chance-constrained qualitative state plan (CCQSP)*. It is an extension of qualitative state plan (QSP), developed and used by Léauté and Williams (2005), Hofmann and Williams (2006), and Blackmore, Li, and Williams (2006). CCQSP specifies a desired evolution of the plant state over time, and is defined by a set of discrete *events*, a set of *episodes*, which impose constraints on the plant state evolution, a set of *temporal constraints* between events, and a set of *chance constraints* that specify reliability constraints on the success of sets of episodes in the plan.

A CCQSP may be depicted as a directed acyclic graph, as shown in Figure 3. The circles represent events and squares represent episodes. Flexible temporal constraints are represented as a simple temporal network (STN) (Dechter, Meiri, & Pearl, 1991), which specifies upper and lower bounds on the duration between two events (shown as the pairs of numbers in parentheses). The plan in Figure 3 describes the PTS scenario depicted in Figure 2, which can be stated informally as:

> "Start from Provincetown, reach the scenic region within 30 time units, and remain there for between 5 and 10 time units. Then end the flight in Bedford. The probability of failure of these episodes must be less than 1%. At all times, remain in the safe region by avoiding the no-fly zones and the storm. Limit the probability of penetrating such obstacles to 0.0001%. The entire flight must take at most 60 time units."

A formal definition of CCQSP is given in Section 2.4.3.

**Objective function**   The user of the p-Sulu Planner can specify an objective function (e.g., a cost function). We assume that it is a convex function.

### 1.1.2 OUTPUT

**Executable control sequence**   The p-Sulu Planner plans over a finite horizon. One of the two outputs of the p-Sulu Planner is an executable control sequence over the horizon that satisfies all constraints specified by the input CCQSP. In the case of the PTS scenario, the outputs are the vehicle's actuation inputs, such as acceleration and ladder angle, that result in the nominal paths shown in Figure 2. In order for the control sequence to be executable, it must be dynamically feasible. For example, the curvature of the PAV's path must not exceed the vehicles' maneuverability.

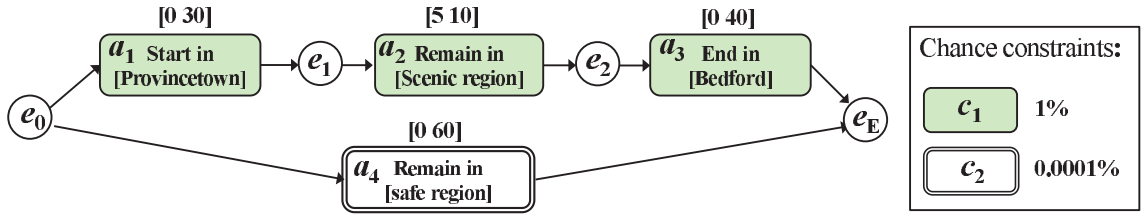

Figure 3: An example of a *chance-constrained qualitative state plan* (*CCQSP*), a new plan representation to specify the desired evolution of the plant state and the acceptable levels of risk. In the PTS scenario in Figure 2, the passengers of a PAV would like to go from Provincetown to Bedford, and fly over the scenic region on the way. The "safe region" means the entire state space except the obstacles. Risk of the episodes must be within the risk bounds specified by chance constraints.

**Optimal schedule**    The other output of the p-Sulu Planner is the optimal schedule, a set of execution time steps for events in the input CCQSP that minimizes a given cost function. In the case of the PTS scenario shown in Figure 3, a schedule specifies when to leave the scenic region and when to arrive at Bedford, for example. The p-Sulu Planner finds a schedule that satisfies all the simple temporal constraints specified by the CCQSP, and minimizes the cost function.

The two outputs – the control sequence and the schedule – must be consistent with each other: the time-evolved goals are achieved on the optimal schedule by applying the control sequence to the given initial conditions.

## 1.2 Approach

The p-Sulu Planner must solve a very difficult problem of generating an executable control sequence for a CCQSP, which involves both combinatorial optimization of a discrete schedule and non-convex optimization of a continuous control sequence. Our approach in this article is to develop the p-Sulu Planner in three technical steps, which we call "spirals".

In the first spiral, described in Section 4, we solve a special case of the CCQSP planning problem, where the feasible state space is convex (e.g., path planning problem *without* obstacles) and the schedule is fixed, as shown in Figure 4-(a). This problem can be transformed into a convex optimization problem by the *risk allocation approach*, which is presented in our previous work (Ono & Williams, 2008a). We obtain a feasible, near-optimal solution to the CCQSP planning problem by optimally solving the convex optimization using an interior point method (Blackmore & Ono, 2009).

In the second spiral, which is presented in Section 5, we consider a CCQSP problem with a *non-convex* state space in order to include obstacles, as in Figure 4-(b). We develop a branch and bound-based algorithm, called non-convex iterative risk allocation (NIRA). Subproblems of the branch-and-bound search of NIRA are convex chance-constrained optimal control problems, which are solved in the first spiral. The NIRA algorithm cannot handle a problem with a flexible schedule.

In the third spiral, which is described in Section 6, we develop another branch and bound-based algorithm, namely the p-Sulu Planner, which can solve a general CCQSP planning problem with a *flexible* schedule and obstacles. Subproblems of the branch-and-bound search of the p-Sulu Planner are non-convex chance-constrained optimal control problems, which are solved by the NIRA algorithm.

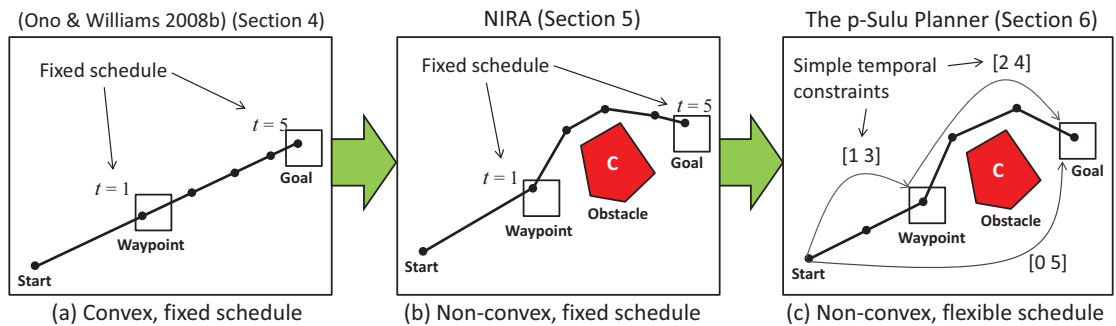

Figure 4: Three-spiral approach to the CCQSP planning problem

## 1.3 Related Work

Recall that the CCQSP planning problem is distinguished by its use of time-evolved goals, continuous states and actions, stochastic optimal solutions and chance constraints. While the planning and control disciplines have explored aspects of this problem, its solution in total is novel, and our approach to solving this problem efficiently through risk allocation is novel.

More specifically, there is an extensive literature on planning with discrete actions to achieve temporally extended goals (TEGs), such as TLPlan (Bacchus & Kabanza, 1998) and TALPlan (Kvarnstrom & Doherty, 2000), which treat TEGs as temporal domain control knowledge and prune the search space by progressing the temporal formula. However, since these TEG planners assume discrete state spaces, they cannot handle problems with continuous states and effects without discretization. Ignoring chance constraints, the representation of time evolved goals used by TLPlan and the p-Sulu Planner is similar. TLPlan uses a version of metric interval temporal logic (MITL) (Alur, Feder, & Henzinger, 1996) applied to discrete states, while the p-Sulu Planner uses qualitative state plans (QSPs) (Léauté & Williams, 2005; Hofmann & Williams, 2006; Li, 2010) over continuous states. Li (2010) shows that, for a given state space, any QSP can be expressed in MITL. The key difference that defines the p-Sulu Planner is the addition of chance constraints, together with its use of continuous variables.

Several planners, particularly those that are employed as components of model-based executives, command actions in continuous state space. For example, Sulu (Léauté & Williams, 2005) takes as input a deterministic linear model and QSP, which specifies a desired evolution of the plant state as well as flexible temporal constraints, and outputs a continuous control sequence. Chekhov (Hofmann & Williams, 2006) also takes as input a QSP and a nonlinear deterministic system model, and outputs a continuous control sequence. In order to enable fast real-time plan execution, Chekhov precomputes flow tubes, the sets of continuous state trajectories that end in the goal regions specified by the given plan. Kongming (Li, 2010) provides a generative planning capability for hybrid systems, involving both continuous and discrete actions. It employs a compact representation of hybrid plans, called a Hybrid Flow Graph, which combines the strengths of a Planning Graph for discrete actions and flow tubes for continuous actions. These planners adapt to the effects of uncertainty, but do not explicitly reason about the effects of uncertainty during planning. For example, Sulu employs a receding horizon approach, which continuously replans the control sequence using the latest measurements. Chekhov's flow tube representation of feasible policies allows the executive to generate new control sequences in response to disturbances on-line. The p-Sulu Planner is distinct from these continuous planners in that it plans with a model of uncertainty in dynamics, instead of just reacting to it. Its plan guarantees the user-specified probability of success by explicitly reasoning about the effects of uncertainty.

In AI planning literatures, a planning domain description language, PDDL+, supports mixed discrete-continuous planning domains (Fox & Long, 2006). Probabilistic PDDL (Younes & Littman, 2004) and the Relational Dynamic influence Diagram Language (RDDL) (Sanner, 2011) can handle stochastic systems. Recently, Coles, Coles, Fox, and Long (2012) developed a forward-chaining heuristic search planner named COLIN, which can deal with continuous linear change and duration-dependent effects. However, these planners do not handle chance constraints. We note that the outputs of the p-Sulu Planner is continuous in space but discrete in time. The time-dependent MDP developed by Boyan and Littman (2000) can handle continuous time by encoding time in the state. Extension of the p-Sulu Planner to continuous-time planning would be an interesting future direction.

Most work within the AI community on probabilistic planning has focused on planning in discrete domains and builds upon the Markov decision process (MDP) framework. A growing subcommunity has focused on extensions of MDPs to the continuous domain. However, tractability is an issue, since they typically require partitioning or approximation of continuous state space. A straightforward partitioning of continuous state and action spaces into discrete states and actions often leads to an exponential blow-up in running time. Furthermore, when the feasible state space is unbounded, it is impossible to partition the space into a finite number of compact subspaces. An alternative approach is the function approximation (Boyan & Moore, 1995), but its convergence is guaranteed only when the approximation error is bounded (Bertsekas & Tsitsiklis, 1996; Lagoudakis & Parr, 2003). Time-dependent MDPs (Boyan & Littman, 2000; Feng, Dearden, Meuleau, & Washington, 2004) can do efficient partitioning of continuous state space, but make an assumption that the set of available states and actions are finite (i.e., discrete). Hence, planning by these MDPs in a continuous state space, such as $\mathbb{R}^n$, requires to approximate the state space by a finite number of discrete states. Our approach is essentially different from the MDP approaches in that the continuous variables are directly optimized through convex optimization without discretization of continuous state space. Hence, the continuity of the state space does not harm the tractability of the p-Sulu Planner.

A second point of comparison is the treatment of risk. Like the p-Sulu Planner, the MDP framework offers an approach to marrying utility and risk. However, most MDP algorithms balance the utility and risk by assigning a large negative utility to the event of constraint violation. Such an approach cannot guarantee bounds on the probability of constraint violation. The constrained MDP approach (Altman, 1999) can explicitly impose constraints. Dolgov and Durfee (2005) showed that stationary deterministic policies for constrained MDPs can be obtained by solving a mixed integer linear program (MILP). However, the constrained MDP framework can only impose bounds on the *expected value* of costs, and again, cannot guarantee strict upper bounds on the probability of constraint violation. In contrast, the p-Sulu Planner allows users to impose chance constraints, which explicitly restrict the probability of constraint violation. As far as the authors know, the risk-sensitive reinforcement learning approach proposed by Geibel and Wysotzki (2005) is the only work that considers chance constraints in the MDP framework. They developed a reinforcement learning algorithm for MDPs with a constraint on the probability of entering error states. Our work is distinct from theirs in that the p-Sulu Planner is goal-directed, by which we mean that it achieves time-evolved goals within user-specified temporal constraints. To summarize, no prior MDP work supports continuous state and actions in combination with general continuous noise on transitions while ensuring that the probability of failure is bounded.

Risk-sensitive control methods in a continuous domain have been extensively studied in the discipline of control theory. For example, the celebrated $H_\infty$ control method minimizes the effect of disturbances on the output of a system while guaranteeing the stability of the system (Stoorvogel, 1992). Risk-sensitive control approaches allow users to choose the level of risk averseness through the minimization of an expected exponentiated cost function (Jacobson, 1973; Fleming & McEneaney, 1995). However, these approaches do not address chance constraints and optimal scheduling. Several methods have been proposed for solving stochastic optimal control problems over continuous variables with chance constraints. The method proposed by van Hessem (2004) turns a stochastic problem into a deterministic problem using a very conservative ellipsoidal relaxation. Blackmore (2006) proposes a sampling-based method called Particle Control, which evaluates joint chance constraints by a Monte-Carlo simulation, instead of using a conservative bound. As a result, the stochastic planning problem is reduced to a MILP problem. Although it has a theoretical guarantee that it can obtain the exactly optimal solution when an infinite number of samples are used, computation time is an issue. Blackmore et al. (2006) and Nemirovski and Shapiro (2006) employed Boole's inequality to decompose a joint chance constraint into individual chance constraints. Although Boole's inequality is less conservative than the ellipsoidal relaxation, their approach still has non-negligible conservatism since it fixes each individual risk bound to a uniform value. Our approach builds upon this approach, with modifications to allow flexible individual risk bounds.

To the best of the authors' knowledge, the p-Sulu Planner is the first goal-directed planner that is able to plan in a continuous state space with chance constraints.

## 1.4 Innovations

The p-Sulu Planner is enabled by six innovations presented in this article.

First, in order to allow users to command stochastic systems intuitively, we develop a new plan representation, CCQSP (Section 2.4.3).

Second, in order to decompose a chance constraint over a disjunctive clause into a disjunction of individual chance constraints, we introduce the risk selection approach (Section 5.1.2).

Third, in order to obtain lower bounds for the branch-and-bound search in NIRA, we develop the fixed risk relaxation (FRR), a linear program relaxation of the subproblems (Section 5.4.2).

Fourth, we minimize the search space for the optimal schedule by introducing a new forward checking method that efficiently prunes infeasible assignment of execution time steps (Section 6.2).

Fifth, in order to enhance the computation time of schedule optimization, we introduce a method to obtain a lower bound for the branch-and-bound by solving fixed-schedule planning problems with an partial assignment of a schedule. (Section 6.3)

Sixth, in order to minimize the number of non-convex subproblems solved in the branch-and-bound search, we introduce a variable ordering heuristic, namely the convex-episode-first (CEF) heuristic, which explores the episodes with a convex feasible state region before the ones with a non-convex state region (Section 6.2.2).

The rest of this article is organized as follows. Section 2 formally defines the CCQSP and states the CCQSP planning problem. Section 3 derives the encoding of the problem as a chance-constrained optimization problem, as well as the encodings of two limited versions of the CCQSP planning problem: one with a fixed schedule and a convex state space, and another with a fixed schedule and a non-convex state space. Section 4 reviews the solution to a fixed-schedule CCQSP planning problem with a convex state space. Section 5 develops the NIRA algorithm, which solves a fixed-schedule CCQSP planning problem with a *non-convex* state space, and Section 6 introduces the p-Sulu Planner, which solves a CCQSP planning problem with a *flexible schedule* and a non-convex state space. Finally, Section 7 shows simulation results on various scenarios, including the personal transportation system (PTS).

## 2. Problem Statement

Recall that the p-Sulu Planner takes as input a linear stochastic plant model, which specifies the effects of actions; an initial state description, describing a distribution over initial states; a CCQSP, which specifies desired evolutions of the state variables, as well as acceptable levels of risk; and an objective function. Its output is an executable control sequence and an optimal schedule. Planning is performed over a finite horizon, since the p-Sulu Planner is incorporated with the finite-horizon optimal control. We first define the variables used in the problem formulations. Then we define elements of the inputs and outputs.

### 2.1 Definition of Time Step

We consider a series of discretized finite time steps $t = 0, 1, 2, \cdots N$ with a fixed time interval $\Delta T$, where integer $N$ is the size of the planning horizon. Since the time interval $\Delta T$ can take any positive real value, it suffices to consider time steps with only integer indices to approximate the system's dynamics. We use the term "time step" to mean an integer index of the discretized time steps, while using the term "time" to mean a real-valued time. We define sets $\mathbb{T}$ and $\mathbb{T}^-$ as follows:

$$\mathbb{T} := \{0, 1, 2, \cdots N\}. \tag{1}$$

$$\mathbb{T}^- := \{0, 1, 2, \cdots N-1\}. \tag{2}$$

We limit the scope of this article to a discrete-time stochastic system. This is because optimizing a control sequence for a continuous-time stochastic system requires solving a stochastic differential equation (SDE) repeatedly. Performing such a computation is not tractable except for very simple problems.

### 2.2 Definitions of Events

An event denotes the start or end of an episode of behavior in our plan representation.

**Definition 1.** *An **event** $e \in \mathcal{E}$ is a instance that is executed at a certain time step in $\mathbb{T}$.*

We define two special events, the start event $e_0$ and the end event $e_E$. Without loss of generality, we assume that $e_0$ is executed at $t = 0$. The end event $e_E$ represents the termination of the entire plan.

## 2.3 Definitions of Variables

Variables used in our problem formulation involve a discrete schedule, a continuous state vector, and a continuous control vector.

We formally define an event as well as a schedule as follows:

**Definition 2.** *An **execution time step** $s(e) \in \mathbb{T}$ is an integer-valued scalar that represents the time step at which an event $e \in \mathcal{E}$ is executed. A **schedule** $s := [s(e_0), s(e_1), \cdots s(e_E)]$ is a sequence of execution time steps of all the events $e \in \mathcal{E}$. Finally, a **partial schedule** $\sigma := [\sigma(e) \in s \mid e \in \mathcal{E}_\sigma \subseteq \mathcal{E}]$ is an ordered set of execution time steps of a subset of events $\mathcal{E}_\sigma$.*

By definition, the start event is executed at $t = 0$ i.e, $s(e_0) = 0$. Following the notation of a schedule, we denote by $\sigma(e)$ the execution time of an event $e \in \mathcal{E}_\sigma$. See also the definition of a schedule (Definition 2).

We consider a continuous state space, where a state vector and a state sequence are defined as follows:

**Definition 3.** *A **state vector** $\boldsymbol{x}_t \in \mathbb{R}^{n_x}$ is a real-valued vector that represents the state of the plant at time step $t$. A **state sequence** $\boldsymbol{x}_{0:N} := [\boldsymbol{x}_0 \cdots \boldsymbol{x}_N]$ is a vector of state variables from time step $0$ to $N$.*

Our actions are assignments to continuous decision variables, which are referred to as a *control vector*:

**Definition 4.** *A **control vector** $\boldsymbol{u}_t \in \mathbb{R}^{n_u}$ is a real-valued vector that represents the control input to the system at time step $t$. A **control sequence** $\boldsymbol{u}_{0:N-1} := [\boldsymbol{u}_0 \cdots \boldsymbol{u}_{N-1}]$ is a vector of control inputs from time $0$ to $N - 1$.*

## 2.4 Definitions of Inputs

This subsection defines the four inputs of the p-Sulu Planner: an initial condition, a stochastic plant model, a CCQSP, and an objective function.

### 2.4.1 INITIAL CONDITION

The belief state at the beginning of the plan is represented by an initial state, which is assumed to have a Gaussian distribution with a known mean $\bar{\boldsymbol{x}}_0$ and a covariance matrix $\boldsymbol{\Sigma}_{\boldsymbol{x}_0}$:

$$\boldsymbol{x}_0 \sim \mathcal{N}(\bar{\boldsymbol{x}}_0, \boldsymbol{\Sigma}_{\boldsymbol{x}_0}). \tag{3}$$

The parameters in (3) are specified by an *initial condition*, which is defined as follows:

**Definition 5.** *An **initial condition** $I$ is a pair $I = \langle \bar{\boldsymbol{x}}_0, \boldsymbol{\Sigma}_{\boldsymbol{x}_0} \rangle$, where $\bar{\boldsymbol{x}}_0$ is the mean initial state and $\boldsymbol{\Sigma}_{\boldsymbol{x}_0}$ is the covariance matrix of the initial state.*

### 2.4.2 STOCHASTIC PLANT MODEL

The p-Sulu Planner controls dynamical systems in which actions correspond to the settings of continuous control variables, and whose effects are on continuous state variables. The p-Sulu Planner specifies these actions and their effects through a *plant model*. A plant model is considered as a state transition model in a continuous space. We employ a variant of a linear plant model with additive Gaussian uncertainty that is commonly used in the context of chance-constrained stochastic optimal control (Charnes & Cooper, 1959; Nemirovski & Shapiro, 2006; Oldewurtel, Jones, & Morari, 2008; van Hessem, 2004), with a modification to consider controller saturation. Specifically, we assume the following plant model:

$$\boldsymbol{x}_{t+1} = \boldsymbol{A}_t \boldsymbol{x}_t + \boldsymbol{B}_t \mu_{\mathbb{U}}(\boldsymbol{u}_t) + \boldsymbol{w}_t \tag{4}$$

where $\boldsymbol{w}_t \in \mathbb{R}^{n_x}$ is a state-independent disturbance at $t$-th time step that has a zero-mean Gaussian distribution with a *given* covariance matrix denoted by $\boldsymbol{\Sigma}_{w_t}$:

$$\boldsymbol{w}_t \sim \mathcal{N}(\boldsymbol{0}, \boldsymbol{\Sigma}_{w_t}). \tag{5}$$

Although this model prohibits state-dependent disturbance, most types of noise involved in our target applications are state independent. For example, in the PTS scenario introduced in Section 1, the primary source of uncertainty is a wind turbulence, which is typically not dependent on the state of a vehicle. In the space rendezvous scenario discussed in Section 7.5, the main sources of perturbations for a space craft are the tidal force and unmodeled gravitational effects of Sun, Moon, and other planets (Wertz & Wiley J. Larson, 1999). Such noises can be modeled as a state-dependent noise in practice when the scale of the planned actions is significantly smaller than that of the Solar System.

not dependent on the state of the space craft. We note that our problem formulation can encode time-varying noise by specifying different covariance matrices $\boldsymbol{\Sigma}_{w_t}$ for each time step.

The set $\mathbb{U} \subset \mathbb{R}^{n_u}$ is a compact convex set that represents the continuous domain of the feasible control inputs. If an infeasible control input $\boldsymbol{u}_t \notin \mathbb{U}$ is given to the plant, its actuators saturate. The function $\mu_{\mathbb{U}}(\cdot) : \mathbb{R}^{n_u} \mapsto \mathbb{U}$ in (4) represents the effect of actuator saturation as follows:

$$\mu_{\mathbb{U}}(\boldsymbol{u}) := \left\{ \begin{array}{ll} \boldsymbol{u} & (\text{if } \boldsymbol{u} \in \mathbb{U}) \\ P_{\mathbb{U}}(\boldsymbol{u}) & (\text{otherwise}) \end{array} \right. ,$$

where $P_{\mathbb{U}}(\boldsymbol{u})$ is a projection of $\boldsymbol{u}$ on $\mathbb{U}$. For example, when $\boldsymbol{u}$ is one-dimensional and $\mathbb{U} = [l, u]$, $P_{\mathbb{U}}(\boldsymbol{u}) = \max(\min(\boldsymbol{u}, u), l)$. Note that $\mu_{\mathbb{U}}$ introduces nonlinearity in the plant.

The parameters in (4) and (5) are specified by a *stochastic plant model*, which is defined as follows:

**Definition 6.** *A stochastic plant model $M$ is a four-tuple $M = \langle \boldsymbol{A}_{0:N-1}, \boldsymbol{B}_{0:N-1}, \boldsymbol{\Sigma}_{w_{0:N-1}}, \mathbb{U} \rangle$, where $\boldsymbol{A}_{0:N-1}$ and $\boldsymbol{B}_{0:N-1}$ are sets of $N$ matrices $\boldsymbol{A}_{0:N-1} := \{\boldsymbol{A}_0, \boldsymbol{A}_1, \cdots \boldsymbol{A}_{N-1}\}$, $\boldsymbol{B}_{0:N-1} := \{\boldsymbol{B}_0, \boldsymbol{B}_1, \cdots \boldsymbol{B}_{N-1}\}$, $\boldsymbol{\Sigma}_{w_{0:N-1}}$ is a set of $N$ covariance matrices $\boldsymbol{\Sigma}_{w_{0:N-1}} = \{\boldsymbol{\Sigma}_{w_0}, \boldsymbol{\Sigma}_{w_1}, \cdots, \boldsymbol{\Sigma}_{w_{N-1}}\}$, and $\mathbb{U} \subset \mathbb{R}^{n_u}$ is a compact convex set that represents the domain of the feasible control inputs.*

Note that $\boldsymbol{x}_t$, as well as $\boldsymbol{w}_t$, is a random variable, while $\boldsymbol{u}_t$ is a deterministic variable. Figure 5 illustrates our plant model. In a typical plant model, the probability circles grow over time since disturbance $\boldsymbol{w}_t$ is added at every time step, as drawn in the figure. This effect represents a commonly observed tendency that the distant future involves more uncertainty than the near future.

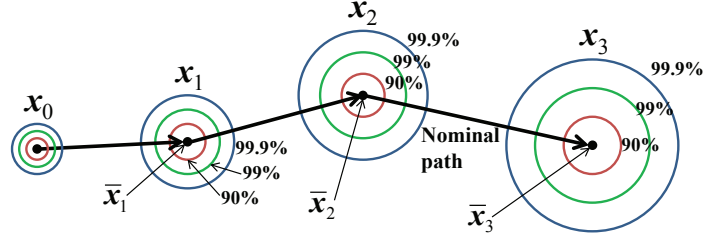

Figure 5: Illustration of the stochastic plant model used by the p-Sulu Planner.

In order to mitigate the accumulation of uncertainty, we employ a close-loop control approach, which generates the control input $\boldsymbol{u}_t$ by incorporating a nominal control input $\bar{\boldsymbol{u}}_t \in \mathbb{R}^{n_u}$ with an error feedback, as follows:

$$\boldsymbol{u}_t = \bar{\boldsymbol{u}}_t + \boldsymbol{K}_t(\boldsymbol{x}_t - \bar{\boldsymbol{x}}_t), \tag{6}$$

where $\boldsymbol{K}_t$ is a matrix representing a constant stabilizing feedback gain at time $t$ and $\bar{\boldsymbol{x}}_t$ is the nominal state vector. The nominal state $\bar{\boldsymbol{x}}_t$ is obtained by the following recursion:

$$\bar{\boldsymbol{x}}_0 := \boldsymbol{x}_0 \tag{7}$$

$$\bar{\boldsymbol{x}}_{t+1} = \boldsymbol{A}_t\bar{\boldsymbol{x}}_t + \boldsymbol{B}_t\bar{\boldsymbol{u}}_t. \tag{8}$$

A closed-loop control approach has been employed by Geibel and Wysotzki (2005) and Oldewurtel et al. (2008) in the context of chance-constrained optimal control and shown that it significantly improves performance.

In this closed-loop planning method, the *nominal* control input $\bar{\boldsymbol{u}}_t$ is planned before the execution. The actual control input $\boldsymbol{u}_t$ is computed in real time by using (6). The feedback term in (6) linearly responds to the error $\boldsymbol{x}_t - \bar{\boldsymbol{x}}_t$. By choosing the feedback gain $\boldsymbol{K}_t$ appropriately, the growth of the probability circles in Figure 5 can be slowed down. Neglecting the effect of controller saturation (i.e., assuming $\mathbb{U} = \mathbb{R}^{n_x}$), it follows from (4) and (6) that $\boldsymbol{x}_t$ has a Gaussian distribution with a covariance matrix $\boldsymbol{\Sigma}_{x_t}$, which evolves as follows:

$$\boldsymbol{\Sigma}_{x_{t+1}} = (\boldsymbol{A}_t + \boldsymbol{B}_t\boldsymbol{K}_t)\boldsymbol{\Sigma}_{x_t}(\boldsymbol{A}_t + \boldsymbol{B}_t\boldsymbol{K}_t)^T + \boldsymbol{\Sigma}_{w_t}. \tag{9}$$

In a typical plant, some of the eigenvalues of $A$ are one. Therefore, when there is no error feedback (i.e., $\boldsymbol{K}_t = 0$), the "size" of $\boldsymbol{\Sigma}_{x_t}$ grows by $\boldsymbol{\Sigma}_{w_t}$ at each iteration. By choosing $\boldsymbol{K}_t$ so that the norm of the largest eigenvalue of $(\boldsymbol{A}_t + \boldsymbol{B}_t\boldsymbol{K}_t)$ is less than one, the covariance $\boldsymbol{\Sigma}_{x_t}$ does not grow continuously. Such a feedback gain can be found by using standard control techniques, such as a linear quadratic regulator (LQR) (Bertsekas, 2005). Since we consider a finite-horizon, discrete-time planning problem, the optimal time-varying LQR gain $\boldsymbol{K}_t$ is obtained by solving the finite-horizon, discrete-time Riccati equation. In practice, it often suffices to use the steady-state (i.e., time-invariant) LQR gain, which is obtained by solving the infinite-horizon, discrete-time Riccati equation for simplicity. We note that the feedback gain $\boldsymbol{K}_t$ can also be optimized in real time. This approach is often used for robust and stochastic model predictive controls (Goulart, Kerrigan, & Maciejowski, 2006; Oldewurtel et al., 2008; Ono, 2012). However, such an extension is beyond the scope of this paper.

An issue is that, if the error $x_t - \bar{x}_t$ happens to be very large, the control input $u_t$ may exceed its feasible domain $\mathbb{U}$, resulting in actuator saturation. Therefore, (9) does not hold due to the nonlinearity of the function $\mu_{\mathbb{U}}(\cdot)$. We address this issue through the risk allocation approach. More specifically, we impose chance constraints on control saturation, and allocate risk to both state and control constraints. This approach is discussed more in detail in Section 4.1.5.

### 2.4.3 CHANCE-CONSTRAINED QUALITATIVE STATE PLAN (CCQSP)

A qualitative state plan (QSP) (Léauté & Williams, 2005) is a temporally flexible plan that specifies the desired evolution of the plant state. The activities of a QSP are called episodes and specify constraints on the plant state. CCQSP is an extension of QSPs to stochastic plans that involve chance constraints, defined as follows:

**Definition 7.** *A **chance-constrained qualitative state plan** (CCQSP) is a four-tuple $P = \langle \mathcal{E}, \mathcal{A}, \mathcal{T}, \mathcal{C} \rangle$, where $\mathcal{E}$ is a set of discrete events, $\mathcal{A}$ is a set of episodes, $\mathcal{T}$ is a set of simple temporal constraints, and $\mathcal{C}$ is a set of chance constraints.*

The four elements of a CCQSP are defined precisely in a moment. Like a QSP, a CCQSP can be illustrated diagrammatically by a directed acyclic graph in which the discrete events in $\mathcal{E}$ are represented by vertices, drawn as circles, and the episodes as arcs with ovals. A CCQSP has a start event $e_0$ and an end $e_E$, which corresponds to the beginning and the end of the mission, respectively.

For example, Figure 3 shows the CCQSP of the PTS scenario. The state regions and obstacles in the CCQSP are illustrated in Figure 2. It involves four events: $\mathcal{E} = \{e_0, e_1, e_2, e_E\}$. Their meanings are described as follows.

1. The start event $e_0$ corresponds to the take off of the PAV from Provincetown.

2. The second event $e_1$ corresponds to the time step when PAV reaches the scenic region.

3. Event $e_2$ is associated with the time instant when the PAV has just left the scenic region.

4. The end event $e_E$ corresponds to the arrival of the PAV in Bedford.

The CCQSP has four episodes $\mathcal{A} = \{a_1, a_2, a_3, a_4\}$ and two chance constraints $\mathcal{C} = \{c_1, c_2\}$.

A natural language expression of the CCQSP is:

> " Start from Provincetown, reach the scenic region within 30 time units, and remain there for between 5 and 10 time units. Then end the flight in Bedford. The probability of failure of these activities must be less than 1%. At all times, remain in the safe region by avoiding the no-fly zones and the storm. Limit the probability of penetrating such obstacles to 0.0001%. The entire flight must take at most 60 time units."

Below we formally define the three types of constraints - episodes, temporal constraints, and chance constraint.

**Episodes**  Each *episode* $a \in \mathcal{A}$ specifies the desired state of the system under control over a time interval.

**Definition 8.** *An **episode** $a = \langle e_a^S, e_a^E, \Pi_a(t_S, t_E), R_a \rangle$ has an associated start event $e_a^S$ and an end event $e_a^E$. $R_a \in \mathbb{R}^N$ is a region in a state space. $\Pi_a \subseteq \mathbb{T}$ is a set of time steps at which the state $x_t$ must be in the region $R_a$.*

The feasible region $R_a$ can be any subset of $\mathbb{R}^N$. We will approximate $R_a$ with a set of linear constraints later in Section 3.1.1.

$\Pi_a(t_S, t_E)$ is a subset of $\mathbb{T}$ given as a function of the episode's start time step $t_S = s(e_a^S)$ and its end time step $t_E = s(e_a^E)$. Different forms of $\Pi_a(t_S, t_E)$ result in various types of episodes. The following three types of episodes are particularly of interest to us:

1. *Start-in episode*: $\Pi_a(t_S, t_E) = \{t_S\}$

2. *End-in episode*: $\Pi_a(t_S, t_E) = \{t_E\}$

3. *Remain-in episode*: $\Pi_a(t_S, t_E) = \{t_S, t_S + 1, \cdots, t_E\}$

For a given episode $a$, the set of time steps at which the plant state must be in the region $R_a$ is obtained by substituting $s(e_a^S)$ and $s(e_a^E)$, the execution time steps of the start event and the end event of the episode, into $t_S$ and $t_E$. In other words, an episode $a$ requires that the plant state is in $R_a$ for all time steps in $\Pi_a\left(s(e_a^S), s(e_a^E)\right)$. For the rest of the article, we use the following abbreviated notation:

$$\Pi_a(s) := \Pi_a\left(s(e_a^S), s(e_a^E)\right).$$

Using this notation, an episode is equivalent to the following state constraint:

$$\bigwedge_{t \in \Pi_a(s)} \boldsymbol{x}_t \in R_a. \tag{10}$$

For example, in the CCQSP shown in Figure 3, there are four episodes: $a_1$ ("Start in [Province-town]"), $a_2$ ("Remain in [Scenic region]"), $a_3$ ("End in Bedford"), and $a_4$ ("Remain in [safe region]").

In Section 6, we solve a relaxed optimization problem with a partial schedule (Definition 2) in order to obtain a lower bound on the optimal objective value. In such relaxed problems, only a subset of the episodes that are relevant to the given partial schedule are imposed. We formally define *a partial episode set* of a partial schedule $\sigma$ as follows:

**Definition 9.** *Given a partial schedule $\sigma$, $\mathcal{A}(\sigma) \subseteq \mathcal{A}$ is its **partial episode set**, which is a subset of $\mathcal{A}$ that involves the episodes whose start event and end event are assigned execution time steps.*

$$\mathcal{A}(\sigma) = \left\{ a \in \mathcal{A} \mid e_a^S \in \mathcal{E}_\sigma \wedge e_a^E \in \mathcal{E}_\sigma \right\},$$

*where the definition of $\mathcal{E}_\sigma$ is given in Definition 2.*

**Chance constraint** Recall that a chance constraint is a probabilistic constraint that requires the constraints defining each episode to be satisfied within a user-specified probability. A CCQSP can have multiple chance constraints. A chance constraint is associated with at least one episode.

A chance constraint is formally defined as follows:

**Definition 10.** *A **chance constraint** $c = \langle \Psi_c, \Delta_c \rangle$ is a constraint requiring that:*

$$\Pr\left[\bigwedge_{a \in \Psi_c} \bigwedge_{t \in \Pi_a(s)} \boldsymbol{x}_t \in R_a\right] \geq 1 - \Delta_c, \tag{11}$$

*where $\Delta_c$ is a user-specified risk bound and $\Psi_c \subseteq \mathcal{A}$ is a set of episodes associated with the chance constraint $c$.*

Note that every episode in a CCQSP must be associated with exactly one chance constraint. Any episode in $\mathcal{A}$ must not be involved in more than one chance constraint or unassociated with any chance constraint.

For example, the CCQSP shown in Figure 3 has two chance constraints, $c_1$ and $c_2$. Their associated episodes are $\Psi_{c_1} = \{a_1, a_2, a_3\}$ and $\Psi_{c_2} = \{a_4\}$. Therefore, $c_1$ requires that the probability of satisfying the three episodes $a_1$, $a_2$, and $a_3$ (colored in green) is more than than 99%, while $c_2$ requires that the probability of satisfying the episode $a_4$ is more than 99.99999%.

We make the following assumption, which is necessary in order to guarantee the convexity of constraints in Section 4.2.

**Assumption 1.**

$$\Delta_c \leq 0.5$$

This assumption requires that the risk bounds are less than 50%. We claim that this assumption does not constrain practical applications, since typically the user of an autonomous system would not accept more than 50% risk.

**Temporal constraint** A CCQSP includes *simple temporal constraints (STCs)* (Dechter et al., 1991), which impose upper and lower bounds on the duration of episodes and on the temporal distances between two events in $\mathcal{E}$.

**Definition 11.** *A **simple temporal constraint** $\tau = \langle e_\tau^S, e_\tau^E, b_\tau^{\min}, b_\tau^{\max} \rangle$ is a constraint, specifying that the duration from a start event $e_\tau^S$ to an end event $e_\tau^E$ be in the real-valued interval $[b_\tau^{\min}, b_\tau^{\max}] \subseteq [0, +\infty]$.*

Temporal constraints are represented diagrammatically by arcs between nodes, labeled with the time bounds $[b_\tau^{\min}, b_\tau^{\max}]$, or by labels over episodes. For example, the CCQSP shown in Figure 3 has four simple temporal constraints. One requires the time between $e_0$ and $e_1$ to be at most 30 time units. One requires the time between $e_1$ and $e_2$ to be at least 5 units and at most 10 units. One requires the time between $e_2$ and $e_E$ to be at most 40 time units. One requires the time between $e_0$ and $e_E$ to be at most 60 time units.

A schedule $s$ is feasible if it satisfies all temporal constraints in the CCQSP. The number of feasible schedules is finite, since $\mathbb{T}$ is discrete and finite. We denote by $\mathcal{S}_F$ the domain of feasible schedules, which is formally defined as follows:

$$\mathcal{S}_F = \{s \in \mathbb{T}^{|\mathcal{E}|} \mid \forall_{\tau \in \mathcal{T}} \quad b_\tau^{\min} \leq \Delta T \{s(e_\tau^E) - s(e_\tau^S)\} \leq b_\tau^{\max}\}, \tag{12}$$

where $|\mathcal{E}|$ is the number of events in the CCQSP. The temporal duration is multiplied by the time interval $\Delta T$ because $b_\tau^{\min}$ and $b_\tau^{\min}$ are real-valued time, while $s$ is a set of discrete time steps in $\mathbb{T}$.

### 2.4.4 OBJECTIVE FUNCTION

In this section, we formally define the objective function.

**Definition 12.** *An **objective function** $J : \mathbb{U}^N \times \mathbb{X}^N \times \mathcal{S}_F \mapsto \mathbb{R}$ is a real-valued function over the nominal control sequence $\bar{\boldsymbol{u}}_{0:N-1}$, the nominal state sequence $\bar{\boldsymbol{x}}_{1:N}$, and the schedule $s$. We assume that $J$ is a convex function over $\bar{\boldsymbol{x}}_{1:N}$ and $\bar{\boldsymbol{u}}_{0:N-1}$.*

A typical example of an objective function is the quadratic sum of control inputs, which requires the total control efforts to be minimized:

$$J(\bar{\boldsymbol{u}}_{0:N-1}, \bar{\boldsymbol{x}}_{1:N}, s) = \sum_{t=0}^{N-1} ||\bar{\boldsymbol{u}}_t||^2.$$

Another example is:

$$J(\bar{\boldsymbol{u}}_{0:N-1}, \bar{\boldsymbol{x}}_{1:N}, s) = s(e_E), \tag{13}$$

which minimizes the total plan execution time, by requiring that the end event $e_E$ of the qualitative state plan be scheduled as soon as possible.

There is often a need to minimize the expectation of a cost function. Note that, in our case, the expectation of a function over $\boldsymbol{x}_{1:N}$ and $\boldsymbol{u}_{0:N-1}$ can be reduced to a function over $\bar{\boldsymbol{u}}_{0:N-1}$ because it follows from (4)-(6) that the probability distributions of $\boldsymbol{x}_{1:N}$ and $\boldsymbol{u}_{0:N-1}$ are uniquely determined by $\bar{\boldsymbol{u}}_{0:N-1}$ and $\boldsymbol{K}_t$. In practice, it is often more convenient to express the objective function as a function of $\bar{\boldsymbol{u}}_{0:N-1}$ *and* $\bar{\boldsymbol{x}}_{1:N}$, rather than as a function of $\bar{\boldsymbol{u}}_{0:N-1}$. Since $\bar{\boldsymbol{x}}_{1:N}$ are specified by $\bar{\boldsymbol{u}}_{0:N-1}$ using (8), the two expressions are equivalent. The conversion from the expectation of a cost function to a function over nominal values can be conducted a priori.

If there is no controller saturation, such a conversion can often be obtained in a closed form. The conversion is particularly straight forward when the cost function is polynomial, since the expectation is equivalent to a combination of raw moments, which can be readily derived from the cumulants. Note that the third and higher cumulants of the Gaussian distribution are zero. Below we show examples of the conversion regarding three commonly-used cost functions: linear, quadratic, and the Manhattan norm.

$$\mathbb{E}[\boldsymbol{x}_t] = \bar{\boldsymbol{x}}_t \tag{14}$$

$$\mathbb{E}[\boldsymbol{x}_t^T \boldsymbol{Q} \boldsymbol{x}_t] = \bar{\boldsymbol{x}}_t^T \boldsymbol{Q} \bar{\boldsymbol{x}}_t + tr(\boldsymbol{Q}\boldsymbol{\Sigma}_{x_t}) \tag{15}$$

$$\mathbb{E}[||\boldsymbol{x}_t||_1] = \sum_{i=1}^{n_x} \sigma_{x_t,i} \sqrt{\frac{2}{\pi}} {}_1F_1\left(-\frac{1}{2}, \frac{1}{2}, -\frac{\bar{\boldsymbol{x}}_{t,i}^2}{2\sigma_{x_t,i}^2}\right), \tag{16}$$

where $\boldsymbol{Q}$ is a positive definite matrix, $\sigma_{x_t,i}$ is the $i$th diagonal element of $\boldsymbol{\Sigma}_{x_t}$, and ${}_1F_1(\cdot)$ is a confluent hypergeometric function. All functions above are convex. The expectation of a function of $\boldsymbol{u}_t$ can also be transformed to a function of $\bar{\boldsymbol{u}}_t$ in the same manner. Note that the second term on the right hand side of (15) is a constant. Hence, minimizing $\bar{\boldsymbol{x}}_t^T \boldsymbol{Q} \bar{\boldsymbol{x}}_t$ yields the same solution as minimizing $\mathbb{E}[\boldsymbol{x}_t^T \boldsymbol{Q} \boldsymbol{x}_t]$.

When there is controller saturation, it is difficult to obtain the conversion in a closed-form due to the nonlinearity of $\mu_{\mathbb{U}}(\cdot)$ in (4). In practice, we use an approximation that assumes no saturation. Since our closed-loop control approach explicitly limits the probability of controller saturation to a small probability (see Section 4.1.5 for the detail), the approximation error is trivial. This claim is empirically validated in Section 7.2.4.

## 2.5 Definitions of Outputs

The output of the p-Sulu Planner is an *optimal solution*, which consists of an optimal control sequence $\boldsymbol{u}_{0:N-1}^\star \in \mathbb{U}^N$ and an optimal schedule $s^\star \in \mathcal{S}_F$.

**Definition 13.** *The optimal solution is a pair $\langle \boldsymbol{u}_{0:N-1}^{\star}, s^{\star} \rangle$. The solution satisfies all constraints in the given CCQSP (Definition 7), the initial condition $I$, and the stochastic plant model $M$. The solution minimizes the given objective function $J(\boldsymbol{u}_{0:N-1}, \bar{\boldsymbol{x}}_{1:N}, s)$ (Definition 12).*

## 2.6 Problem Statement

We now formally define the *CCQSP planning* problem.

### Problem 1: CCQSP Planning Problem

Given a stochastic plant model $\mathcal{M} = \langle \boldsymbol{A}_{0:N-1}, \boldsymbol{B}_{0:N-1}, \boldsymbol{\Sigma}_{w_{0:N-1}} \rangle\rangle$, an initial condition $\mathcal{I} = \langle \bar{\boldsymbol{x}}_0, \boldsymbol{\Sigma}_{\boldsymbol{x}_0} \rangle$, a CCQSP $P = \langle \mathcal{E}, \mathcal{A}, \mathcal{T}, \mathcal{C} \rangle$, and an objective function $J(\boldsymbol{u}_{0:N-1}, \bar{\boldsymbol{x}}_{1:N})$, a CCQSP planning problem is to find an optimal solution $\langle \boldsymbol{u}_{0:N-1}^{\star}, s^{\star} \rangle$ for $\mathcal{M}, \mathcal{I}, P$, and $J$.

We note that the p-Sulu Planner gives a *near-optimal* solution to Problem 1. The p-Sulu Planner employs two approximations, namely risk allocation (Section 4.1.1) and risk selection (Section 5.1.1), for the sake of computational tractability. As a result, its solution is not strictly optimal in general. However, we empirically show in Section 7 that the suboptimality due to risk allocation and risk selection is significantly smaller than existing approximation methods.

# 3. Problem Encoding

This section encodes the CCQSP planning problem stated in the previous section into a mathematical programming problem. Sections 4 - 6 then address how to solve this form of mathematical problem. Recall that we build our CCQSP planner, the p-Sulu Planner, in three spirals. We first present the problem encoding of a general CCQSP planning problem with a non-convex state space and a flexible schedule (Figure 4-(c)) in Subsection 3.1. Then we present the encodings of the two special cases of the CCQSP planning problem in Subsections 3.2 and 3.3: one with a non-convex state space and a *fixed* schedule (Figure 4-(b)), and one with a *convex* state space and a fixed schedule (Figure 4-(a)).

## 3.1 Encoding of a CCQSP Planning Problem with a Non-convex State Space and Flexible Schedule

### 3.1.1 Encoding of Feasible Regions

In order to encode Problem 1 into a mathematical programming problem, the geometric constraint in (11), $\boldsymbol{x}_t \in R_a$, must be represented by algebraic constraints. For that purpose, we approximate the feasible state regions $R_a$ by a set of half-spaces, each of which is represented by a linear state constraint.

Figure 6 shows two simple examples. The feasible region of (a) is outside of the obstacle, which is approximated by a triangle. The feasible region of (b) is inside of the pickup region, which is again approximated by a triangle. Each feasible region is approximated by a set of linear constraints as follows:

$$(a) \quad \bigvee_{i=1}^{3} \boldsymbol{h}_i^T \boldsymbol{x} \leq \boldsymbol{g}_i, \qquad (b) \quad \bigwedge_{i=1}^{3} \boldsymbol{h}_i^T \boldsymbol{x} \geq \boldsymbol{g}_i.$$

We approximate the feasible regions so that the set of linear constraints is a sufficient condition of the original state constraint $\boldsymbol{x}_t \in R_a$.

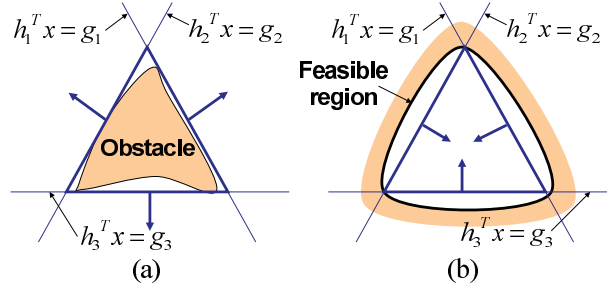

Figure 6: Approximate representation of feasible regions by a set of linear constraints

We assume that the set of linear state constraints that approximates a feasible region has been reduced to conjunctive normal form (CNF) as follows:

$$\bigwedge_{k \in \mathcal{K}_a} \bigvee_{j \in \mathcal{J}_{a,k}} \boldsymbol{h}_{a,k,j}^T \boldsymbol{x}_t - g_{a,k,j} \leq 0, \tag{17}$$

where $\mathcal{K}_a = \{1, 2, \cdots |\mathcal{K}_a|\}$ and $\mathcal{J}_{c,i} = \{1, 2, \cdots |\mathcal{J}_{c,i}|\}$ are sets of indices. By replacing $\boldsymbol{x}_t \in R_a$ in (11) by (17), a chance constraint $c$ is encoded as follows:

$$\Pr\left[\bigwedge_{a \in \Psi_c} \bigwedge_{t \in \Pi_a(s)} \bigwedge_{k \in \mathcal{K}_a} \bigvee_{j \in \mathcal{J}_{a,k}} \boldsymbol{h}_{c,a,k,j}^T \boldsymbol{x}_t - g_{c,a,k,j} \leq 0\right] \geq 1 - \Delta_c. \tag{18}$$

In order to simplify the notation, we merge indices $a \in \Psi_c$, $t \in \Pi_a(s)$, and $k \in \mathcal{K}_a$ into a new index $i \in \mathcal{I}_c(s)$, where $\mathcal{I}_c(s) = \{1, 2, \cdots |\mathcal{I}_c(s)|\}$ and $|\mathcal{I}_c(s)| = |\mathcal{K}_a| \cdot \sum_{a \in \Psi_c} |\Pi_a(s)|$. We let $a_i$, $k_i$, and $t_i$ the indices that correspond to to the combined index $i$, and let $h_{c,i,j} = h_{c,a_i,k_i,j}$. Using these notations, the three conjunctions of (18) are combined into one, and we obtain the following encoding of a chance constraint:

$$\Pr\left[\bigwedge_{i \in \mathcal{I}_c(s)} \bigvee_{j \in \mathcal{J}_{c,i}} \boldsymbol{h}_{c,i,j}^T \boldsymbol{x}_{t_i} - g_{c,i,j} \leq 0\right] \geq 1 - \Delta_c. \tag{19}$$

The specification of chance constraints given in (19) requires that all $|\mathcal{I}_c(s)|$ disjunctive clauses of state constraints must be satisfied with a probability $1 - \Delta_c$. The $i$'th disjunctive clause of the $c$'th chance constraint is composed of $|\mathcal{J}_{c,i}|$ linear state constraints.

### 3.1.2 CCQSP PLANNING PROBLEM ENCODING

Using (3), (4), (5), (6), and (19), a CCQSP planning problem (Problem 1), which is solved in the third spiral, is encoded as follows:

**Problem 2: General CCQSP Planning Problem**

$$\min_{\bar{\boldsymbol{u}}_{0:N-1}, s} \quad J(\boldsymbol{u}_{0:N-1}, \bar{\boldsymbol{x}}_{1:N}, s) \tag{20}$$

$$\text{s.t.} \quad s \in \mathcal{S}_F \tag{21}$$

$$\boldsymbol{x}_{t+1} = \boldsymbol{A}_t \boldsymbol{x}_t + \boldsymbol{B}_t \mu_{\mathbb{U}}(\boldsymbol{u}_t) + \boldsymbol{w}_t, \quad \forall t \in \mathbb{T}^- \tag{22}$$

$$\boldsymbol{u}_t = \bar{\boldsymbol{u}}_t + \boldsymbol{K}_t(\boldsymbol{x}_t - \bar{\boldsymbol{x}}_t), \quad \forall t \in \mathbb{T}^- \tag{23}$$

$$\bigwedge_{c \in \mathcal{C}} \Pr \left[ \bigwedge_{i \in \mathcal{I}_c(s)} \bigvee_{j \in \mathcal{J}_{c,i}} \boldsymbol{h}_{c,i,j}^T \boldsymbol{x}_{t_i} - g_{c,i,j} \leq 0 \right] \geq 1 - \Delta_c. \tag{24}$$

$$\boldsymbol{x}_0 \sim \mathcal{N}(\bar{\boldsymbol{x}}_0, \boldsymbol{\Sigma}_{\boldsymbol{x}_0}), \quad \boldsymbol{w}_t \sim \mathcal{N}(\boldsymbol{0}, \boldsymbol{\Sigma}_{\boldsymbol{w}_t}), \quad \forall t \in \mathbb{T}^- \tag{25}$$

Recall that $\mathcal{S}_F$, formally defined in (12), is the set of schedules that satisfy all temporal constraints in the given CCQSP. This CCQSP execution problem is a hybrid optimization problem over both discrete variables $s$ (schedule) and continuous variables $\boldsymbol{u}_{0:N-1}$ (control sequence). Note that the temporal constraints within Problem 2 are solved in Section 6. A similar problem encoding is also employed in the chance-constraint MDP proposed by Geibel and Wysotzki (2005). However, our encoding differs from Geibel and Wysotzki in two respects: 1) we optimize not only the continuous control sequence $\boldsymbol{u}_{0:N-1}$ but also the discrete schedule $s$ with temporal constraints; 2) we allow joint chance constraints, which require the satisfaction of *multiple* state constraints for a given probability. Problem 2 is solved in Section 6.

## 3.2 Encoding of a CCQSP Planning Problem with a Non-convex State Space and Fixed Schedule

A restricted version of a CCQSP planning problem with a *fixed* schedule, which is solved in the second spiral, is obtained by fixing $s$ in Problem 2 as follows:

**Problem 3: CCQSP Planning Problem with a Fixed Schedule**

$$J^\star(s) = \min_{\bar{\boldsymbol{u}}_{0:N-1}} \quad J'(\boldsymbol{u}_{0:N-1}, \bar{\boldsymbol{x}}_{1:N}) \tag{26}$$

$$\text{s.t.} \quad \boldsymbol{x}_{t+1} = \boldsymbol{A}_t \boldsymbol{x}_t + \boldsymbol{B}_t \mu_{\mathbb{U}}(\boldsymbol{u}_t) + \boldsymbol{w}_t, \quad \forall t \in \mathbb{T}^- \tag{27}$$

$$\boldsymbol{u}_t = \bar{\boldsymbol{u}}_t + \boldsymbol{K}_t(\boldsymbol{x}_t - \bar{\boldsymbol{x}}_t), \quad \forall t \in \mathbb{T}^- \tag{28}$$

$$\bigwedge_{c \in \mathcal{C}} \Pr \left[ \bigwedge_{i \in \mathcal{I}_c(s)} \bigvee_{j \in \mathcal{J}_{c,i}} \boldsymbol{h}_{c,i,j}^T \boldsymbol{x}_{t_i} - g_{c,i,j} \leq 0 \right] \geq 1 - \Delta_c, \tag{29}$$

$$\boldsymbol{x}_0 \sim \mathcal{N}(\bar{\boldsymbol{x}}_0, \boldsymbol{\Sigma}_{\boldsymbol{x}_0}), \quad \boldsymbol{w}_t \sim \mathcal{N}(\boldsymbol{0}, \boldsymbol{\Sigma}_{\boldsymbol{w}_t}), \quad \forall t \in \mathbb{T}^- \tag{30}$$

where $J^\star(s)$ is the optimal objective value of the CCQSP Planning problem with the schedule fixed to $s$. Note that the schedule $s$, which is a decision variable in Problem 2, is treated as a constant in Problem 3. Therefore, the objective function $J'$ is now a function of only control sequence and mean state, since we have fixed the schedule. Since we assumed that $J$ is a convex function regarding to $\boldsymbol{u}_{0:N-1}$ and $\bar{\boldsymbol{x}}_{1:N}$, $J'$ is also a convex function. Section 5 solves Problem 3.

### 3.3 Encoding of a CCQSP Planning Problem with a Convex State Space and Fixed Schedule

A more restrictive version of a CCQSP planning problem with a fixed schedule and a convex state space, which is solved in the first spiral, is obtained by removing the disjunctions in the chance constraints in Problem 3 as follows:

**Problem 4: CCQSP Planning Problem with a Fixed Schedule and a Convex State Space**

$$\min_{\bar{\boldsymbol{u}}_{0:N-1}} \quad J'(\boldsymbol{u}_{0:N-1}, \bar{\boldsymbol{x}}_{1:N}) \tag{31}$$

$$\boldsymbol{x}_{t+1} = \boldsymbol{A}_t \boldsymbol{x}_t + \boldsymbol{B}_t \mu_{\mathbb{U}}(\boldsymbol{u}_t) + \boldsymbol{w}_t, \quad \forall t \in \mathbb{T}^- \tag{32}$$

$$\boldsymbol{u}_t = \bar{\boldsymbol{u}}_t + \boldsymbol{K}_t(\boldsymbol{x}_t - \bar{\boldsymbol{x}}_t), \quad \forall t \in \mathbb{T}^- \tag{33}$$

$$\bigwedge_{c \in \mathcal{C}} \Pr \left[ \bigwedge_{i \in \mathcal{I}_c(s)} \boldsymbol{h}_{c,i}^T \boldsymbol{x}_{t_i} - g_{c,i} \leq 0 \right] \geq 1 - \Delta_c. \tag{34}$$

$$\boldsymbol{x}_0 \sim \mathcal{N}(\bar{\boldsymbol{x}}_0, \boldsymbol{\Sigma}_{\boldsymbol{x}_0}), \quad \boldsymbol{w}_t \sim \mathcal{N}(\boldsymbol{0}, \boldsymbol{\Sigma}_{\boldsymbol{w}_t}), \quad \forall t \in \mathbb{T}^- \tag{35}$$

Section 4 solves Problem 4.

## 4. CCQSP Planning with a Convex State Space and a Fixed Schedule

This section presents the solution methods to Problem 4, which is the CCQSP planning problem with a convex state space and a fixed schedule, as shown in Figure 4-(a). When there are no obstacles in the environment and the execution time steps to achieve time-evolved goals are fixed, the CCQSP planning problem is reduced to a convex chance-constrained finite-horizon optimal control problem.

In our past work we presented the *risk allocation approach*, which conservatively approximates the chance-constrained finite-horizon optimal control problem by a tractable convex optimization problem (Ono & Williams, 2008a, 2008b; Blackmore & Ono, 2009). Although an optimal solution to the approximated convex optimization problem is not an exactly optimal solution to the original convex chance-constrained finite-horizon optimal control problem, its suboptimality is significantly smaller than previous approaches. This section gives a brief overview of the risk allocation approach, as well as the solution to the convex chance-constrained finite-horizon optimal control problem.

### 4.1 Deterministic Approximation of Problem 4

Evaluating whether a joint chance constraint (34) is satisfied requires computing an integral of a multivariate probability distribution over an arbitrary region, since the probability in (34) involves multiple constraints. Such an integral cannot be obtained in a closed form. We address this issue by decomposing the intractable joint chance constraint (34) into a set of *individual chance constraints*, each of which involves only a univariate probability distribution. The key feature of an individual chance constraint is that it can be transformed into an equivalent deterministic constraint that can be evaluated analytically.

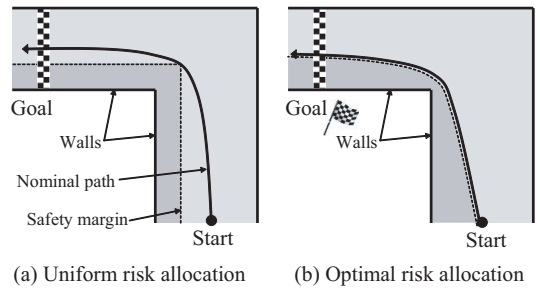

(a) Uniform risk allocation      (b) Optimal risk allocation

Figure 7: Risk allocation strategies on the racing car example

### 4.1.1 RISK ALLOCATION APPROACH

The decomposition can be considered as an allocation of risk. Through the decomposition, the risk bound of the joint chance constraint is distributed to the individual chance constraints. There are many feasible risk allocations. The problem is to find a risk allocation that results in the minimum cost. We offer readers an intuitive understanding of the risk allocation approach using the example below.

**Racing Car Example**  Consider a racing car example, shown in Figure 7. The dynamics of the vehicle have Gaussian-distributed uncertainty. The task is to plan a path that minimizes the time to reach the goal, with the guarantee that the probability of crashing into a wall during the race is less than 0.1% (*chance constraint*). Planning the control sequence is equivalent to planning the nominal path, which is shown as the solid lines in Figure 7. To limit the probability of crashing into the wall, a good driver would set a safety margin, which is colored in dark gray in Figure 7, and then plan the nominal path outside of the safety margin.

The driver wants to set the safety margin as small as possible in order to make the nominal path shorter. However, since the probability of crashing during the race is bounded, there is a certain lower bound on the size of the safety margin. Given this constraint, there are different ways of setting a safety margin; in Figure 7(a) the width of the margin is uniform; in Figure 7(b) the safety margin is narrow around the corner, and wide at the other places.

An intelligent driver would take the strategy of (b), since he knows that going closer to the wall at the corner makes the path shorter, while doing so at the straight line does not. A key observation here is that taking a risk (i.e., setting a narrow safety margin) at the corner results in a greater reward (i.e. time saving) than taking the same risk at the straight line. This gives rise to the notion of risk allocation. The good risk allocation strategy is to save risk when the reward is small, while taking it when the reward is great. As is illustrated in this example, the risk allocation must be optimized in order to minimize the objective function of a joint chance-constrained stochastic optimization problem.

#### 4.1.2 Decomposition of Conjunctive Joint Chance Constraints through Risk Allocation

We derive the mathematical representation of risk allocation by reformulating each chance constraint over a conjunction of constraints into a conjunction of chance constraints. The reformulation was initially presented by Prékopa (1999) and introduced to chance-constrained optimal control by Ono and Williams (2008b). The concept of risk allocation was originally developed by Ono and Williams (2008a). Let $C_i$ be a proposition that is either true or false. Then the following lemma holds:

**Lemma 1.**

$$\Pr\left[\bigwedge_{i=1}^{N} C_i\right] \geq 1 - \Delta \quad \Leftarrow \quad \exists \delta_i \geq 0, \ \bigwedge_{i=1}^{N} \Pr[C_i] \geq 1 - \delta_i \ \wedge \ \sum_{i=1}^{N} \delta_i \leq \Delta$$

*Proof.*

$$\Pr\left[\bigwedge_{i=1}^{N} C_i\right] \geq 1 - \Delta \ \Leftrightarrow \ \Pr\left[\bigvee_{i=1}^{N} \overline{C_i}\right] \leq \Delta \tag{36}$$

$$\Leftarrow \ \bigwedge_{c \in \mathcal{C}} \sum_{i=1}^{N} \Pr\left[\overline{C_i}\right] \leq \Delta \tag{37}$$

$$\Leftrightarrow \ \exists \delta_i \geq 0 \quad \bigwedge_{i=1}^{N} \Pr\left[\overline{C_i}\right] \leq \delta_i \wedge \sum_{i=1}^{N} \delta_i \leq \Delta$$

$$\Leftrightarrow \ \exists \delta_i \geq 0 \quad \bigwedge_{i=1}^{N} \Pr\left[C_i\right] \geq 1 - \delta_i \wedge \sum_{i=1}^{N} \delta_i \leq \Delta. \tag{38}$$

The overline $\overline{C}$ is the negation of a literal $C$. We use the following Boole's inequality to obtain (37) from (36):

$$\Pr\left[\bigvee_{i=1}^{N} C_{c,i}\right] \leq \sum_{i=1}^{N} \Pr[C_{c,i}].$$

$\square$

The following result immediately follows from Lemma 1 by substituting a linear constraint $\boldsymbol{h}_{c,i}^T \boldsymbol{x}_{t_i} - g_{c,i} \leq 0$ for $C_i$ for each chance constraint $c$.

**Corollary 1.** *The following set of constraints is a sufficient condition of the joint chance constraint (34) in Problem 4:*

$$\exists \delta_{c,i} \geq 0 \quad \bigwedge_{c \in \mathcal{C}} \left\{ \bigwedge_{i \in \mathcal{I}_c(s)} \Pr\left[\boldsymbol{h}_{c,i}^T \boldsymbol{x}_{t_i} - g_{c,i} \leq 0\right] \geq 1 - \delta_{c,i} \ \wedge \ \sum_{i \in \mathcal{I}_c(s)} \delta_{c,i} \leq \Delta_c \right\} \tag{39}$$

The newly introduced variables $\delta_{c,i}$ represent the upper bounds on the probability of violating each linear state constraint. We refer to them as *individual risk bounds*. Each individual risk bound, $\delta_{c,i}$, can be viewed as the amount of risk *allocated* to the $i$'th clause. The fact that $\delta_{c,i}$ is a bound on probability implies that $0 \leq \delta_{c,i} \leq 1$. The second term of (39) requires that the total amount of risk is upper-bounded to the original risk bound $\Delta_c$. Here we find an analogue to the resource allocation problem, where the allocation of a resource is optimized with an upper bound on the total amount of available resource. Likewise, the allocation of risk $\delta_{c,i}$ must be optimized in order to minimize the cost. Therefore, we call this decomposition method a risk allocation.

### 4.1.3 CONSERVATISM OF RISK ALLOCATION APPROACH

As mentioned previously, the risk allocation approach gives a conservative approximation of the original chance constraint. This subsection evaluates the level of conservatism of the risk allocation approach.

Let $P_{fail}$ be the true probability of failure, defined as the probability that a solution violates the constraints (i.e., the left hand side of (34)). Since (39) is a sufficient but not necessary condition for (34), $P_{fail}$ is smaller than or equal to the risk bound $\Delta$ in general: $\Delta \geq P_{fail}$. Hence, the conservatism introduced by risk allocation is represented as

$$\Delta - P_{fail}.$$

The best-case scenario for the risk allocation approach is when the violations of all constraints are mutually exclusive, meaning that a solution that violates one constraint always satisfies all the other constraints. In that case, (39) becomes a *necessary and* sufficient condition for (34) and hence, risk allocation does not involve any conservatism. Therefore,

$$\Delta - P_{fail} = 0.$$

On the other hand, the worst-case scenario is when all constraints are equivalent, meaning that a solution that violates one constraint always violates all the other constraints. In such a case,

$$\Delta - P_{fail} = \frac{N-1}{N}\Delta,$$

where $N$ is the number of constraints.

Most practical problems lie somewhere between the best-case scenario and the worst-case scenario, but typically closer to the best-case than to the worst-case scenario. For example, if there are two separate obstacles in a path planning problem, collisions with the two obstacles are mutually exclusive events. Collision with an obstacle at one time step does not usually imply collisions at other time steps. A rough approximation of such a real-world situation is to assume that the satisfaction of constraints are probabilistically independent. With such an assumption, the true probability of failure is:

$$P_{fail} = \prod_{i \in \mathcal{I}_c} \Pr\left[q_{c,i}(u) \leq 0\right] \leq 1 - \prod_{i \in \mathcal{I}_c}(1 - \delta_i),$$

where $\mathcal{I}_c$ is the set of the index of all state constraints. Note that $\delta_i \leq \Delta$. Therefore, the conservatism introduced by risk allocation is at the second order of $\Delta$:

$$\Delta - P_{fail} \sim \mathcal{O}(\Delta^2).$$

For example, if $\Delta = 1\%$, the true probability of failure is approximately $P_{fail} \sim 0.99\%$. In most practical cases, the users prefer to set very small risk bounds, typically less than $1\%$. In such cases, the conservatism introduced by risk allocation becomes very small.

### 4.1.4 CONVERSION TO DETERMINISTIC CONSTRAINTS

Each individual chance constraint in (39) only involves a single linear constraint. Furthermore, assuming that there is no actuator saturation, $\boldsymbol{x}_{t_i}$ has a Gaussian distribution with the covariance matrix given by (9). Hence, $\boldsymbol{h}_{c,i}^T \boldsymbol{x}_{t_i}$ has a univariate Gaussian distribution. The following lemma transforms an individual chance constraint into an equivalent deterministic constraint that involves the mean of state variables, instead of the random state variables:

**Lemma 2.** *The following two conditions are equivalent.*

$$\Pr\left[\boldsymbol{h}_{c,i}^T \boldsymbol{x}_{t_i} - g_{c,i} \leq 0\right] \geq 1 - \delta_{c,i} \quad \Leftrightarrow \quad \boldsymbol{h}_{c,i}^T \bar{\boldsymbol{x}}_{t_i} - g_{c,i} \leq -m_{c,i}(\delta_{c,i})$$

*where*

$$m_{c,i}(\delta_{c,i}) = -\sqrt{2\boldsymbol{h}_{c,i}^T \boldsymbol{\Sigma}_{x,t_i} \boldsymbol{h}_{c,i}} \; \mathrm{erf}^{-1}(2\delta_{c,i} - 1). \tag{40}$$

Note that $\mathrm{erf}^{-1}$ is the inverse of the Gauss error function and $\boldsymbol{\Sigma}_{x,t_i}$ is the covariance matrix of $\boldsymbol{x}_{t_i}$. This lemma holds because $-m_{c,i}(\cdot)$ is the inverse of cumulative distribution function of univariate, zero-mean Gaussian distribution with variance $\boldsymbol{h}_{c,i}^T \boldsymbol{\Sigma}_{x,t_i} \boldsymbol{h}_{c,i}$.

### 4.1.5 RISK ALLOCATION APPROACH FOR THE CLOSED-LOOP CONTROL POLICY

When a close-loop control policy is employed (i.e., $\boldsymbol{K}_t \neq 0$ in (6)), there is a risk of actuator saturation. Since the nonlinearity of the function $\mu_{\mathbb{U}}(\cdot)$ in (5) makes the probability distribution of $\boldsymbol{x}_{t_i}$ non-Gaussian, $m_{c,i}(\cdot)$ cannot be obtained by (40). Although it is theoretically possible to derive $m_{c,i}(\cdot)$ for non-Gaussian distributions, it is very difficult in our case since the inverse of the cumulative distribution function of $\boldsymbol{x}_{t_i}$ cannot be obtained in a closed-form.

Our solution to this issue is summarized in Lemma 3 below, which allows us to assume that $\boldsymbol{x}_{t_i}$ is Gaussian-distributed and hence to use (40), even if there is a possibility of actuator saturation. This approach is enabled by imposing additional chance constraints that bound the risk of actuator saturation as follows:

$$\Pr\left[\boldsymbol{u}_t \in \mathbb{U}\right] \geq 1 - \epsilon_t, \quad \forall t \in \mathbb{T}^-, \tag{41}$$

where $\epsilon_t$ is the bound on the risk of actuator saturation at time step $t$. Using the method presented in Section 3.1.2, we approximate $\mathbb{U}$ by a polytope as follows:

$$\boldsymbol{u}_t \in \mathbb{U} \Longleftrightarrow \bigwedge_{i \in \mathcal{I}_{\mathbb{U}}} \boldsymbol{h}_{\mathbb{U},i} \boldsymbol{u}_t - g_{\mathbb{U},i} \leq 0$$

Assuming that $\boldsymbol{x}_{t_i}$ is Gaussian-distributed, we use Lemma 2 to transform (41) into deterministic constraints on *nominal* control inputs as follows:

$$\bigwedge_{i \in \mathcal{I}_{\mathbb{U}}} \boldsymbol{h}_{\mathbb{U},i} \bar{\boldsymbol{u}}_t - g_{\mathbb{U},i} \leq -m_{\mathbb{U},t,i}(\epsilon_{t,i}) \; \wedge \; \sum_{i \in \mathcal{I}_{\mathbb{U}}} \epsilon_{t,i} \leq \epsilon_t, \quad \forall t \in \mathbb{T}^-, \tag{42}$$

where

$$m_{\mathbb{U},t,i}(\epsilon_{c,i}) = -\sqrt{2\boldsymbol{h}_{\mathbb{U},i}^T \boldsymbol{\Sigma}_{x,t} \boldsymbol{h}_{\mathbb{U},i}} \; \mathrm{erf}^{-1}(2\epsilon_{c,i} - 1). \tag{43}$$

The following lemma holds:

**Lemma 3.** *The following set of constraints is a sufficient condition of the joint chance constraint (34) in Problem 4:*

$$\exists \delta_{c,i} \geq 0, \epsilon_t \geq 0 \quad \bigwedge_{c \in \mathcal{C}} \Bigg\{ \bigwedge_{i \in \mathcal{I}_c(s)} \boldsymbol{h}_{c,i}^T \bar{\boldsymbol{x}}_{t_i} - g_{c,i} \leq -m_{c,i}(\delta_{c,i})$$

$$\wedge \sum_{i \in \mathcal{I}_c(s)} \delta_{c,i} + \sum_{t=0}^{T_c^{\max}} \sum_{i \in \mathcal{I}_{\mathbb{U}}} \epsilon_{t,i} \leq \Delta_c \Bigg\}$$

$$\wedge \bigwedge_{t \in \mathbb{T}^-} \bigwedge_{i \in \mathcal{I}_{\mathbb{U}}} \boldsymbol{h}_{\mathbb{U},i} \bar{\boldsymbol{u}}_t - g_{\mathbb{U},i} \leq -m_{\mathbb{U},t,i}(\epsilon_{t,i}), \qquad (44)$$

$$(45)$$

*where $m_{c,i}(\cdot)$ and $m_{\mathbb{U},t,i}$ are given by (40) and (43). $T_c^{\max}$ is the last time step that the episodes associated with the chance constraint $c$ are executed, given the schedule $s$:*

$$T_c^{\max} = \max_{a \in \Psi_c} s(e_a^E).$$

*Intuitively, the constraint (44) requires that, with probability $1 - \Delta_c$, the episode constraints are satisfied and the actuators do not saturate until all episodes associated with $c$ are executed.*

*Proof.* We consider two plants: $M = \langle \boldsymbol{A}_{0:N-1}, \boldsymbol{B}_{0:N-1}, \boldsymbol{\Sigma}_{w_{0:N-1}}, \mathbb{U} \rangle$ and $M' = \langle \boldsymbol{A}_{0:N-1}, \boldsymbol{B}_{0:N-1}, \boldsymbol{\Sigma}_{w_{0:N-1}}, \mathbb{R}^{n_u} \rangle$, where $\mathbb{U} \subset \mathbb{R}^{n_u}$ is a compact convex set (see Definition 6). The difference between the two plants is that $M$ has a possibility of actuator saturation, while $M'$ does not. As a result, while the probability distribution of the state variables of $M$ is non-Gaussian, that of $M'$ is Gaussian. Note that $M$ and $M'$ result in different probability distributions of $\boldsymbol{x}_{t_i}$ and $\boldsymbol{u}_t$. In order to explicitly show which plant model is considered, we use notations such as $\boldsymbol{x}_{t_i}^M$ and $\boldsymbol{u}_t^{M'}$ in this proof.

We first consider $M'$. It follows from Lemmas 1 and 2 that:

$$(44) \Longrightarrow \bigwedge_{c \in \mathcal{C}} \Bigg\{ \Pr\Bigg[ \Bigg( \bigwedge_{i \in \mathcal{I}_c(s)} \boldsymbol{h}_{c,i}^T \boldsymbol{x}_{t_i}^{M'} - g_{c,i} \leq 0 \Bigg) \wedge \Bigg( \bigwedge_{t=0}^{T_c^{\max}} \boldsymbol{u}_t^{M'} \in \mathbb{U} \Bigg) \Bigg] \geq 1 - \Delta_c \Bigg\}.$$

Let $\boldsymbol{w}_{0:N-1} := [\boldsymbol{w}_0 \cdots \boldsymbol{w}_{N-1}]$. We define a *feasible disturbance set*, $W_c(\boldsymbol{v}_{0:N-1}, s) \subset \mathbb{R}^{N n_x}$, as follows:

$$W_c(\boldsymbol{v}_{0:N-1}, s) := \Bigg\{ \boldsymbol{w}_{0:N-1} \in \mathbb{R}^{N n_x} \Bigg| \Bigg( \bigwedge_{i \in \mathcal{I}_c(s)} \boldsymbol{h}_{c,i}^T \boldsymbol{x}_{t_i}^{M'} - g_{c,i} \leq 0 \Bigg) \wedge \Bigg( \bigwedge_{t=0}^{T_c^{\max}} \boldsymbol{u}_t^{M'} \in \mathbb{U} \Bigg) \Bigg\}. \qquad (46)$$

Then, by definition,

$$\Pr\Bigg[ \Bigg( \bigwedge_{i \in \mathcal{I}_c(s)} \boldsymbol{h}_{c,i}^T \boldsymbol{x}_{t_i}^{M'} - g_{c,i} \leq 0 \Bigg) \wedge \Bigg( \bigwedge_{t=0}^{T_c^{\max}} \boldsymbol{u}_t^{M'} \in \mathbb{U} \Bigg) \Bigg] = \Pr\left[ \boldsymbol{w}_{0:N-1} \in W_c(\boldsymbol{v}_{0:N-1}, s) \right].$$

Next we consider $M$. Note that $M$ and $M'$ are identical as long as there is no actuator saturations (i.e., $\boldsymbol{u}_t^M \in \mathbb{U}$). Therefore, for a given $\boldsymbol{w}_{0:N-1} \in W_c(\boldsymbol{v}_{0:N-1}, s)$, it follows from (46) that $\boldsymbol{x}_t^M = \boldsymbol{x}_t^{M'}$ and $\boldsymbol{u}_t^M = \boldsymbol{u}_t^{M'}$. Hence,

$$\boldsymbol{w}_{0:N-1} \in W_c(\boldsymbol{v}_{0:N-1}, s) \Longrightarrow \left( \bigwedge_{i \in \mathcal{I}_c(s)} \boldsymbol{h}_{c,i}^T \boldsymbol{x}_{t_i}^M - g_{c,i} \leq 0 \right) \wedge \left( \bigwedge_{t=0}^{T_c^{\max}} \boldsymbol{u}_t^M \in \mathbb{U} \right).$$

Accordingly, for a given $c \in \mathcal{C}$,

$$\Pr \left[ \left( \bigwedge_{i \in \mathcal{I}_c(s)} \boldsymbol{h}_{c,i}^T \boldsymbol{x}_{t_i}^M - g_{c,i} \leq 0 \right) \right]$$

$$\geq \Pr \left[ \left( \bigwedge_{i \in \mathcal{I}_c(s)} \boldsymbol{h}_{c,i}^T \boldsymbol{x}_{t_i}^M - g_{c,i} \leq 0 \right) \wedge \left( \bigwedge_{t=0}^{T_c^{\max}} \boldsymbol{u}_t^M \in \mathbb{U} \right) \right]$$

$$\geq \Pr \left[ \boldsymbol{w}_{0:N-1} \in W_c(\boldsymbol{v}_{0:N-1}, s) \right]$$

$$= \Pr \left[ \left( \bigwedge_{i \in \mathcal{I}_c(s)} \boldsymbol{h}_{c,i}^T \boldsymbol{x}_{t_i}^{M'} - g_{c,i} \leq 0 \right) \wedge \left( \bigwedge_{t=0}^{T_c^{\max}} \boldsymbol{u}_t^{M'} \in \mathbb{U} \right) \right]$$

$$\geq 1 - \Delta_c.$$

This completes the proof of Lemma 3 □

We note that Lemma 3 is a probabilistic extension of the closed-loop robust model predictive control (RMPC) methods proposed by Acikmese, Carson III, and Bayard (2011) and Richards and How (2006). These methods avoid the risk of actuator saturation by imposing tightened control constraints on $\bar{\boldsymbol{u}}_t$. Since we consider stochastic uncertainty, we replace the constraint tightening by chance constraints.

## 4.2 Convex Programming Solution to Problem 4

Using Lemma 3, we replace the stochastic optimization problem, Problem 4, with the deterministic convex optimization problem:

**Problem 5: Deterministic Approximation of Problem 4**

$$\min_{\bar{\boldsymbol{u}}_{1:N}, \delta_{c,i} \geq 0, \epsilon_{t,i} \geq 0} \quad J'(\boldsymbol{u}_{1:N}, \bar{\boldsymbol{x}}_{1:N}) \tag{47}$$

$$\text{s.t.} \quad \forall t \in \mathbb{T}^-, \quad \bar{\boldsymbol{x}}_{t+1} = \boldsymbol{A}_t \bar{\boldsymbol{x}}_t + \boldsymbol{B}_t \boldsymbol{u}_t \tag{48}$$

$$\bigwedge_{c \in \mathcal{C}} \bigwedge_{i \in \mathcal{I}_c(s)} \boldsymbol{h}_{c,i}^T \bar{\boldsymbol{x}}_{t_i} - g_{c,i} \leq -m_{c,i}(\delta_{c,i}) \tag{49}$$

$$\bigwedge_{t \in \mathbb{T}^-} \bigwedge_{i \in \mathcal{I}_{\mathbb{U}}} \boldsymbol{h}_{\mathbb{U},i} \bar{\boldsymbol{u}}_t - g_{\mathbb{U},i} \leq -m_{\mathbb{U},t,i}(\epsilon_{t,i}) \tag{50}$$

$$\bigwedge_{c \in \mathcal{C}} \sum_{i \in \mathcal{I}_c(s)} \delta_{c,i} + \sum_{t=0}^{T_c^{\max}} \sum_{i \in \mathcal{I}_{\mathbb{U}}} \epsilon_{t,i} \leq \Delta_c. \tag{51}$$

It follows immediately from Corollaries 1 and 2 that a feasible solution to Problem 5 is always a feasible solution to Problem 4. Furthermore, Blackmore and Ono (2009) showed that an optimal solution to Problem 5 is a near-optimal solution to Problem 4. The following lemma guarantees the tractability of Problem 5.

**Lemma 4.** *Problem 5 is a convex optimization problem.*

*Proof.* The inverse error function $\mathrm{erf}^{-1}(x)$ is concave for $x$. Since we assume in Section 2.4.3 that $\Delta_c \leq 0.5$, the feasible ranges of $\delta$ and $\epsilon$ are upperbounded by 0.5. Since the safety margin function $m_{c,i}(\delta_{c,i})$ and $m_{\mathbb{U},t,i}(\epsilon_{t,i})$ are convex for $0 < \delta_{c,i} \leq 0.5$ and $0 < \epsilon_{t,i} \leq 0.5$, the constraints (49) and (50) are convex within the feasible region. All other constraints are also convex since they are linear. Finally, the objective function is convex by assumption (Section 2.4.4). Therefore, Problem 5 is a convex optimization problem. □

Since Problem 5 is a convex optimization problem, it can be solved by an interior point method optimally and efficiently. This completes our first spiral, planning for CCQSPs with a fixed schedule and convex constraints. In the next section we present a solution method for a non-convex problem through a branch-and-bound algorithm, whose subproblems are convex problems.

## 5. CCQSP Planning with a Non-convex State Space

Next, we consider the second spiral, comprised of Problem 3 in Section 3.2, a variant of the CCQSP planning problem that involves a fixed schedule and non-convex constraints, such as obstacles, as shown in Figure 4-(b). Once again, this is encoded as a chance-constrained optimization problem, but the addition of the obstacle avoidance constraints requires disjunctive state constraints. Hence, the problem results in a non-convex, chance-constrained optimization. This section introduces a novel algorithm, called Non-convex Iterative Risk Allocation (NIRA), that optimally solves a deterministic approximation of Problem 3.

The solution to a CCQSP planning problem with a non-convex state space is two-fold. In the first step, described in Section 5.1, we obtain a deterministic approximation of Problem 3. In order to handle disjunctive chance constraints, we develop an additional decomposition approach called *risk selection*, which reformulates each chance constraint over a disjunction of constraints into a disjunction of individual chance constraints. Once the chance constraints in (29) are decomposed into a set of individual chance constraints through risk allocation and risk selection, the same technique as in Section 4.1.4 is used to obtain equivalent deterministic constraints. As a result, we obtain a disjunctive convex programming problem (Problem 6 in Section 5.1.3).

The deterministic disjunctive convex programming problem is solved in the second step, described in Sections 5.2-5.4. We introduce the NIRA algorithm (Algorithm 1) that significantly reduces the computation time without making any compromise in the optimality of the solution. The reduction in computation time is enabled by our new bounding approach, Fixed Risk Relaxation (FRR). FRR relaxes nonlinear constraints in the subproblems of the branch-and-bound algorithm with linear constraints. In many cases, FRR of the nonlinear subproblems is formulated as a linear programming (LP) or approximated by an LP. NIRA obtains a strictly optimal solution of Problem 6 by solving the subproblems *exactly* without FRR at unpruned leaf nodes of the search tree, while other subproblems are solved approximately with FRR in order to reduce the computation time.

## 5.1 Deterministic Approximation

As in Section 4, we first obtain a deterministic approximation of Problem 3.

### 5.1.1 RISK SELECTION APPROACH

The deterministic approximation is obtained by decomposing the non-convex joint chance constraint (29) into a set of individual chance constraints, through risk allocation and *risk selection*. We revisit the race car example to explain the concept of risk selection intuitively.

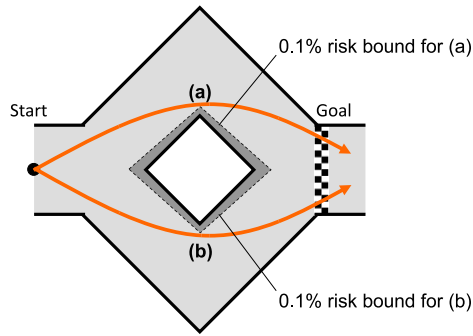

Figure 8: In the racing car example, the risk selection approach guarantees the $0.1\%$ risk bound for both paths, and lets the vehicle choose the better one.

**Racing Car Example**    We consider the example shown in Figure 8, where a vehicle with uncertain dynamics plans a path that minimizes the time to reach the goal. The vehicle is allowed to choose one of the two routes shown in Figure 8. We impose a chance constraint that limits the probability of crashing into a wall during the mission to $0.1\%$.

The satisfaction of the chance constraint can be guaranteed by the following process. First, for each of the routes, we find a safety margin that limits the probability of crash throughout the route to $0.1\%$ from the start to the goal. Then, we let the vehicle plan a nominal path that operates within the safety margins. Since both routes have a $0.1\%$ safety margin, the chance constraint is satisfied no matter which route the vehicle chooses. Therefore, the vehicle can optimize the path by choosing the route that results in a smaller cost. The optimization process can be considered as a selection of risk; the vehicle is given two options as in Figure 8, routes (a) and (b), both of which involve the same level of risk; then the vehicle selects the one that results in less cost. Hence, we name this decomposition approach as the risk selection.

### 5.1.2 DECOMPOSITION OF CONJUNCTIVE JOINT CHANCE CONSTRAINT THROUGH RISK SELECTION

In this subsection, we derive the mathematical representation of risk selection. Let $C_i$ be a proposition that is either true or false. Then the following lemma holds:

**Lemma 5.**

$$\Pr\left[\bigvee_{i=1}^{N} C_i\right] \geq 1 - \Delta \quad \Leftarrow \quad \bigvee_{i=1}^{N} \Pr\left[C_i\right] \geq 1 - \Delta$$

*Proof.* The following inequality always holds:

$$\forall i \quad \Pr\left[\bigvee_{i=1}^{N} C_i\right] \geq \Pr\left[C_i\right]. \tag{52}$$

Hence,

$$\Pr\left[\bigvee_{i=1}^{N} C_i\right] \geq 1 - \Delta \Leftarrow \exists i \ \Pr\left[C_i\right] \geq 1 - \Delta \Leftrightarrow \bigvee_{i=1}^{N} \Pr\left[C_i\right] \geq 1 - \Delta. \tag{53}$$

$$\square$$

The following corollary follows immediately from Lemmas 3 and 5.

**Corollary 2.** *The following set of constraints is a sufficient condition of the disjunctive joint chance constraint (29) in Problem 3:*

$$\exists \delta_{c,i} \geq 0, \epsilon_t \geq 0 \quad \bigwedge_{c \in \mathcal{C}} \left\{ \bigwedge_{i \in \mathcal{I}_c(s)} \bigvee_{j \in \mathcal{J}_{c,i}} \boldsymbol{h}_{c,i,j}^T \boldsymbol{x}_{t_i} - g_{c,i,j} \leq m_{c,i}(\delta_{c,i}) \right.$$

$$\wedge \sum_{i \in \mathcal{I}_c(s)} \delta_{c,i} + \sum_{t=0}^{T_c^{\max}} \sum_{i \in \mathcal{I}_\mathbb{U}} \epsilon_{t,i} \leq \Delta_c \left.\right\} 5$$

$$\wedge \bigwedge_{t \in \mathbb{T}^-} \bigwedge_{i \in \mathcal{I}_\mathbb{U}} \boldsymbol{h}_{\mathbb{U},i} \bar{\boldsymbol{u}}_t - g_{\mathbb{U},i} \leq -m_{\mathbb{U},t,i}(\epsilon_{t,i}). \tag{54}$$

Note that the resulting set of constraints (54) is a *sufficient* condition for the original chance constraint (29). Therefore, any solution that satisfies (54) is guaranteed to satisfy (29). Furthermore, although (54) is a conservative approximation of (29), the conservatism introduced by risk selection is generally small in many practical applications. This claim is empirically validated in Section 7.2.3.

### 5.1.3 DETERMINISTIC APPROXIMATION OF PROBLEM 3

Using Corollary 2, the non-convex fixed-schedule CCQSP planning problem (Problem 3) is approximated by the following deterministic convex optimization problem. For later convenience, we label each part of the optimization problem as $O$ (objective function), $M$ (plant model), $C$ (chance constraints on states), $D$ (chance constraints on control inputs), and $R$ (risk allocation constraint).

**Problem 6: Deterministic Approximation of Problem 3**

$$\min_{\bar{\boldsymbol{u}}_{1:N}, \delta_{c,i} \geq 0, \epsilon_{t,i} \geq 0} \quad (O:) \quad J'(\boldsymbol{u}_{1:N}, \bar{\boldsymbol{x}}_{1:N}) \tag{55}$$

$$\text{s.t.} \quad (M:) \quad \forall t \in \mathbb{T}^-, \quad \bar{\boldsymbol{x}}_{t+1} = \boldsymbol{A}_t \bar{\boldsymbol{x}}_t + \boldsymbol{B}_t \boldsymbol{u}_t \tag{56}$$

$$(C:) \quad \bigwedge_{c \in \mathcal{C}} \bigwedge_{i \in \mathcal{I}_c(s)} \bigvee_{j \in \mathcal{J}_{c,i}} \boldsymbol{h}_{c,i,j}^T \bar{\boldsymbol{x}}_{t_i} - g_{c,i,j} \leq -m_{c,i,j}(\delta_{c,i}) \tag{57}$$

$$(D:) \quad \bigwedge_{t \in \mathbb{T}^-} \bigwedge_{i \in \mathcal{I}_\mathbb{U}} \boldsymbol{h}_{\mathbb{U},i} \bar{\boldsymbol{u}}_t - g_{\mathbb{U},i} \leq -m_{\mathbb{U},t,i}(\epsilon_{t,i}) \tag{58}$$

$$(R:) \quad \bigwedge_{c \in \mathcal{C}} \sum_{i \in \mathcal{I}_c(s)} \delta_{c,i} + \sum_{t=0}^{T_c^{\max}} \sum_{i \in \mathcal{I}_\mathbb{U}} \epsilon_{t,i} \leq \Delta_c. \tag{59}$$

It follows immediately from Corollary 2 that an optimal solution to Problem 6 is guaranteed to be a feasible solution to the original problem with regard to satisfying the chance constraints (Problem 3). Furthermore, we empirically demonstrate in Section 7.2.3 that it is a near-optimal solution to Problem 3 in our applications.

## 5.2 NIRA: Branch and Bound-Based Solution to Problem 6

We next present the Non-convex Iterative Risk Allocation (NIRA) algorithm. Recall that NIRA optimally solves Problem 6 by a branch-and-bound algorithm. The standard branch-and-bound solution to problems involving disjunctive nonlinear constraints, such as those in Problem 6, is to use a bounding approach in which the nonlinear convex relaxed subproblems are constructed by removing all non-convex constraints below the corresponding disjunction. This approach was used by Balas (1979) and Li and Williams (2005) for a different problem known as disjunctive linear programming, whose subproblems are LPs instead of convex programmings. However, although the standard branch-and-bound algorithm is guaranteed to find a globally optimal solution to Problem 6, its computation time is slow because the algorithm needs to solve numerous nonlinear subproblems in order to compute relaxed bounds.

Our new bounding approach, Fixed Risk Relaxation (FRR), addresses this issue by computing lower bounds more efficiently. We observe that the relaxed subproblems are nonlinear convex optimization problems. FRR relaxes the nonlinear constraints to linear constraints. Particularly, when the objective function is linear, an FRR of a subproblem (Problem 8) is an LP, which can be very efficiently solved. The optimal objective value of an FRR of a subproblem is a lower bound of the optimal objective value of the original subproblem.

NIRA solves the FRRs of the subproblems in order to efficiently obtain the lower bounds, while solving the original subproblems *exactly* without relaxation at unpruned leaf nodes in order to obtain an exact optimal solution. As a result, NIRA achieves significant reduction in computation time, without any loss in optimality.

### 5.2.1 THE NIRA ALGORITHM OVERVIEW

Algorithm 1 shows the pseudocode of the NIRA algorithm. Its input is the deterministic approximation of a non-convex chance-constrained optimal control problem (Problem 6), which is a five-tuple

---

**Algorithm 1** Non-convex Iterative Risk Allocation (NIRA) algorithm

---

**function** NIRA($problem$) **returns** optimal solution to Problem 6

 1: Set up $queue$ as a FILO queue
 2: $Incumbent \leftarrow \infty$
 3: $rootSubproblem \leftarrow$ obtainRootSubproblem($problem$)
 4: $queue \leftarrow rootSubproblem$
 5: **while** $queue$ is not empty **do**
 6:    $subproblem \leftarrow$ the last entry in $queue$
 7:    Remove $subproblem$ from $queue$
 8:    $lb \leftarrow$ obtainLowerBound($subproblem$)
 9:    **if** $lb \leq Incumbent$ **then**
10:       **if** $c = |\mathcal{C}| \wedge i = |\mathcal{I}_c(s)|$ **then**
11:          $(J, \bar{U}) \leftarrow$ Solve($subproblem$)
12:          **if** $J^{\star} < Incumbent$ **then**
13:             $Incumbent \leftarrow J, \quad \bar{U}^{\star} \leftarrow \bar{U}$ //Update the optimal solution
14:          **end if**
15:       **else**
16:          $i \leftarrow i + 1$
17:          **if** $i > |\mathcal{I}_c(s)|$ **then**
18:             $c \leftarrow c + 1, i \leftarrow 1$
19:          **end if**
20:          **for** $j \in \mathcal{J}_{c,i}$ **do**
21:             $newSubproblems \leftarrow$ Expand($subproblem$,$problem$,c,i,j)
22:             Add $newSubproblems$ to $queue$
23:          **end for**
24:       **end if**
25:    **end if**
26: **end while**
27: **return** $\bar{U}^{\star}$

---

$\langle O, M, C, D, R \rangle$, as well as a fixed schedule $s$. Its output is an optimal nominal control sequence $\bar{U}^{\star} := [\bar{u}_0^{\star} \cdots \bar{u}_{N-1}^{\star}]$.

Each node of the branch-and-bound search tree corresponds to a subproblem that is a convex chance-constrained optimization problem (Problem 5). We use a FILO queue to store subproblems so that the search is conducted in a depth-first manner (Line 1). At each node, the corresponding subproblem is solved to obtain a lower bound of the objective value of all subsequent subproblems (Line 8). The details of the bounding approaches are explained in Subsection 5.4. If the lower bound is larger than the incumbent, the algorithm prunes the branch. Otherwise, the branch is expanded (Line 21). If a branch is expanded to the leaf without being pruned, subproblems are solved exactly (Line 11). Subsection 5.3 explains our expansion procedure in detail. The NIRA algorithm always results in a globally optimal solution to Problem 6, since the solution $\bar{U}^{\star}$ is obtained by solving the subproblems at leaf nodes exactly. The next two subsections introduces the branching and bounding methods.

### 5.3 Branching

This subsection explains how NIRA constructs the root subproblem (Line 3 of Algorithm 1), as well as how it expands the nodes (Line 21 of Algorithm 1). The root subproblem is a convex optimal CCQSP planning problem without any chance constraints. When a node is expanded, the subproblems of its children nodes are constructed by adding one constraint in a disjunction to the subproblem of the parent node. In order to simplify notations, we let $C_{c,i,j}$ represent each individual chance constraint (57) in Problem 6:

$$C_{c,i,j} := \begin{cases} True & (\text{if } \boldsymbol{h}_{c,i,j}^T - g_{c,i,j}\bar{\boldsymbol{x}}_{t_i} \leq -m_{c,i,j}(\delta_{c,i})) \\ False & (\text{otherwise}). \end{cases}$$

#### 5.3.1 Walk-through Example

We first present a walk-through example to intuitively explain the branching procedure. The example is an instance of Problem 6, which involves four individual chance constraints:

$$\bigwedge_{i \in \{1,2\}} \bigvee_{j \in \{1,2\}} \boldsymbol{h}_{1,i,j}^T \bar{\boldsymbol{x}}_{t_i} - g_{1,i,j} \leq -m_{1,i,j}(\delta_{1,i}) \tag{60}$$

Using this notation defined above, the set of individual chance constraints (57) is represented as follows:

$$(C_{1,1,1} \vee C_{1,1,2}) \wedge (C_{1,2,1} \vee C_{1,2,2}) \tag{61}$$

Figure 9-(a) shows a tree obtained by dividing the original problem into subproblems sequentially. The subproblems corresponding to the tree's four leaf nodes (Nodes 4-7 in Figure 9-(a)) exhaust all conjunctive (i.e., *convex*) combinations among the chance constraints (61). On the other hand, the subproblems corresponding to the three branch nodes (Nodes 1-3 in Figure 9-(a)) involve disjunctive (i.e., *nonconvex*) clauses of chance constraints. We relax such non-convex subproblems to convex subproblems by removing all clauses that contain disjunctions in order to obtain the search tree shown in Figure 9-(b).

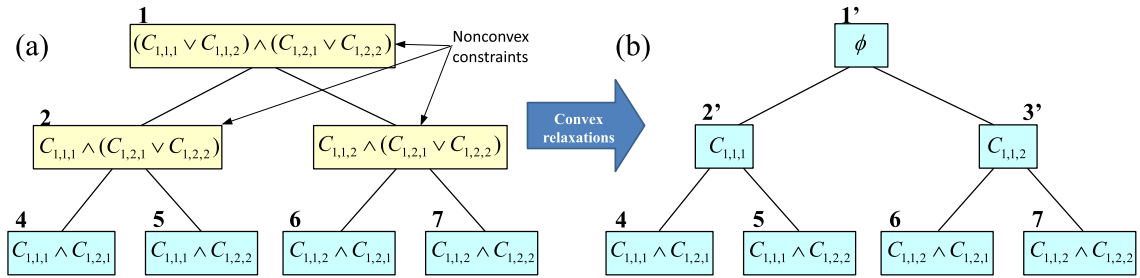

Figure 9: Branch-and-bound search tree for a sample disjunctive convex programming problem (Problem 6) with constraints (60). (a) Tree of non-convex subproblems, (b) Tree of relaxed convex subproblems.

The non-convex problem (Problem 6) can be optimally solved by repeatedly solving the relaxed convex subproblems using the algorithms presented in Section 4. The following subsections introduce the algorithms that construct a search tree with relaxed *convex* subproblems, such as the one in Figure 9-(b).

### 5.3.2 RELAXED CONVEX SUBPROBLEM

The formulation of the relaxed convex subproblems is given in Problem 7. We represent the index $j$ as $j(c, i)$ since the convex relaxation chooses only one disjunct for each disjunction specified by $(c, i)$. Let $I_c$ be a set of indices for $i$. We denote by $J_{SP}^\star$ the optimal objective value of the relaxed subproblem.

**Problem 7: Convex Relaxed Subproblem of NIRA**

$$
\begin{aligned}
J_{SP}^\star = \min_{\bar{\boldsymbol{u}}_{1:N}, \delta_{c,i} \geq 0, \epsilon_{t,i} \geq 0} \quad & (O:) \quad J'(\boldsymbol{u}_{1:N}, \bar{\boldsymbol{x}}_{1:N}) \\
\text{s.t.} \quad & (M:) \quad \forall t \in \mathbb{T}^-, \quad \bar{\boldsymbol{x}}_{t+1} = \boldsymbol{A}_t \bar{\boldsymbol{x}}_t + \boldsymbol{B}_t \boldsymbol{u}_t \\
& (C:) \quad \bigwedge_{c \in \mathcal{C}} \bigwedge_{i \in I_c} \boldsymbol{h}_{c,i,j(c,i)}^T \bar{\boldsymbol{x}}_{t_i} - g_{c,i,j(c,i)} \leq -m_{c,i,j(c,i)}(\delta_{c,i}) \quad (62) \\
& (D:) \quad \bigwedge_{t \in \mathbb{T}^-} \bigwedge_{i \in \mathcal{I}_{\mathbb{U}}} \boldsymbol{h}_{\mathbb{U},i} \bar{\boldsymbol{u}}_t - g_{\mathbb{U},i} \leq -m_{\mathbb{U},t,i}(\epsilon_{t,i}) \quad (63) \\
& (R:) \quad \bigwedge_{c \in \mathcal{C}} \sum_{i \in I_c} \delta_{c,i} + \sum_{t=0}^{T_c^{\max}} \sum_{i \in \mathcal{I}_{\mathbb{U}}} \epsilon_{t,i} \leq \Delta_c. \quad (64)
\end{aligned}
$$

Note that Problem 7 is identical to Problem 5. Hence, the algorithms introduced in Section 4 can be used to solve the relaxed subproblems.

### 5.3.3 CONSTRUCTION OF ROOT SUBPROBLEM

The root subproblem is a special case of Problem 7 above with $I_c$ being an empty set for all $c \in \mathcal{C}$. The function presented in Algorithm 2 is used in Line 3 of the NIRA algorithm (Algorithm 1) to construct the root subproblem of the branch-and-bound tree. Note that, in Algorithm 2, we use an object-oriented notation, such as *subproblem.O*, to represent the objective function $O$ of *subproblem*. The resulting root subproblem is as follows:

### 5.3.4 EXPANSION OF SUBPROBLEMS

In order to create a child subproblem of a subproblem, the function described in Algorithm 3 is used in Line 21 of the NIRA algorithm (Algorithm 1). It adds the individual chance constraint specified by the indices $(c, i, j)$ as a conjunct. Note that the resulting child subproblem is still a convex optimization, because the individual chance constraint is added conjunctively. The NIRA algorithm (Algorithm 1) enumerates children nodes for all disjuncts in $\mathcal{J}_{c,i}$ (Lines 20-23).

---

**Algorithm 2** Construction of the root subproblem of NIRA

---

**function** obtainRootSubproblem($problem$) **returns** root subproblem

1: $rootSubproblem.O \leftarrow problem.O$
2: $rootSubproblem.M \leftarrow problem.M$
3: $rootSubproblem.D \leftarrow problem.D$
4: **for** $c \in \mathcal{C}$ **do**
5: $\quad rootSubproblem.I_c \leftarrow \phi$
6: $\quad rootSubproblem.R_c.lhs \leftarrow \sum_{t=0}^{T_c^{\max}} \sum_{i \in \mathcal{I}_{\mathbb{U}}} \epsilon_{t,i}$
7: $\quad rootSubproblem.R_c.rhs \leftarrow problem.R_c.rhs$
8: **end for**
9: **return** $rootSubproblem$

---

**Algorithm 3** Expansion of a subproblem of NIRA

---

**function** Expand($subproblem, problem, c, i, j$) **returns** a child subproblem

1: $subproblem.I_c \leftarrow subproblem.I_c \cup i$
2: $subproblem.R_c.lhs \leftarrow subproblem.R_c.lhs + \delta_{c,i}$
3: **return** $subproblem$

---

## 5.4 Bounding

In this subsection, we present two implementations of the *obtainLowerBound* function in Line 8 of Algorithm 1. The first one uses the optimal solution of the convex subproblems (Problem 7) as lower bounds. This approach typically results in extensive computation time. The second one solves an LP relaxation of the convex subproblems, called fixed risk relaxation (FRR). FRR dramatically reduces the computation time compared to the first implementation. The NIRA algorithm employs the second implementation.

### 5.4.1 SIMPLE BOUNDING

Algorithm 4 shows the most straightforward way to obtain lower bounds. It simply solves the convex relaxed subproblems (Problem 7) using the methods presented in Section 4.2. The optimal objective value of a relaxed subproblem gives a lower bound of the optimal objective value of all the subproblems below it. For example, the optimal solution of the relaxed subproblem at Node $2'$ in Figure 9-(b) gives a lower bound of the objective value of the subproblems at Nodes 4 and 5. This is because the constraints of the relaxed subproblems are always a subset of the constraints of the subproblems below. Note that optimization problems are formulated as minimizations.

However, despite the simplicity of this approach, its computation time is slow because the algorithm needs to solve a myriad of subproblems. For example, a simple path planning problem with

---

**Algorithm 4** A simple implementation of the obtainLowerBound function in Line 8 of Algorithm 1

---

**function** obtainLowerBound-Naive($subproblem$) **returns** a lower bound

1: Solve $subproblem$ using algorithms presented in Section 4.2
2: **return** the optimal objective value

---

ten time steps and one rectangular obstacle requires the solution of $4^{10} = 1,048,576$ in the worst case, although the branch-and-bound process often significantly reduces the number of subproblems to be solved. Moreover, the subproblems (Problem 7) are *nonlinear* convex optimization problems due to the nonlinearity of $m_{c,i,j}$ and $m_{\mathbb{U},t,i}$ in (62) and (63). A general nonlinear optimization problem requires significantly more solution time than more specific classes of optimization problems, such as linear programmings (LPs) and quadratic programmings (QPs).

### 5.4.2 FIXED RISK RELAXATION

Our new relaxation approach, fixed risk relaxation (FRR), addresses this issue. FRR linearizes the nonlinear constraints (62) and (63) in Problem 7 by fixing all the individual risk allocations, $\delta_{c,i}$ and $\epsilon_{t,i}$, to their upper bound $\Delta$. When the objective function is linear, an FRR is an LP. An FRR with a convex piecewise linear objective function can also be reformulated as an LP by introducing slack variables (See Section 7.1.1 for an example.). A general convex objective function can be approximated by a convex piecewise linear function. Hence, in many cases, the FRRs of subproblems result in LPs, which can be solved very efficiently. The fixed risk relaxation of Problem 7 is as follows:

**Problem 8: Fixed Risk Relaxation of Problem 7**

$$
\begin{aligned}
J_{FRR}^\star = \min_{\bar{\boldsymbol{u}}_{1:N}} \quad & J'(\boldsymbol{u}_{1:N}, \bar{\boldsymbol{x}}_{1:N}) \\
\text{s.t.} \quad & \forall t \in \mathbb{T}^-, \quad \bar{\boldsymbol{x}}_{t+1} = \boldsymbol{A}_t \bar{\boldsymbol{x}}_t + \boldsymbol{B}_t \boldsymbol{u}_t \\
& \bigwedge_{c \in \mathcal{C}} \bigwedge_{i \in \mathcal{I}_c(s)} h_{c,i}^T \bar{\boldsymbol{x}}_{t_i} - g_{c,i} \le -m_{c,i,j(c,i)}(\Delta_c) \quad (65) \\
& \bigwedge_{t \in \mathbb{T}^-} \bigwedge_{i \in \mathcal{I}_\mathbb{U}} \boldsymbol{h}_{\mathbb{U},i} \bar{\boldsymbol{u}}_t - g_{\mathbb{U},i} \le -m_{\mathbb{U},t,i}(\Delta_c) \quad (66)
\end{aligned}
$$

Note that the nonlinear terms in (62) and (63), $m_{c,i,j}$ and $m_{\mathbb{U},t,i}$, become constant by fixing $\delta_{c,i}$ and $\epsilon_{t,i}$ to $\Delta_c$, which is a constant. The optimal objective value of the FRR provides a tightest lower bound among the linear relaxations of constraints (62) and (63). The following lemmas hold:

**Lemma 6.** *Problem 8 gives a lower bound to the optimal objective value of Problem 7:*

$$J_{FRR}^\star \le J_{SP}^\star$$

*Proof.* $m_{c,i,j}(\cdot)$ and $m_{\mathbb{U},t,i}(\cdot)$ are monotonically decreasing functions. Since $\delta_{c,i} \le \Delta_c$ and $\epsilon_{t,i} \le \Delta_c$, all individual chance constraints (65) and (66) of the Fixed Risk Relaxation are less stricter than the first conjunct of (62) and (63). Therefore, the cost of the optimal solution of the Fixed Risk Relaxation is less than or equal to the original subproblem. □

**Lemma 7.** *FRR gives the tightest lower bound among the linear relaxations of constraints (62) and (63).*

*Proof.* The linear relaxation of (62) and (63) becomes tighter by fixing $\delta_{c,i}$ and $\epsilon_{t,i}$ to a lesser value. However, setting $\delta_{c,i}$ and $\epsilon_{t,i}$ to values less than $\Delta_c$ may exclude feasible solutions, such as the one

---

**Algorithm 5** An FRR implementation of the obtainLowerBound function in Line 8 of Algorithm 1

**function** obtainLowreBound-FRR($subproblem$) **returns** lower bound

1: **for** $\forall (c, i, j)$ in $subproblem.C$ **do**
2:    $subproblem.C_{c,i,j}.rhs \leftarrow -m_{c,i,j}(\Delta_c)$    //Apply fixed risk relaxation
3: **end for**
4: **for** $\forall (t, i)$ **do**
5:    $subproblem.D_{t,i}.rhs \leftarrow -m_{\mathbb{U},t,i}$    //Apply fixed risk relaxation
6: **end for**
7: Remove $subproblem.R$
8: Solve $subproblem$ using an LP solver
9: **return** the optimal objective value

---

that sets $\delta_{c,i} = \Delta_c$ for some $(c, i)$. Hence, FRR is the tightest linear relaxation of (62) and (63), resulting in the tightest lower bound. $\qquad\square$

Note that the optimal solution of Fixed Risk Relaxation (Problem 8) is typically an infeasible solution to Problem 7, since setting $\delta_{c,i} = \epsilon_{t,i} = \Delta_c$ violates the constraint (64).

Algorithm 5 implements the fixed risk relaxation. The LP relaxation is solved by an LP solver, and its optimal objective value is returned.

This completes our second spiral, planning for CCQSPs with a fixed schedule and nonconvex constraints. In the next section, we turn to our final spiral, which involves flexible temporal constraints.

## 6. CCQSP Planning with a Flexible Schedule

This section presents the complete p-Sulu Planner, which efficiently solves the general CCQSP planning problem with a *flexible* schedule and a non-convex state space (Problem 2 in Section 3.1.2). The problem is to find a schedule of events $s$ that satisfies simple temporal constraints, as well as a nominal control sequence $\bar{u}_{0:N-1}$ that satisfies the chance constraints and minimizes cost. Our approach is to first generate a feasible schedule and then to extend it to a control sequence for that schedule, while iteratively improving the candidate schedules using branch-and-bound.

We build the p-Sulu Planner upon the NIRA algorithm presented in the previous section. Recall that NIRA optimizes the nominal control sequence $\bar{u}_{0:N-1}$ given a *fixed* schedule $s$. The p-Sulu Planner uses NIRA as a subroutine that takes a schedule $s$ as an input, and outputs the optimal objective value as well as an executable control sequence. We denote the optimal objective value for a given schedule $s$ as $J^\star(s)$. Using this notation, the CCQSP planning problem with a *flexible* schedule (Problem 2) can be rewritten as a schedule optimization problem as follows:

$$\min_{s \in \mathcal{S}_F} J^\star(s). \tag{67}$$

Recall that the domain of feasible schedules $\mathcal{S}_F$ (Definition 11) is a finite set, since we consider a discretized, finite set of time steps $\mathbb{T}$ (see Section 2.1). Hence, the schedule optimization problem (67) is a combinatorial constraint optimization problem, where the constraints are given in the form of simple temporal constraints.

---

**Algorithm 6** The the p-Sulu Planner

---

**function** pSulu($ccqsp$) **returns** optimal schedule and control sequence

1: $Incumbent = \infty$
2: Set up $queue$ as a FILO queue
3: $\mathcal{E}_{\sigma_0} = \{e_0\}$, $\sigma_0(e_0) = 0$    //initialize the partial schedule
4: $queue \leftarrow \langle \mathcal{E}_{\sigma_0}, \sigma_0 \rangle$
5: **while** $queue$ is not empty **do**
6:    $\langle \mathcal{E}_\sigma, \sigma \rangle \leftarrow$ the last entry in $queue$
7:    Remove $\langle \mathcal{E}_\sigma, \sigma \rangle$ from $queue$
8:    $[J^\star, \boldsymbol{u}_{0:N-1}] \leftarrow$ obtainLowerBound($ccqsp, \mathcal{E}_\sigma, \sigma$)
9:    **if** $J^\star < Incumbent$ **then**
10:      **if** $\mathcal{E}_\sigma = \mathcal{E}$ **then**
11:        $Incumbent \leftarrow J^\star$,   $OptCtlSequence \leftarrow \boldsymbol{u}_{0:N-1}$,   $OptSchedule \leftarrow \sigma$
12:      **else**
13:        expand($ccqsp, queue, e, \mathcal{E}_\sigma, \sigma$)
14:      **end if**
15:    **end if**
16: **end while**
17: **return** $OptCtlSequence, OptSchedule$

---

## 6.1 Algorithm Overview

Our solution approach is again to use a branch-and-bound algorithm. In the branch-and-bound search, the p-Sulu Planner incrementally assigns an execution time step to each event in order to find the schedule that minimizes $J^\star(s)$ in (67). The objective function is evaluated by solving the fixed schedule CCQSP planning problem using the NIRA algorithm. Although the combination of the two branch-and-bound searches in the p-Sulu Planner and NIRA are equivalent to one unified branch-and-bound search in practice, we treat them separately for ease of explanation.

As shown in Figure 12, the branch-and-bound algorithm searches for an optimal schedule by incrementally assigning execution time steps to each event in a depth-first manner. Each node of the search tree corresponds to a partial schedule (Definition 2), which assigns execution time steps to a subset of the events included in the CCQSP. The partial schedule at the root node only involves an assignment to the start node $e_0$. The tree is expanded by assigning an execution time step to one new event at a time. For example, the node $\sigma(e_1) = 2$ in Figure 12-(a) represents a partial schedule that assigns the execution time step $t = 0$ to the event $e_0$ and $t = 2$ to $e_1$, while leaving $e_E$ unassigned.

The p-Sulu Planner obtains the lower bound of the objective function value $J^\star(s)$ by solving a CCQSP planning problem with a *partial* schedule that can be extended to $s$. The the p-Sulu Planner minimizes the search space by dynamically pruning the domain through forward checking. More specifically, after an execution time is assigned to an event at each iteration of the branch-and-bound search, the the p-Sulu Planner runs a shortest-path algorithm to tighten the real-valued upper and lower bounds of the execution time step of unassigned events according to the newly assigned execution time step.

Algorithm 6 shows the pseudocode of the algorithm. At each node of the search tree, a fixed-schedule CCQSP planning problem is solved with the given partial schedule. If the node is at the leaf of the tree and the optimal objective value is less than the incumbent, the optimal solution is updated (Line 11). If the node is not at the leaf, the optimal objective value of the corresponding subproblem is a lower bound for the optimal objective value of subsequent nodes. If the lower bound is less than the incumbent, the node is expanded by enumerating the feasible execution time assignments to an unassigned event (Line 13). Otherwise, the node is not expanded, and hence pruned. Details of this branch-and-bound process are described in later subsections.

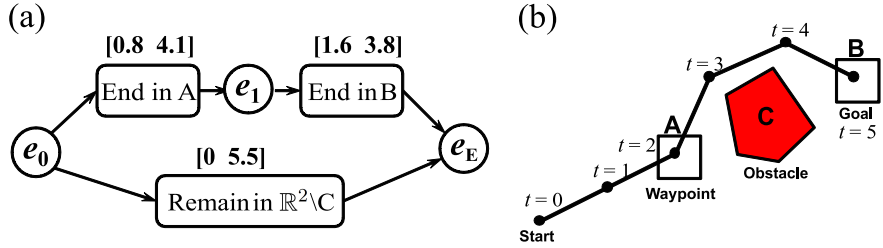

Figure 10: **(a)** An example of CCQSP; **(b)** a plan that satisfies the CCQSP in (a)

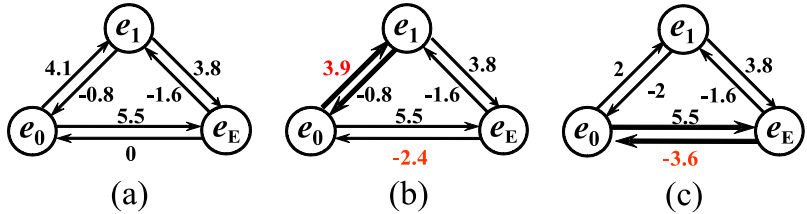

Figure 11: **(a)** The directed distance graph representation of the CCQSP in Figure 10-(a); **(b)** the d-graph of (a), which shows the shortest distances between nodes; **(c)** the updated d-graph after the execution time $t = 2$ is assigned to the event $e_1$.

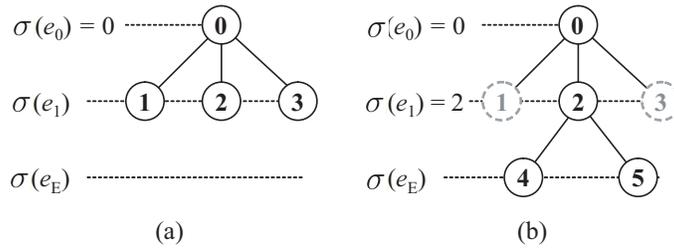

Figure 12: Branch-and-bound search over a schedule $s$. We assume that the time interval is $\Delta T = 1.0$. **(a)** The node $\sigma(e_0) = 0$ is expanded; $D_{e_1}(\sigma) = \{1, 2, 3\}$ given $\sigma(e_0) = 0$, since $\left[d_e^{\max}(\sigma), d_e^{\min}(\sigma)\right] = [0.8, 3.9]$ from Figure 11-(b); **(b)** the node $\sigma(e_1) = 2$ is expanded; $D_{e_E}(\sigma) = 4, 5$ given $\sigma(e_0) = 0$ and $\sigma(e_1) = 2$, since $\left[d_e^{\max}(\sigma), d_e^{\min}(\sigma)\right] = [3.6, 5.5]$ from Figure 11-(c).

**Walk-through example** We present a walk-through example to give readers insight into the solution process. We consider a CCQSP shown in Figure 10-(a). The CCQSP specifies a mission to go through a waypoint A and get to the goal region B while avoiding the obstacle C, as shown in Figure 10-(b). We assume that the time interval is $\Delta T = 1.0$.

Figures 11 and 12 illustrate the solution process. The the p-Sulu Planner algorithm is initialized by assigning the execution time 0 to the start event $e_0$. Figure 11-(a) is the distance graph representation of the simple temporal constraints (Dechter, 2003) of the CCQSP. Note that a simple chance constraint is equivalently represented as a pair of inequality constraints as follows:

$$s(e) - s(e') \in [l, u] \iff s(e) - s(e') \leq u \ \wedge \ s(e') - s(e) \leq -l.$$

The two inequality constraints are represented by two directional edges between each two nodes in the distance graph. The the p-Sulu Planner runs an all-pair shortest-path algorithm on the distance graph to obtain the *d-graph* shown in Figure 11-(b). A d-graph is a completed distance graph where each edge is labeled by the shortest-path length. The d-graph represents the tightest temporal constraints. Then the algorithm enumerates the feasible execution-time assignments for the event $e_1$ using the d-graph. According to the d-graph, the execution time for the event $e_1$ must be between 0.8 and 3.9. Since we consider discrete time steps with the time interval $\Delta T = 1.0$, the feasible execution time steps for $e_1$ are $\{1, 2, 3\}$. The idea behind enumerating all feasible execution time steps is to assign an event, and thus to tighten the bounds of all unassigned events in order to ensure feasibility.

At the node $\sigma(e_1) = 1$, the the p-Sulu Planner solves the FRR of the fixed-schedule CCQSP planning problem *only* with the "End in A" episode and the execution schedule $\sigma(e_1) = 1$. In other words, it tries to find the optimal path that goes through A at $t = 1$, but neglects the goal B and obstacle C. If a solution exists, its optimal cost gives a lower bound on the objective value of all feasible paths that go through A at $t = 1$. Assume here that such a solution does not exist. Then, the the p-Sulu Planner prunes the node $\sigma(e_1) = 1$, and goes to the next node $\sigma(e_1) = 2$. It solves the FRR of the corresponding fixed-schedule subproblem to find the best path that goes through A at $t = 2$. Assume that the the p-Sulu Planner finds a solution. Then, the the p-Sulu Planner expands the node in the following process. First, it fixes the execution time $\sigma(e_1) = 2$ in the d-graph, and runs a shortest-path algorithm in order to tighten the temporal constraints (11-(c)). Then the the p-Sulu Planner uses the updated d-graph to enumerate the feasible execution-time assignments for the event $e_E$, which are $\{4, 5\}$. It visits both nodes and solves the fixed-schedule subproblems exactly with all episodes and a fully assigned schedule. For example, at the node $\sigma(e_E) = 5$, it computes the best path that goes through A at $t = 2$ and reaches B at $t = 5$ while avoiding the obstacle C, as shown in Figure 10-(b). Assume that the optimal objective values of the subproblems are 10.0 for $\sigma(e_E) = 4$ and 8.0 for $\sigma(e_E) = 5$. The algorithm records the solution with $\sigma(e_E) = 5$ and its cost 8.0 as the incumbent.

The algorithm then backs up and visits the node $\sigma(e_1) = 3$, where a relaxed subproblem with only the "End in A" episode is solved to obtain the lower bound of the objective value of subsequent nodes. The lower bound turns out to be 9.0, which exceeds the incumbent. Therefore, the branch is pruned. Since there are no more nodes to expand, the algorithm is terminated, and the incumbent solution is returned.

---

**Algorithm 7** Implementation of expand function in Line 13 of Algorithm 6

---

**function** expand($ccqsp, queue, e, \mathcal{E}_\sigma, \sigma$)

1: Fix the distance between $e_0$ and $e$ to $\sigma(e)\Delta T$ on the d-graph of $ccqsp$
2: Update the d-graph by running a shortest-path algorithm
3: Choose $e'$ from $\mathcal{E}\backslash\mathcal{E}_\sigma$   //choose an unassigned event
4: $\mathcal{E}_{\sigma'} := \mathcal{E}_\sigma \cup e'$
5: $D_{e'}(\sigma) := \{\, t \in \mathbb{T} \mid d_{e'}^{\min}(\sigma) \le t\Delta T \le d_{e'}^{\max}(\sigma)\,\}$
6: **for** $t$ in $D_{e'}(\sigma)$ **do**
7: $\quad \sigma'(e) := \begin{cases} \sigma(e) & (e \in \mathcal{E}_\sigma) \\ t & (e = e') \end{cases}$   //update the partial schedule
8: $\quad queue \leftarrow \langle \mathcal{E}_{\sigma'}, \sigma' \rangle$
9: **end for**

---

## 6.2 Branching

Algorithm 7 outlines the implementation of the expand() function in Algorithm 6. It takes a partial schedule $\sigma$ as an input, and adds to the queue a set of schedules that assign an execution time step to an additional event $e'$. In other words, the domain of the newly added schedules $\mathcal{E}_{\sigma'}$ has one more assigned event than the domain of the input partial schedule $\mathcal{E}_\sigma$. The details of Algorithm 7 are explained in the following parts of this subsection.

### 6.2.1 ENUMERATION OF FEASIBLE TIME STEP ASSIGNMENTS USING D-GRAPH

When enumerating all feasible time steps, the simple temporal constraints must be respected. To accomplish this, we use a d-graph to translate the bounds on the durations between two events into the bounds on the execution time step of each event. It is shown by Dechter et al. (1991) that the set of feasible execution times for an event $e$ is bounded by the distance between $e$ and $e_0$ on the d-graph. A d-graph is a directed graph, where the weights of the edges represent the *shortest* distances between nodes, as in Figure 11-(b). In order to obtain the d-graph representation, we first translate the simple temporal constraints into a directed distance graph, as in Figure 11-(a). The weight of an edge between two nodes (events) corresponds to the maximum duration of time from the origin node to the destination node, as specified by the corresponding simple temporal constraint. The distance takes a negative value to represent lower bounds. The d-graph (Figure 11-(b)) is obtained from the distance graph (Figure 11-(a)) by running an all-pair shortest-path algorithm (Dechter et al., 1991).

**Forward checking over a d-graph**  The the p-Sulu Planner algorithm incrementally assigns an execution time step to each event, as explained in the walk-through example. The p-Sulu Planner minimizes the search space through forward checking using the d-graph. As in forward checking methods of Constraint Programming, our method prunes all values of unassigned variables (i.e., execution times of an unassigned event) that violate simple temporal constraints. What is different here from normal forward checking is that no back tracking is performed, due to decomposability of d-graph. The forward checking is conducted in the following process. Once an execution time step $t$ is assigned to an event $e$ (i.e., $\sigma(e) = t$), the distance from $e_0$ to $e$ is fixed to $t\Delta T$, and the distance from $e$ to $e_0$ is fixed to $-t\Delta T$ on the distance graph (Line 1 of Algorithm 7). Recall that $t$ is an index of discretized time steps with a fixed interval $\Delta T$, while the temporal bounds are given as real-valued times (Section 2.1). We then run a shortest-path algorithm to update the d-graph (Line 2). Given a partial schedule $\sigma$, we denote the updated shortest distance from the start event $e_0$ to $e'$ on the d-graph by $d_{e'}^{\max}(\sigma)$, and the distance from $e'$ to $e_0$ by $d_{e'}^{\min}(\sigma)$.

For example, the execution time 2 is assigned to the event $e_1$ in Figure 11-(c) (i.e., $\sigma(e_1) = 2$), so the distance between $e_0$ and $e_1$ is fixed to 2 and the distance in the opposite direction is fixed to $-2$. Then we run a shortest-path algorithm again to update the d-graph. As a result, we obtain updated distances $d_{e_E}^{\max}(\sigma) = 5.5$ and $d_{e_E}^{\min}(\sigma) = -3.6$.

Dechter et al. (1991) showed that $d_{e'}^{\max}(\sigma)$ corresponds to the upper bound of the feasible execution time for an unassigned event $e'$, while $d_{e_E}^{\min}(\sigma)$ corresponds to the negative of the lower bound. Hence, after a partial schedule $\sigma$ is assigned to events $e \in \mathcal{E}_\sigma$, the updated domain for an unassigned event $e' \notin \mathcal{E}_\sigma$ is bounded by $d_{e'}^{\min}(\sigma)$ and $d_{e'}^{\max}(\sigma)$. Note that the domain of the execution time steps $e'$ is included in, but not equal to $[d_{e'}^{\min}(\sigma), d_{e'}^{\max}(\sigma)]$, because we only consider discrete execution time steps in a finite set $\mathbb{T}$. During the forward checking, the p-Sulu Planner only computes the real-valued bounds $[d_{e'}^{\min}(\sigma), d_{e'}^{\max}(\sigma)]$. The feasible values of an unassigned variable $e'$ are not enumerated until the search tree is expanded to $e'$.

**Enumerating the domain of execution time steps for an unassigned event** We can readily extract the feasible execution time steps for any unassigned event $e' \notin \mathcal{E}_\sigma$ from the updated d-graph with a partial schedule $\sigma$. Let $D_{e'}(\sigma)$ be the domain of execution time steps for an unassigned event $e' \notin \mathcal{E}_\sigma$, given a partial schedule $\sigma$. The finite domain $D_{e'}(\sigma)$ is obtained as follows:

$$D_{e'}(\sigma) := \{\, t \in \mathbb{T} \mid d_{e'}^{\min}(\sigma) \leq t\Delta T \leq d_{e'}^{\max}(\sigma)\}.$$

Note that $D_e(\sigma)$ may be empty when the temporal constraints are tight, even though they are feasible. The user of the p-Sulu Planner must make $\Delta T$ small enough so that $D_e$ is not empty.

For example, Figure 11-(b) is the d-graph given the partial schedule $\{\sigma(e_0) = 0\}$. According to the d-graph, $e_1$ must be executed between 0.8 and 3.9. Assuming that $\Delta T = 1$, the set of feasible execution time steps for $e_1$ is $D_{e_1}(\sigma) = \{1, 2, 3\}$, as shown in Figure 12-(a). Likewise, Figure 11-(c) is the d-graph given the partial schedule $\{\sigma(e_0) = 0, \sigma(e_1) = 2\}$; the feasible execution time of $e_E$ is between 3.6 and 5.5. Hence, the set of feasible execution time steps for $e_E$ is $D_{e_E}(\sigma) = \{4, 5\}$, as shown in Figure 12-(b).

The enumeration is conducted in Line 6 in Algorithm 7. Then the algorithm creates extensions of the input partial schedule by assigning each of the time steps to $e'$ (Line 7), and puts the extended partial schedules in the queue (Line 8).

### 6.2.2 EFFICIENT VARIABLE ORDERING OF BRANCH-AND-BOUND SEARCH

When choosing the next event to assign a time step in Line 3 of Algorithm 7, two variable ordering heuristics are found to be effective in order to reduce computation time.

The first heuristic is our new *convex-episode-first* (CEF) heuristic, which prioritizes events that are not associated with non-convex constraints. The idea of the CEF heuristic is based on the observation that subproblems of the branch-and-bound algorithm are particularly difficult to solve when the episodes in $\mathcal{A}(\mathcal{E}_\sigma)$ involve non-convex state constraints. The "Remain in $\mathbb{R}^2 \backslash C$" (2D plane minus the obstacle C) episode in the walk-through example in Figures 10 is an example of such non-convex episodes. Therefore, an effective approach to reduce the computation time of the p-Sulu Planner is to minimize the number of non-convex subproblems solved in the branch-and-bound process. This idea can be realized by sorting the events so that the episodes with a convex feasible region are always examined in the branch-and-bound process before the episodes with a non-convex feasible region. In the walk-through example, note that we visited the event $e_1$ before the event $e_E$ in this example. This is because the "End in A" episode only involves a convex state constraint while "Remain in $\mathbb{R}^2 \backslash C$" (2D plane minus the obstacle C) is non-convex.

The second one is the well-known most constrained variable heuristic. When the p-Sulu Planner expands a node, it counts the number of feasible time steps of all unassigned events, and chooses the one with the least number of feasible time steps. The second heuristic used to break ties in the first heuristic.

### 6.3 Bounding

We next present the implementation of the obtainLowerBound() function in Line 8 of Algorithm 6. The algorithm obtains the lower bound by solving a relaxed CCQSP planning problem with a fixed partial schedule.

Algorithm 8 outlines the implementation of the obtainLowerBound() function. It takes a partial schedule $\sigma$ as an input, and outputs the lower bound of the objective function, as well as the optimal control sequence, given the partial schedule $\sigma$. It constructs a relaxed optimization problem, which only involves episodes whose start and end events are both assigned execution time steps (Line 1). If the optimization problem involves non-convex constraints, the NIRA algorithm is used to obtain the solution to the problem (Line 3). Otherwise we solve the FRR of the convex optimization problem to obtain the lower bound efficiently (Line 5). If the input is a fully assigned schedule ($\mathcal{E}_\sigma = \mathcal{E}$), the corresponding node is a leaf node. In such case we obtain an *exact* solution to the CCQSP planning problem with the fixed schedule $\sigma$ by running the NIRA algorithm (Line 3). The details of Algorithm 8 are explained in the subsequent part of this subsection.

---

**Algorithm 8** Implementation of obtainLowerBound function in Line 8 of Algorithm 6

**function** obtainLowerBound($ccqsp, \mathcal{E}_\sigma, \sigma$) **returns** optimal objective value and control sequence

1: $subprblem \leftarrow$ Problem 9 with $\sigma$ given $ccqsp$
2: **if** $\mathcal{E}_\sigma = \mathcal{E}$ **or** $\mathcal{A}(\sigma)$ has episodes with non-convex state regions, **then**
3: $\quad [J^\star, \boldsymbol{u}_{0:N-1}] \leftarrow$ NIRA($subprblem$)    //Algorithm 1
4: **else**
5: $\quad J^\star \leftarrow$ obtainLowreBound-FRR($subprblem$)    //Algorithm 5
6: $\quad \boldsymbol{u}_{0:N-1} \leftarrow \Phi$
7: **end if**
8: **return** $[J^\star, \boldsymbol{u}_{0:N-1}]$

---

### 6.3.1 RELAXED OPTIMIZATION PROBLEM WITH PARTIAL SCHEDULE

We consider a relaxed optimization problem as follows:

**Problem 9: Relaxed Optimization Problem for a Partial Schedule $\sigma$**

$$J^\star(\sigma) = \min_{\boldsymbol{u}_{0:N-1} \in \mathbb{U}^N} \quad J(\boldsymbol{u}_{0:N-1}, \bar{\boldsymbol{x}}_{1:N}, \sigma) \tag{68}$$

$$\text{s.t.} \quad \forall t \in \mathbb{T}^-, \quad \boldsymbol{x}_{t+1} = \boldsymbol{A}_t \bar{\boldsymbol{x}}_t + \boldsymbol{B}_t \boldsymbol{u}_t \tag{69}$$

$$\bigwedge_{c \in \mathcal{C}} \bigwedge_{a \in (\Psi_c \cap \mathcal{A}(\sigma))} \bigwedge_{t \in \Pi_a(\sigma)} \bigwedge_{k \in \mathcal{K}_a} \bigvee_{j \in \mathcal{J}_{a,k}} \boldsymbol{h}_{c,a,k,j}^T \boldsymbol{x}_t - g_{c,a,k,j} \leq -m_{c,a,k,j}(\delta_{c,a,k}) \tag{70}$$

$$\sum_{k \in \mathcal{K}_a, a \in (\Psi_c \cap \mathcal{A}(\sigma))} \delta_{c,a,k} \geq 1 - \Delta_c, \tag{71}$$

where $J^\star(\sigma)$ is the optimal objective value of the relaxed subproblem with a partial schedule $\sigma$. Recall that $\mathcal{A}(\sigma)$ is the partial episode set of $\sigma$, which only involves the episodes whose start and end nodes are both assigned execution time steps by the partial schedule $\sigma$ (Definition 9). For notational simplicity, we merge the three conjunctions of (70) and obtain the following:

$$\bigwedge_{c \in \mathcal{C}} \bigwedge_{i \in \mathcal{I}_c(\sigma)} \bigvee_{j \in \mathcal{J}_{c,i}} \boldsymbol{h}_{c,i,j}^T \bar{\boldsymbol{x}}_{t_i} - g_{c,i,j} \leq -m_{c,i,j}(\delta_{c,i}).$$

Note that this chance constraint is exactly the same as (57), except that a partial schedule $\sigma$ is specified instead of a fully assigned schedule $s$. Hence, Problem 9 is an instance of a non-convex CCQSP planning problem with a fixed schedule (Problem 6), and can be optimally solved by the NIRA algorithm. Also note that $\sigma$ is a fully assigned schedule at the leaf node of the branch-and-bound search tree.

The optimal objective value of Problem 9 gives a lower bound of the optimal objective value of all the subsequent subproblems in the branch-and-bound tree. This property is formally stated in Lemma 8 below. In order to prove this feature, we first define the concept of an *extension* of a partial schedule as follows:

**Definition 14.** *A schedule $s : \mathcal{E} \mapsto \mathbb{T}$ is an **extension** of a partial schedule $\sigma : \mathcal{E}_\sigma \mapsto \mathbb{T}$ if and only if both assign the same time steps to all the events in the domain of $\sigma$:*

$$\sigma(e) = s(e) \quad \forall e \in \mathcal{E}_\sigma.$$

For example, in Figure 12-(b), a fully assigned schedule $\{s(e_0) = 0, s(e_1) = 2, s(e_E) = 4\}$ and $\{s(e_0) = 0, s(e_1) = 2, s(e_E) = 5\}$ is an extension of a partial schedule $\{\sigma(e_0) = 0, \sigma(e_1) = 2\}$.

The following lemma always holds:

**Lemma 8.** *If a schedule $s$ is an extension of a partial schedule $\sigma$, then the optimal objective value of Problem 9 with $\sigma$ is a lower bound of the optimal objective value with $s$:*

$$J^\star(\sigma) \leq J^\star(s).$$

*Proof.* Since $\sigma$ is a partial schedule, $\mathcal{E}_\sigma \subset \mathcal{E}$, and hence $\mathcal{A}(\sigma) \subseteq \mathcal{A}$. Also, since $\sigma(e) = s(e)$ for all $e \in \mathcal{E}_\sigma$, all the state constraints in the chance constraint (70) of Problem 9 with a partial schedule $\sigma$ are included in the problem with a full schedule $s$. This means that the feasible state space of the problem with $s$ is a subset of the one with $\sigma$. Hence, if the chance constraint (24) of the problem with $s$ is satisfied, the chance constraint (70) of the problem with $\sigma$ is also satisfied. Therefore, the problem with $\sigma$ always results in a better (less) or equal cost than the problem with $\sigma'$, because the former has looser constraints. □

For example, in Figure 12-(b), $e_1$ has been assigned an execution time step but $e_E$ has not. Therefore, at node $\sigma(e_1) = 2$, the chance-constrained optimization problem with only the "End in A" episode is solved with the partial schedule $\{\sigma(e_0) = 0, \sigma(e_1) = 2\}$ (see Figure 10-(a)). It gives a lower bound of the cost of the problems with the fully assigned schedules $\{s(e_0) = 0, s(e_1) = 2, s(e_E) = 4\}$ and $\{s(e_0) = 0, s(e_1) = 2, s(e_E) = 5\}$.

Algorithm 8 obtains a lower bound by solving Problem 9 exactly using the NIRA algorithm, if it involves episodes with non-convex state regions (Line 3). If the function is called on a leaf node, Problem 9 is also solved exactly by NIRA. This is because the solutions of leaf subproblems are candidate solutions of an optimal solution of the overall problem. Hence, by solving them exactly, we can ensure the optimality of the branch-and-bound search.

### 6.3.2 FURTHER BOUNDING WITH FRR

If the relaxed subproblem (Problem 9) is convex, then the p-Sulu Planner solves the FRR of the subproblem, instead of solving it exactly with NIRA, in order to obtain a lower bound more efficiently (Line 5 of Algorithm 8). Many practical CCQSP execution problems have only one episode that has a non-convex feasible region. For example, in the CCQSP planning problem shown in Figures 2 and 3, only the "safe region" ($\mathbb{R}^2$ minus the obstacles) is non-convex, while "Provincetown" (start region), "Scenic region," and "Bedford" (goal region) are convex. In such a case subproblems are solved exactly only at the leaf nodes, and their lower bounds are always evaluated by approximate solutions of FRRs of the subproblems at the non-leaf nodes.

## 7. Results

In this section we empirically demonstrate that the p-Sulu Planner can efficiently operate various systems within the given risk bound. We first present the simulation settings in Section 7.1. Section 7.2 presents the simulation results of the NIRA algorithm, and validates our claim that it can efficiently compute a feasible and near-optimal solution. Section 7.3 demonstrates the p-Sulu Planner on two different benchmark problems. The simulation results highlight the p-Sulu Planner's capability to operate within the user-specified risk bound. Section 7.4 deploys the p-Sulu Planner on the PTS scenarios, while Section 7.5 applies the p-Sulu Planner to the space rendezvous of an autonomous cargo spacecraft to the International Space Station.

### 7.1 Simulation Settings

Recall that, as we stated in Section 2.4, the p-Sulu Planner takes four inputs: a stochastic plant model $\mathcal{M}$, an an initial condition $\mathcal{I}$, a CCQSP $P$, and an objective function $J$. This section specifies $\mathcal{M}$ and $J$, which are commonly used by all the problems in Sections 7.2-7.4. We specify $P$ and $\mathcal{I}$ for each problem in the corresponding section.

### 7.1.1 Stochastic Plant Model

This section explains the plant model used in Sections 7.2 - 7.4. Section 7.5 uses a different plant model that is described in detail in Section 7.5.2. We consider a point-mass double-integrator plant, as shown in (72)-(73). Parameters, such as $u_{max}$, $v_{max}$, $\sigma^2$, and $\Delta T$ are set individually for each problem. This plant model is commonly assumed in literatures on unmanned aerial vehicle (UAV) path planning (Kuwata & How, 2011; Léauté, 2005; Wang, Yadav, & Balakrishnan, 2007).

Our state vector $\boldsymbol{x}_t$ consists of positions and velocities in $x$ and $y$ directions, while the control vector consists of the accelerations:

$$\boldsymbol{x}_t := [x \; y \; v_x \; v_y]^T, \quad \boldsymbol{u}_t := [a_x \; a_y]^T.$$

The plant model is specified by the following matrices:

$$\boldsymbol{A} = \begin{pmatrix} 1 & 0 & \Delta t & 0 \\ 0 & 1 & 0 & \Delta t \\ 0 & 0 & 1 & 0 \\ 0 & 0 & 0 & 1 \end{pmatrix}, \; \boldsymbol{B} = \begin{pmatrix} \Delta t^2/2 & 0 \\ 0 & \Delta t^2/2 \\ \Delta t & 0 \\ 0 & \Delta t \end{pmatrix}, \; \boldsymbol{\Sigma}_w = \begin{pmatrix} \sigma^2 & 0 & 0 & 0 \\ 0 & \sigma^2 & 0 & 0 \\ 0 & 0 & 0 & 0 \\ 0 & 0 & 0 & 0 \end{pmatrix} \quad (72)$$

$$\forall t \in \mathbb{T}, \quad ||\boldsymbol{u_t}|| \leq u_{\max}, \quad ||C\boldsymbol{x_t}|| \leq v_{\max}, \quad (73)$$

where

$$\boldsymbol{C} = \begin{pmatrix} 0 & 0 & 1 & 0 \\ 0 & 0 & 0 & 1 \end{pmatrix}.$$

The first constraint in (73) is imposed in order to limit the acceleration. This nonlinear constraint is approximated by the following set of linear constraints:

$$\forall t \in \mathbb{T}, \quad \boldsymbol{r}_n \cdot \boldsymbol{u}_t \leq u_{\max} \quad (n = 1, 2, \cdots, N_r)$$
$$\boldsymbol{r}_n = \left[ \cos \frac{2\pi n}{N_r}, \sin \frac{2\pi n}{N_r} \right]$$

We choose $N_r = 16$. The second constraint in (73) is imposed in order to limit the velocity. We use the same linear approximation as above.

### 7.1.2 Objective Function

In Sections 7.2.3, 7.3, and 7.4, the cost function is the Manhattan norm of the control input over the planning horizon, as follows:

$$J(\bar{\boldsymbol{x}}_{t_i}, \boldsymbol{U}, s) = \sum_{t=1}^{T} \left( |u_{x,t}| + |u_{y,t}| \right).$$

This cost function represents the total change in momentum, which is roughly proportional to the fuel consumption of an aerial vehicle. Note that a minimization problem with the piece-wise linear cost function above can be equivalently replaced by the following minimization problem with a linear cost function and additional linear constraints by introducing slack variables $\mu_{x,t}$ and $\mu_{y,t}$:

$$\min \quad \sum_{t=1}^{T} (\mu_{x,t} + \mu_{y,t})$$
$$\text{s.t.} \quad \forall t \in \mathbb{T}, \quad \mu_{x,t} \geq u_{x,t} \; \wedge \; \mu_{x,t} \geq -u_{x,t} \; \wedge \; \mu_{y,t} \geq u_{y,t} \; \wedge \; \mu_{y,t} \geq -u_{y,t}$$

In Section 7.2.4, we minimize expected quadratic cost as follows:

$$J(\bar{\boldsymbol{x}}_{t_i}, \boldsymbol{U}, s) = \sum_{t=1}^{T} \mathbb{E}\left[u_{x,t}^2 + u_{y,t}^2\right]. \tag{74}$$

### 7.1.3 COMPUTING ENVIRONMENT

All simulations except for the ones in Section 7.2 are conducted on a machine with a dual-core Intel Xeon CPU clocked at 2.40 GHz, and with 16 GB of RAM. The algorithms are implemented in C/C++, and run on Debian 5.0.8 OS. The simulations in Section 7.2 are conducted on a machine with a quad-core Intel Core i7 CPU clocked at 2.67 GHz, and with 8 GB of RAM. The algorithms are implemented in Matlab, and run on Windows 7 OS. We used IBM ILOG CPLEX Optimization Solver Academic Edition version 12.2 as the linear program solver, and SNOPT version 7.2-9 as the convex optimization solver.

## 7.2 NIRA Simulation Results

We first statistically compare the performance of NIRA with the prior art. Recall that NIRA is a solver for CCQSP planning problems with non-convex state constraints and a fixed schedule (Problem 3), and used as a subroutine in the p-Sulu Planner.

### 7.2.1 COMPARED ALGORITHMS

There are two existing algorithms that can solve the same problem:

1. **Fixed risk allocation** (Blackmore et al., 2006) - This approach *fixes* the risk allocation to a uniform value. As a result, with an assumption that the cost function is linear, Problem 6 can be reformulated to a mixed-integer linear programming (MILP) problem, which can be solved efficiently by a MILP solver, such as CPLEX.

2. **Particle Control** (Blackmore, 2006) - Particle Control is a sampling-based method, which uses a finite number of samples to approximate the joint chance constraints. The control sequence is optimized so that the number of samples that violate constraints is less than $\Delta_c N_p$, where $N_p$ is the total number of samples. The optimization problem is again reformulated into MILP, with an assumption that the cost function is linear.

We also compare NIRA with an MDP in Section 7.2.5. Although an MDP does not solve exactly the same problem as NIRA, it can also avoid risk by considering a penalty cost of constraint violations. The purpose of the comparison is to highlight the capabilities of chance-constrained planning to provide a guarantee on the probability of failure.

### 7.2.2 PROBLEM SETTINGS

We compare closed-loop and open-loop NIRAs with the two algorithms on a 2-D path planning problem with a randomized location of an obstacle, as shown in Figure 13. A vehicle starts from $[0,0]$ and heads to the goal at $[1.0, 1.0]$, while avoiding a rectangular obstacle. The obstacle with edge length 0.6 is placed at a random location within the square region with its corners at $[0,0]$, $[1,0]$, $[1,1]$, and $[0,1]$. We consider ten time steps with the time interval $\Delta t = 1.0$. We require that the mean state at $t = 10$ is at $[1.0, 1.0]$. The risk bound is set to $\Delta = 0.01$. We set the standard deviation of the disturbance as $\sigma = 0.01$. We use the expected quadratic cost function given in (74). The steady-state LQR gain is used for the closed-loop NIRA with $\mathbf{Q} = I_4$ and $\mathbf{R} = 10000 I_2$, where $I_n$ is the $n \times n$ identity matrix and $\mathbf{Q}$ and $\mathbf{R}$ are cost matrices for the state and control variables, respectively.

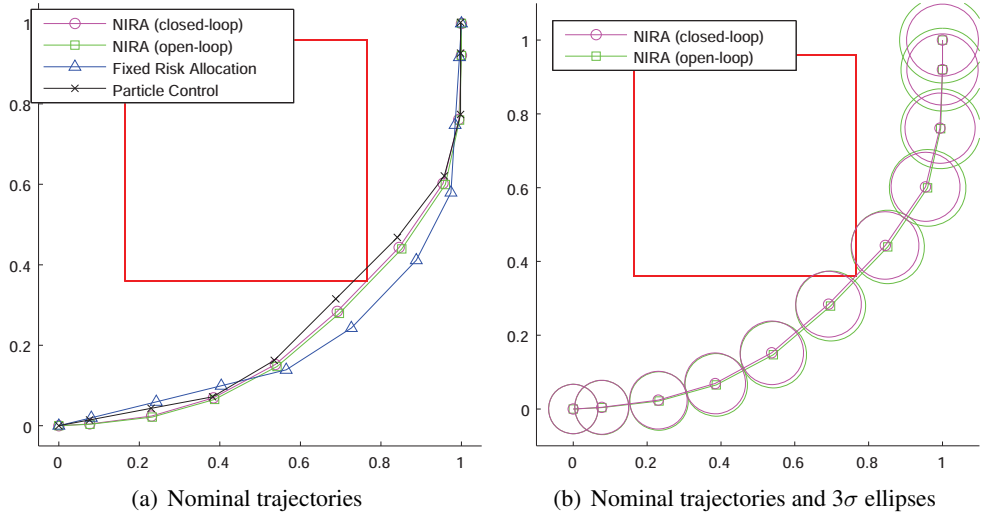

(a) Nominal trajectories        (b) Nominal trajectories and $3\sigma$ ellipses

Figure 13: (a) An instance of the 2-D path planning problem used in 7.2.3. The obstacle with a fixed size is randomly placed within the unit square for each run. (b) The mean and the standard deviation of the closed-loop and open-loop NIRAs.

### 7.2.3 Performance Comparison

Recall that the solution of the NIRA algorithm, which is used by the p-Sulu Planner to solve sub-problems, is not an exactly optimal solution to Problem 3, since risk allocation (Section 4.1.1) and risk selection (Section 5.1.1) replace the chance constraint (29) with its sufficient condition (57) $\wedge$ (59). Since the chance constraint (29) is very difficult to evaluate, all the previously proposed methods solve optimization with its approximation. We provide empirical evidence that our risk allocation/selection approach results in a solution that is significantly closer to the optimal solution than the prior art, while satisfaction of the original constraint (29) is guaranteed.

We evaluate the suboptimality of the solutions by the difference between the risk bound, $\Delta = 0.001$, and the resulting probability of constraint violation, $P_{fail}$, estimated by a Monte-Carlo simulation. $1 - P_{fail}$ is equal to the left-hand-side value of (29) in Problem 3. Hence, the chance constraint (29) is equivalent to:

$$P_{fail} \leq \Delta.$$

The strictly optimal solution to this problem should achieve $P_{fail} = \Delta$, although such an exact solution is unavailable, since there is no algorithm to solve Problem 3 exactly. A solution is sub-optimal if $P_{fail} < \Delta$, and their ratio $\Delta/P_{fail}$ represents the degree of suboptimality. A solution violates the chance constraint if $P_{fail} > \Delta$.

| Algorithm | Computation time [sec] | Probability of failure | Cost |
|---|---|---|---|
| **NIRA (Closed-loop)** | $54.8 \pm 36.9$ | $0.0096 \pm 0.0008$ | $0.666 \pm 0.061$ |
| **NIRA (Open-loop)** | $25.0 \pm 13.1$ | $0.0095 \pm 0.0008$ | $0.672 \pm 0.068$ |
| Fixed Risk Allocation | $0.42 \pm 0.04$ | $(2.19 \pm 0.40) \times 10^{-4}$ | $0.726 \pm 0.113$ |
| Particle Control (100 particles) | $41.7 \pm 12.8$ | $0.124 \pm 0.036$ | $0.635 \pm 0.048$ |

Table 1: The averages and the standard deviations of the computation time, the probability of constraint violation, and the cost of the four algorithms. Each algorithms are run 100 times with random location of an obstacle. The risk bound is set to $\Delta = 0.01$. Note that Particle Control results in less cost than the other two methods because its solutions violate the chance constraint.

Table 1 compares the performance of the four algorithms. The values in the table are the averages and the standard deviations of 100 runs with random locations for the obstacle. The probability of constraint violation, $P_{fail}$, is evaluated by Monte-Carlo simulations with $10^6$ samples.

**Comparison of closed-loop and open-loop NIRAs**   Before comparing NIRA with existing algorithms, we first compare the two variants of NIRA: the closed-loop and open loop NIRAs. Table 1 shows that the closed-loop NIRA results in less cost than the open-loop NIRA. Importantly, the former outperforms the latter in *all* the 100 test cases. This reduction in cost by the closed-loop approach is explained by Figure 13-(b), which shows the $3\sigma$ ellipses of the probability distribution of the state. Since the closed-loop NIRA assumes a feedback control, the future position is less uncertain. As a result, the plan generated by the closed-loop NIRA is less conservative. In fact, Table 1 shows that $P_{fail}$ of the closed-loop NIRA is closer to the risk bound than that of the open-loop NIRA. However, the closed-loop planning problem requires about twice as much solution time as the open-loop one since it is more complicated due to additional chance constraints on control input.

**Comparison with the fixed risk allocation approach**   Table 1 shows that closed and open NIRAs result in the average probabilities of failure 0.0096 and 0.0095 respectively, which is within the user-specified risk bound $\Delta = 0.01$. On the other hand, the fixed risk allocation approach results in a very conservative probability of failure, $P_{fail} = 0.000219$, which is 98% smaller than $\Delta$. This result indicates that the solution by NIRA is significantly closer to the exactly optimal solution than the fixed risk allocation approach. In fact, the NIRA algorithm results in less cost than the fixed risk allocation approach in all the 100 runs. This is because it optimizes the risk allocation while the fixed risk allocation approach uses the predetermined risk allocation.

Figure 14 shows the suboptimality measure $\Delta/P_{fail}$ of the open-loop NIRA with different settings of the risk bound $\Delta$. For all values of $\Delta$, the suboptimality of NIRA is significantly smaller than the fixed risk allocation approach. The graph shows a tendency that the suboptimality of NIRA gets smaller for less $\Delta$, while the suboptimality of the fixed risk allocation approach is approximately constant.

NIRA achieves the improvement in solution optimality with a cost of computation time; Table 1 shows that NIRA takes longer computation time than the risk allocation approach by the factor of two. Hence, NIRA and the fixed risk allocation approach provide users with a trade-off between suboptimality and computation time.

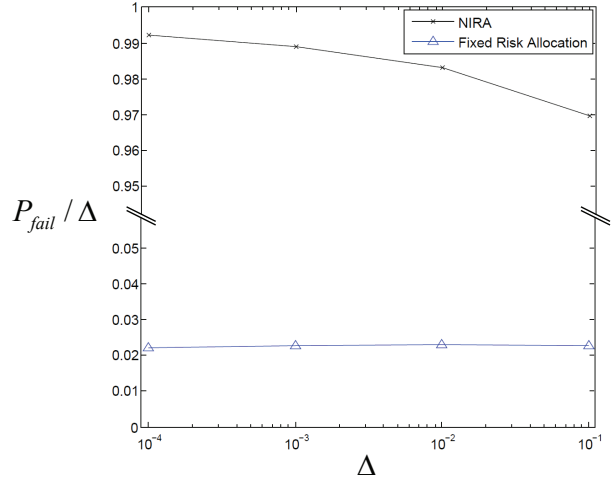

Figure 14: Suboptimality of NIRA and the fixed risk allocation approach. Strictly optimal solution has $\Delta/P_{fail} = 1$. A smaller value of $\Delta/P_{fail}$ indicates that the solution is suboptimal.

**Comparison with the Particle Control**   Table 1 shows that the average probability of failure of the Particle Control approach is higher than the risk bound $\Delta = 0.01$, meaning that the approach tends to generate infeasible solutions. On the other hand, NIRA guarantees the satisfaction of the chance constraint since it employs a conservative approximation of the joint chance constraint.

Particle Control has a guarantee that its solution converges to an optimal solution when increasing the number of samples to infinity. However, using a large number of samples is impractical, since computation time and memory usage grow exponentially as the number of samples increases. For example, we used only 100 samples in the analysis in Table 1. When using 300 samples, it took 4596 seconds (about 1.5 hours) to solve the same problem with the obstacle's centered at $[0.5, 0.5]$. Computation with 1000 samples could not be conducted, because of the shortage of memory. On the other hand, the computation time of NIRA is significantly shorter than PC, while guaranteeing the feasibility of the solution.

### 7.2.4 OPTIMAL PLANNING WITH EXPECTED COST

Next we demonstrate the capability of the p-Sulu Planner to handle expected cost, instead of the cost of the expected trajectory, for the same path planning problem presented above. Specifically, we consider the expected quadratic cost function shown in (74). When conducting open-loop planning, this cost function can be transformed to a function of nominal control inputs with a constant term by using the equality (15). However, when performing closed-loop planning, this equality is not exact, due to controller saturation. Nevertheless, we use (15) as an approximation of the expected cost, as explained in Section 2.4.4. In this subsection we empirically evaluate the error of this approximation.

| Approximate expected cost | Actual expected cost |
|---|---|
| $0.048434950 \pm 0.010130589$ | $0.048434956 \pm 0.010130588$ |

Table 2: Comparison of the approximate expected cost obtained by the closed-loop NIRA with the actual expected cost. The table shows the mean and variance of 100 runs with random location of the obstacle.

Table 2 compares the approximate expected cost function value obtained by the closed-loop NIRA with the actual expected cost estimated by Monte-Carlo simulation with one million samples. The path planning problem is solved 100 times with a randomized location of the obstacle. The risk bound is set to $\Delta = 0.01$. As shown in the table, the approximate cost almost exactly agrees with the actual cost. This is because our closed-loop planning approach explicitly bounds the risk of controller saturation.

### 7.2.5 COMPARISON WITH MDP

Next we compare NIRA with an MDP formulation. For the sake of tractability of the MDP, we consider a single integrator dynamics with a two-dimensional state space and a two-dimensional control input, which specifies the velocity of a vehicle. The rest of the problem setting is the same, except that the state space is discretized into a 100-by-100 grid. We implement a finite-horizon MDP-based path planner, which imposes a penalty $c$ on an event of failure and minimizes the expected cost based on explicit state dynamic programming. The MDP-based path planner imposes a cost as follows:

$$\mathbb{E}\left[\sum_{t=1}^{T}\left(u_{x,t}^2 + u_{y,t}^2 + cI(\boldsymbol{x}_t)\right)\right],$$

where $I(\boldsymbol{x}_t)$ is an indicator function that is one if $\boldsymbol{x}_t$ is in a obstacle and zero otherwise. The resulting optimization problem is solved via dynamic programming.

We ran the MDP-based path planner with three values of penalty $c$: 1, 10, and 100. For each choice of $c$, we conducted 100 simulations with a randomized obstacle position. Figure 14 shows a typical output of the MDP-based path planner. Note that, with a small penalty ($c = 1$), the path planner chooses to take a $100\%$ risk of failure by ignoring the obstacle. This is simply because the penalty of failure is smaller than the expected reduction of cost by going through an obstacle. An issue of utilitarian approaches such as MDPs is that minimization of unconstrained cost can sometimes lead to such impractical solution.

Table 3 shows the mean and the standard deviation of path lengths, as well as the maximum, minimum, and the mean of the resulting probability of failure among the 100 runs. As expected, by imposing a larger penalty, the MDP-path planner chooses a more risk-averse path, which has a longer nominal path length. In this sense, an MDP can also conduct a trade-off between cost and risk. MDP is particularly useful when the primary concern of the user is the cost of failure instead of the probability of failure. On the other hand, when a user would like to impose a hard bound on the probability of failure, our chance constrained planning approach has an advantage. Observe that, even with the same penalty value, the MDP-based path planner results in a wide range of failure probabilities depending on the location of the obstacle. Most notably, with $c = 10$, some of the paths move directly across the obstacle, and in so doing, accept a 100% probability of failure, while others go around the obstacle. Undesirable behaviors, such as crossing an obstacle, are likely to be suppressed by imposing a greater penalty, but without a guarantee. Moreover, imposing a heavy penalty on failure often results in an overly conservative, risk averse solution. On the other hand, the behavior of NIRA with regarding to risk is predictable, in a sense that the path is guaranteed to go around the obstacle, regardless of its location. This is because the chance constraint requires that there exists a margin between the path and the boundary of the obstacle. The p-Sulu Planner inherits this property from NIRA.

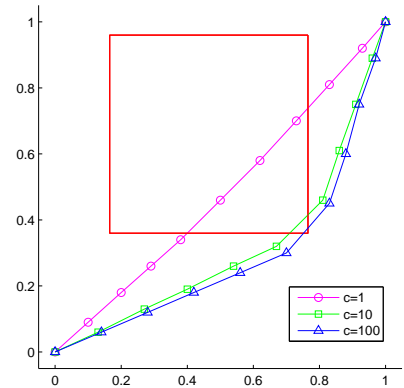

Figure 15: Optimal paths generated by an MDP-based planner with different penalty levels , $c$. The red rectangle represents an obstacle. Note that the path with $c = 1$ cuts through the obstacle.

| Penalty $c$ | path length | Probability of failure | | |
| --- | --- | --- | --- | --- |
| | | Max | Mean | Min |
| 1 | $1.41 \pm 0.00$ | 1.000 | 1.000 | 1.000 |
| 10 | $1.54 \pm 0.05$ | 1.000 | 0.375 | 0.096 |
| 100 | $1.57 \pm 0.06$ | 0.1215 | 0.031 | 0.009 |

Table 3: 100 runs with a randomized obstacle location

### 7.3 The p-Sulu Planner Simulation Results

Next we present the simulation results of the p-Sulu Planner on two problems, in order to illustrate its capability of planning with schedule constraints. We also empirically evaluate the scalability of p-Sulu.

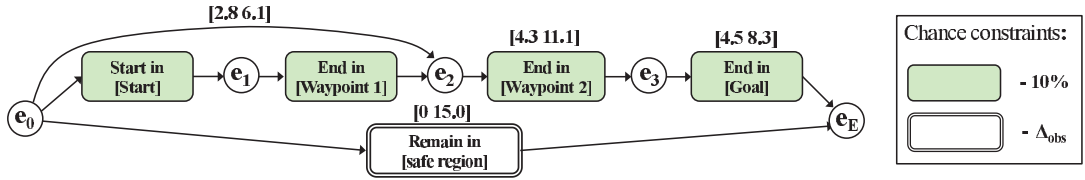

Figure 16: A sample CCQSP for a personal aerial vehicle's path planning and scheduling problem.

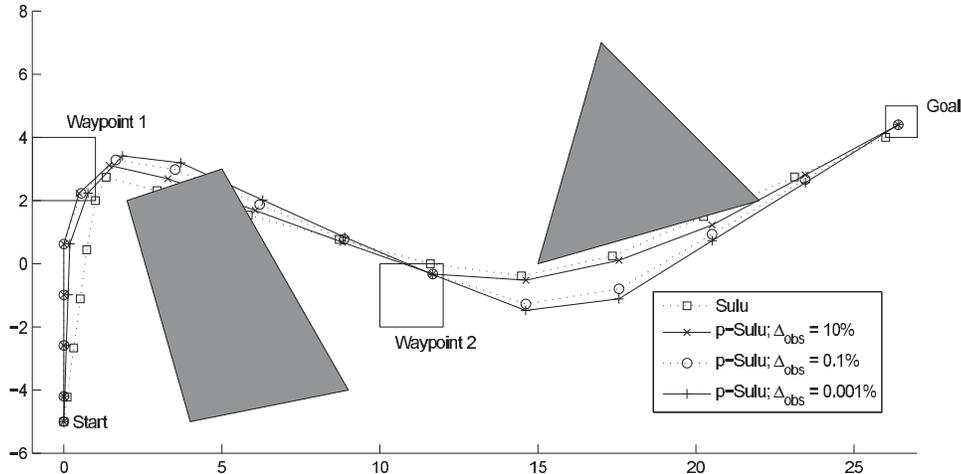

Figure 17: Output of the p-Sulu Planner for the CCQSP in Figure 16 with three different settings of the risk bound $\Delta_{obs}$, compared to the path planned by a deterministic planner, Sulu, which does not consider chance constraints.

### 7.3.1 PATH PLANNING WITH OBSTACLES

In this simulation we test the p-Sulu Planner on a path planning problem in the environment shown in Figure 17. The input CCQSP is shown in Figure 16. The CCQSP requires a vehicle to arrive at the goal region within 15 minutes, by going through Waypoint 1 and Waypoint 2 with the temporal constraints specified in Figure 16. It also imposes two chance constraints: one that requires the vehicle to achieve the time-evolved goals with 90% certainty, and another that requires the vehicle to limit the probability of violating the obstacles to $\Delta_{obs}$. We set $\Delta t = 1$ and $\sigma^2 = 0.0025$.

Figure 17 shows the plans generated by the p-Sulu Planner with three different risk bounds: $\Delta_{obs} = 10\%$, $0.1\%$, and $0.001\%$. The computation times were 79.9 seconds, 86.4 seconds, and 88.1 seconds, respectively. Figure 17 also shows the plan generated by Sulu, a deterministic planner that does not explicitly consider uncertainty (Léauté & Williams, 2005). Observe that Sulu leaves no margin between the path and obstacles. As a result, the Sulu path results in a $94.1\%$ probability of hitting obstacles, as estimated by a Monte-Carlo simulation with $10^7$ samples. On the other hand, the p-Sulu Planner leaves margins between the path and the obstacles in order to satisfy the risk bound, specified in the chance constraint. The margins are larger for the plans with smaller risk bounds. The probabilities of failure of the three plans generated by the p-Sulu Planner, estimated by Monte-Carlo simulations with $10^7$ samples, are $9.53\%$, $0.0964\%$, and $0.00095\%$, respectively. Hence the chance constraints are satisfied. The schedule optimized by the p-Sulu Planner is $\{s(e_0) = 0, s(e_1) = 5, s(e_2) = 10, s(e_E) = 15\}$, which satisfies all the temporal constraints in the CCQSP.

In Figure 16, it appears that the path cuts across the obstacle. This is due to the discretization of time; the optimization problem only requires that the vehicle locations at each discrete time step satisfy the constraints, and does not consider the state in between. This issue can be addressed by a constraint-tightening method (Kuwata, 2003).

### 7.3.2 PATH PLANNING IN AN INDOOR ENVIRONMENT

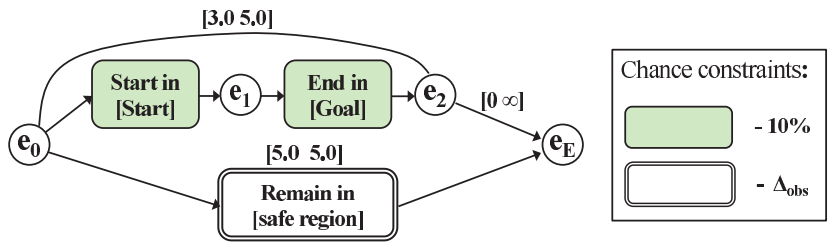

Figure 18: A sample CCQSP for a path planning problem in an indoor environment.

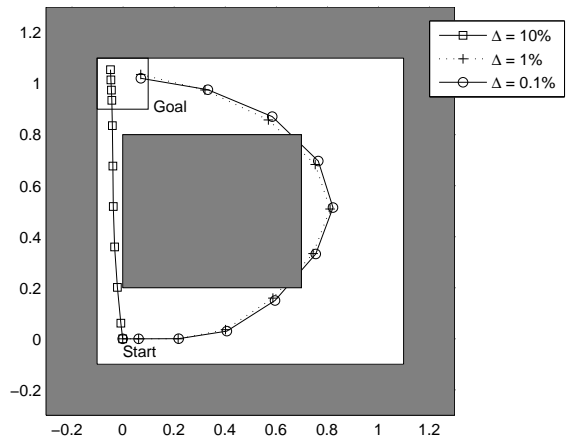

Figure 19: Output of the p-Sulu Planner for the CCQSP in Figure 16 with three different settings of the risk bound $\Delta_{obs}$.

We next give the p-Sulu Planner the CCQSP shown in Figure 18, which simulates a path planning problem in an indoor environment. A vehicle must get to the goal region at the other side of the room in three to five seconds. The "Remain in safe region" episode requires the vehicle to stay within the room and outside of the obstacle during the five-second planning horizon. The CCQSP imposes two chance constraints shown in Figure 18. We set $\Delta t = 0.5$ and $\sigma^2 = 5.0 \times 10^{-5}$.

Given this CCQSP, the planner faces a choice: heading straight to the goal by going through the narrow passage between the left wall and the obstacle minimizes the path length, but involves higher risk of constraint violation; making a detour around the right side of the obstacle involves less risk, but results in a longer path.

Figure 19 shows the p-Sulu Planner's outputs with $\Delta_{obs} = 10\%$, $1\%$, and $0.1\%$. The computation times were 35.1 seconds, 84.5 seconds, and 13.3 seconds, respectively. The result is consistent with our intuition. When the p-Sulu Planner is allowed a 10% risk, the planner chooses to go straight to the goal, resulting in the cost function value of 1.21; when the user gives a 1% or 0.1% risk bound, it chooses the risk-averse path, resulting in the cost function values of 3.64 and 3.84, respectively. This example demonstrates the p-Sulu Planner's capability to make an intelligent choice in order to minimize the cost, while limiting the risks to user-specified levels.

### 7.3.3 SCALABILITY ANALYSIS

In this subsection we conduct an empirical analysis of the scalability of the p-Sulu Planner, as the environment becomes increasingly constrained.. As shown in Figure 20, we measured the computation time to solve a path planning problem with different numbers of obstacles and waypoints. In all simulations, the path starts at $[0, 12]$ and ends in a square region centered at $[24, 12]$. Figure 20 shows twenty simulation results, with zero to three obstacles and zero to four waypoints. Obstacles and waypoints are represented by blue and red squares in the figure, respectively. The positions of the center of the obstacles are $[6, 12]$, $[12, 12]$, and $[18, 12]$, while the positions of the center of the waypoints are $[9, 9]$, $[9, 15]$, $[15, 15]$, and $[15, 9]$. The computation time is shown in the caption of each subfigure in Figure 20.

By comparing the results in Figure 20 horizontally, we observe exponential growth in computation time with the number of obstacles. This result is expected since the number of disjunctive clauses in the state constraint of the p-Sulu Planner increases exponentially with the number of obstacles. Building a tractable extension of the p-Sulu Planner for a large number of obstacles is future work. On the other hand, by comparing the results vertically, we find that the computation time with the same number of obstacles and different number of waypoints stays in the same order of magnitude. This is because adding an extra waypoint only increases the number of conjunctive clauses in the state constraints.

In the remaining sections we describe the application of psulu to two real world problems, air vehicle and space vehicle control. A third application, building energy management, using a variant of the p-Sulu Planner, is reported by Ono, Graybill, and Williams (2012).

## 7.4 PTS Scenarios

Next, we deploy the p-Sulu Planner on PTS scenarios, the robotic air taxi system introduced in Section 1.

### 7.4.1 SCENARIOS

We consider three scenarios, specified by the CCQSPs shown in Figure 21. Scenarios 1 and 2 are similar to the scenic flight scenario introduced at the beginning of this paper (see Figure 1). In

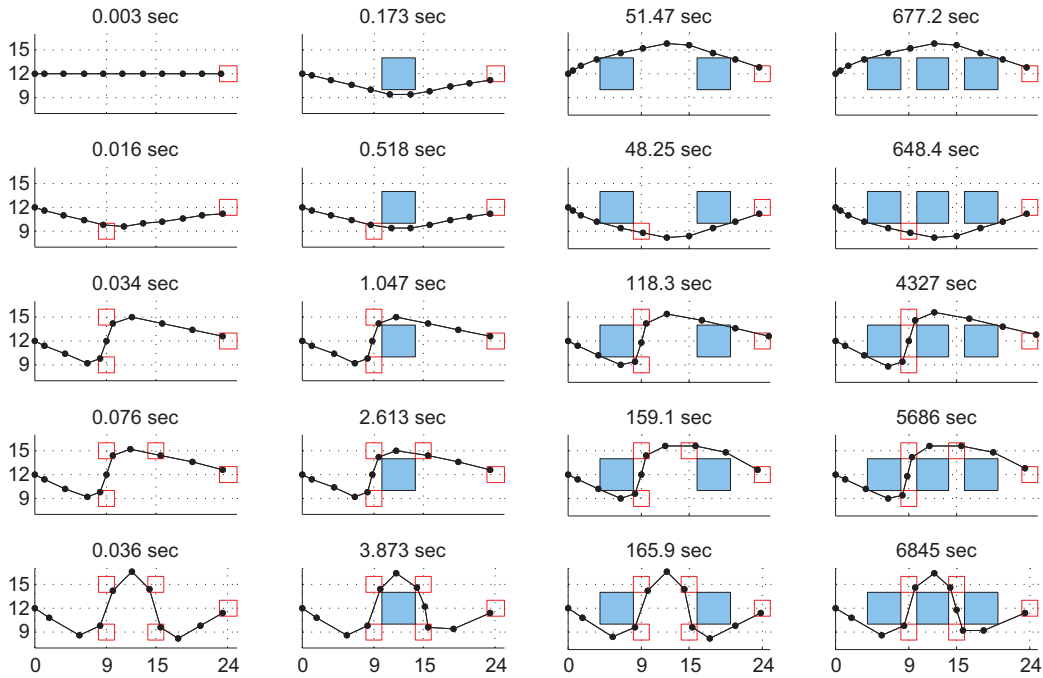

Figure 20: Computation time of the p-Sulu Planner for a path planning problem with different numbers of obstacles and waypoints.

Scenario 1, a personal aerial vehicle (PAV) takes off from Runway 7[1] of Provincetown Municipal Airport (KPVC) in Provincetown, Massachusetts, fly over a scenic region, and lands on Runway 23 of Hanscom Field (KBED) in Bedford, Massachusetts. The vehicle is required to stay within the scenic region at least for 2 minutes and at most for 10 minutes. The entire flight must take more than 13 minutes and less than 15 minutes. Scenario 2 is the same as Scenario 1, except for the runways used for take-off and landing.

Scenario 3 simulates a leisure flight off the coast of Massachusetts. A PAV takes off Runway 7 of Provincetown Municipal Airport, and flies over two regions where whales are often seen. Then the vehicle lands on Runway 11 of Hanscom Field.

We place three no-fly zones, as shown in Figure 22. The entire flight must take more than 13 minutes and less than 15 minutes. Each scenario has three chance constraints, $\{c_1, c_2, c_3\}$, as shown in Figure 21. The first one, $c_1$, is concerned with the vehicle's operation; it requires the vehicle to take off from and land on the right runways at the right airports with less than 10 % probability of failure. The second chance constraint, $c_2$, is concerned with the leisure activities; it requires the vehicle to fly over the scenic regions with less than 10 % probability of failure. Finally, $c_3$ is concerned with the passenger's safety; it requires the vehicle to limit the risk of penetrating the no-fly zones to 0.01 %.

---

1. A runway of an airport is specified by a number, which represents the clockwise angle from the north. For example, Runway 7 points 70 degrees away from the north.

### 7.4.2 PLANT PARAMETERS

We set $\boldsymbol{u}_{\max} = 250$ m/s, which approximates the maximum cruise speed of private jet airplanes, such as Gulfstream V. The maximum acceleration is determined from the maximum bank angle. Assuming that an aircraft is flying at a constant speed, the lateral acceleration $a$ is given as a function of the bank angle $\phi$ as follows:

$$a = g \cdot \tan \phi,$$

where $g$ is the acceleration of gravity. Typically passenger aircraft limits the bank angle to 25 degrees for passenger comfort, even though the aircraft is capable of turning with a larger bank angle. Hence, we use:

$$u_{\max} = 9.8 \text{ m/s}^2 \cdot \tan(25°) = 4.6 \text{ m/s}^2.$$

We set $\sigma = 100$ m and $\Delta T = 60$ seconds.

### 7.4.3 SIMULATION RESULTS

Figure 22 shows the paths planned by the p-Sulu Planner for the three scenarios. In all the scenarios, all the episode requirements in the CCQSPs in Figure 21 are met within the specified temporal and chance constraints.

Table 4 compares the performance of Sulu and the p-Sulu Planner. As expected, Sulu's plans result in excessive probabilities of failure in all scenarios. This is because Sulu does not consider uncertainty in the planning process, although the PAV is subject to disturbance in reality. On the other hand, the p-Sulu Planner successfully limits the probability of failure within the user-specified risk bounds for all three scenarios. Furthermore, although the p-Sulu Planner significantly reduces the risk of failure, its cost is higher than that of Sulu only by 9.5 - 12.8 %. Such a capability of limiting the risk and maximizing the efficiency at the same time is a desirable feature for PTS, which transports passengers.

| Scenario number | 1 | | 2 | | 3 | |
|---|---|---|---|---|---|---|
| Planner | Sulu | **p-Sulu** | Sulu | **p-Sulu** | Sulu | **p-Sulu** |
| Computation time [sec] | 2.58 | 60.2 | 2.00 | 390 | 5.17 | 198 |
| $P_{fail,1}$ | 0.999 | $9.12 \times 10^{-2}$ | 0.996 | $9.14 \times 10^{-2}$ | 0.999 | $9.23 \times 10^{-2}$ |
| $P_{fail,2}$ | 0.807 | $8.46 \times 10^{-2}$ | 0.813 | $8.59 \times 10^{-2}$ | 0.603 | $7.65 \times 10^{-2}$ |
| $P_{fail,3}$ | 0.373 | $2.74 \times 10^{-5}$ | 0.227 | $2.62 \times 10^{-5}$ | 0.372 | $2.81 \times 10^{-5}$ |
| Cost function value $J^{\star}$ | 24.2 | 27.5 | 21.0 | 23.7 | 20.0 | 22.3 |

Table 4: Performance Comparison of the prior art, Sulu, and the p-Sulu Planner. $P_{fail,1}$, $P_{fail,2}$, and $P_{fail,3}$ represent the probabilities of failure regarding the chance constraints $c_1$, $c_2$, and $c_3$ in Figure 21, respectively.

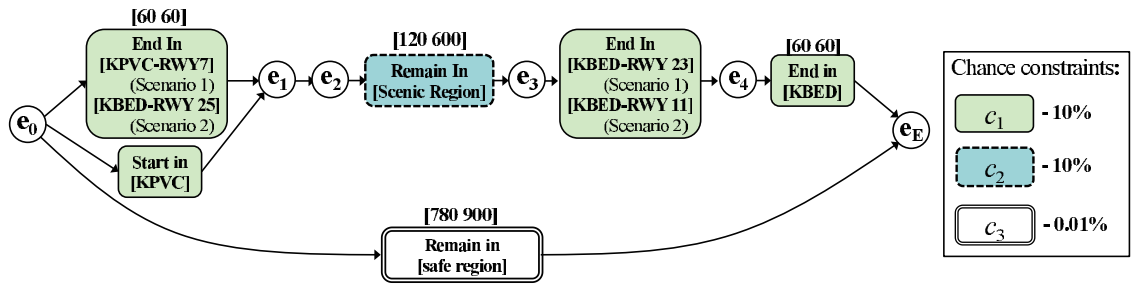

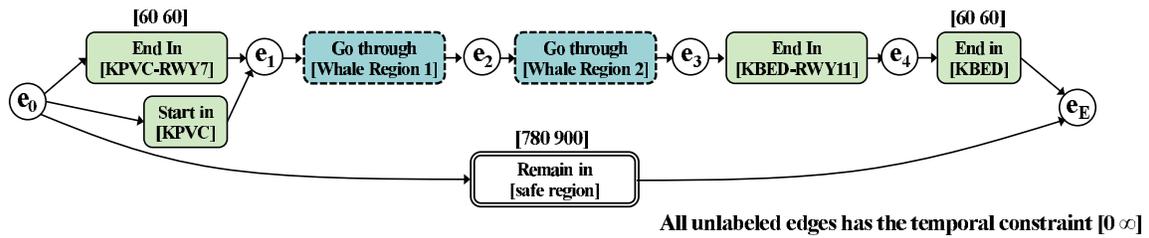

Figure 21: The CCQSPs for the PTS scenarios.

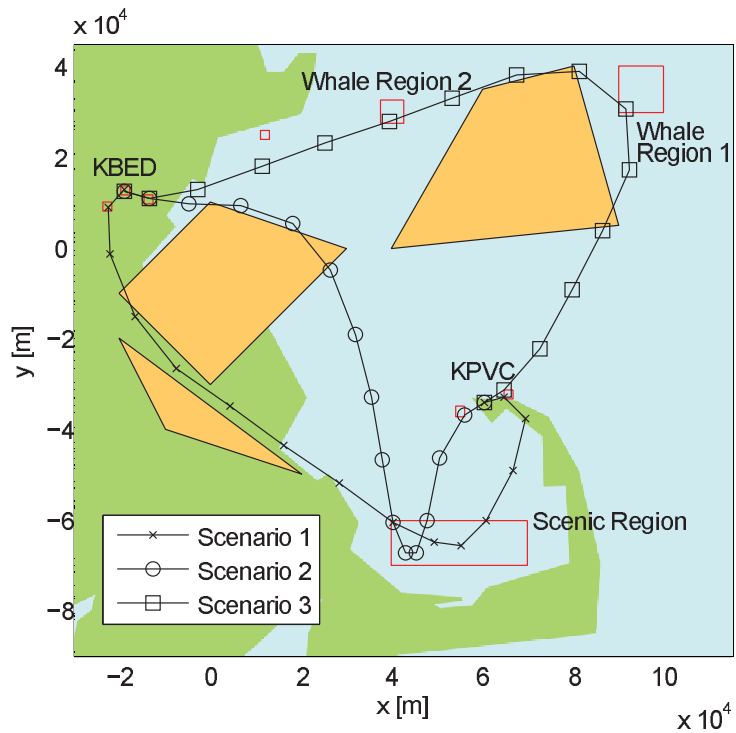

Figure 22: The paths planned by the p-Sulu Planner.

As shown in Table 4, the p-Sulu Planner typically takes several minutes to compute the plan. This length of computation time would be allowed for PTS applications, since we assume that the p-Sulu Planner is used for preplanning; before take-off, the passengers of a PAV specify requirements, and the p-Sulu Planner creates a risk-sensitive flight plan. We assume that a real-time plan executive executes the plan after take-off.

We note that it is more desirable to have a real-time *risk-sensitive* plan executive, since risk factors, such as the location of storms, change over time. Our future work is to reduce the computation time of the p-Sulu Planner so that it can be used for real-time execution.

## 7.5 Space Rendezvous Scenario

The p-Sulu Planner is a general planner whose application is not limited to a specific plant model. In order to show the generality of the planner, we deployed the p-Sulu Planner on a system whose plant model is significantly different from PTS.

Specifically, we chose an autonomous space rendezvous scenario of the H-II Transfer Vehicle (HTV), shown in Figure 23, as our subject. HTV is an unmanned cargo spacecraft developed by the Japanese Aerospace Exploration Agency (JAXA), which is used to resupply the International Space Station (ISS). Collision of the vehicle with the ISS may result in a fatal disaster, even if the collision speed is low. For example, in August 1994, the Russian unmanned resupply vehicle Progress M-34 collided with the Mir space station in a failed attempt to automatic rendezvous and docking. As a result, one of the modules of Mir was permanently depressurized. In order to avoid such an accident, HTV is required to follow a specified safety sequence during the automated rendezvous, as described in the following subsection.

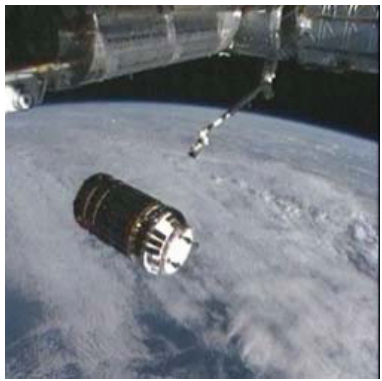

Figure 23: H-II Transfer Vehicle (HTV), a Japanese unmanned cargo vehicle, conducts autonomous rendezvous with the International Space Station. Image courtesy of NASA.

### 7.5.1 HTV RENDEZVOUS SEQUENCE

In HTV's autonomous rendezvous mission, the final approach phase starts from the Approach Initiation (AI) point, which is located 5 km behind the ISS, as shown in Figure 24. First, HTV moves to the R-bar Initiation (RI) point, which is located 500 m below the ISS, guided by the relative GPS navigation. At the RI point, HTV switches the navigation mode to Rendezvous Sensor (RVS) Navigation. In RVS Navigation, HTV measures the distance to ISS precisely by beaming the laser to the reflector placed on the nadir (earth-facing) side of the ISS. Then, HTV proceeds to the Hold Point (HP), located 300 m below the ISS. It is required to hold at HP in order to perform a 180-degree yaw-around maneuver. The new orientation of HTV allows the vehicle to abort the rendezvous quickly in case of emergency. After the yaw-around maneuver, HTV resumes the approach, and holds again at the Parking Point (PP), which is 30 m below the ISS. Finally, HTV approaches at a distance of 10 meters from the ISS, and stops within the Capture Box (CB) of the ISS's robotic arm. The robotic arm then grabs HTV and docks it to the ISS. Please refer to the report by Japan Aerospace Exploration Agency (2009) for the details of the rendezvous sequence.

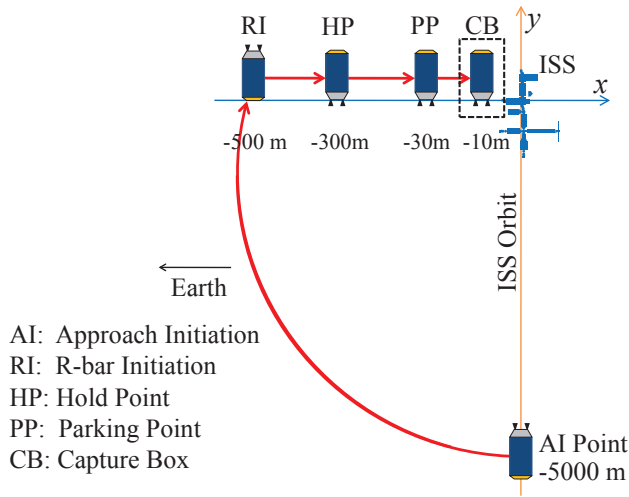

Figure 24: HTV's final approach sequence (Japan Aerospace Exploration Agency, 2009).

The rendezvous sequence described above is represented by the CCQSP shown in Figure 25. In addition to the time-evolved goals specified in the actual rendezvous sequence, we specify temporal constraints and chance constraints in the simulation, as shown in the figure. We require HTV to hold at each intermediate goal for at least 240 seconds. The transition between the goals must take at least 600 seconds, in order to make sure that the vehicle moves slowly enough. The entire mission must be completed within 4800 seconds (1 hour 20 minutes). We require HTV to stay within the Safe Zone, a conic area below the ISS, during the RVS navigation phase with 99.5% probability, since otherwise the laser may not be reflected back to HTV properly. We assume that the goals are square regions, with 10 m sides for RI and HP, 2 m sides for PP, and 1 m sides for CB. Finally, we require that HTV achieves all the time-evolved goals with 99.5% success probability.

### 7.5.2 ORBITAL DYNAMICS

The rendezvous can be considered as a two-body problem, where a chaser spacecraft (e.g., HTV) moves in relation to a target spacecraft (e.g., ISS), which is in a circular orbit. In such a problem, it is convenient to describe the motion of the chaser spacecraft using a rotating frame that is fixed to the target space craft, known as a Hill coordinate frame (Schaub & Junkins, 2003). As shown in Figure 24, we set the $x$-axis pointing away from the center of the earth and the $y$-axis along the orbital velocity of the target spacecraft. Since HTV's path is within the $x$-$y$ plane, we don't consider the $z$-axis.

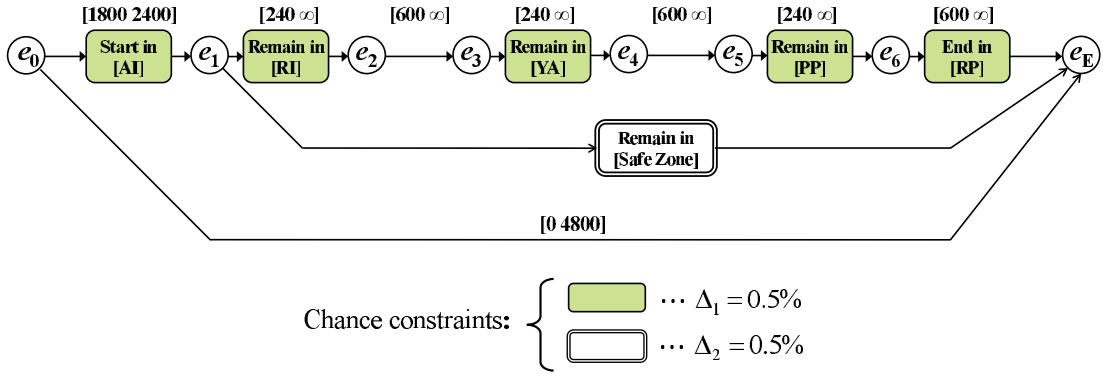

Figure 25:  A CCQSP representation of the HTV's final approach sequence. We assume the same time-evolved goals as the ones used for actual flight missions. The temporal constraints and the chance constraints are added by the authors.

It is known that the relative motion of the chase spacecraft in the Hill coordinate frame is described by the following Clohessy-Wiltshire (CW) equation (Vallado, 2001):

$$\ddot{x} = 2\omega\dot{y} + 3\omega^2 x + F_x$$
$$\ddot{y} = 2\omega\dot{x} + F_y$$

where $\omega$ is the angular speed of the target spacecraft's orbit, and $F_x$ and $F_y$ are the force per unit mass, or the acceleration in $x$ and $y$ directions. The first terms on the right-hand sides represent the Coriolis force.

An object that follows the CW equation moves in an unintuitive manner. Its unforced motion is not in a straight line due to the Coriolis effect; in general, an object cannot stay at the same position without external force. For example, Figure 26 shows the fuel-optimal path to visit two waypoints, A and B, and come back to the start. As can be seen in the figure, the optimal path is not typically a straight line. The virtue of the p-Sulu Planner is that it can handle such irregular dynamic systems in the same way as regular systems, just by setting the $A$ and $B$ matrices of our plant model (4) appropriately.

The state vector consists of positions and velocity in the $x - y$ plane:

$$\boldsymbol{x} = [x\ y\ v_x\ v_y]^T$$

We obtain the discrete-time CW equation using the impulse-invariant discretization:

$$\boldsymbol{x}_{k+1} = A\boldsymbol{x}_k + B\boldsymbol{u}_k,$$

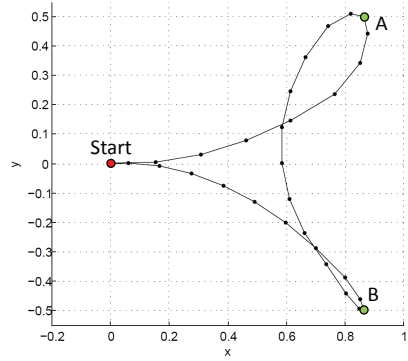

Figure 26: A typical motion of spacecraft in the Hill coordinate frame. The solid line is the fuel optimal path to visits A and B and returns to the Start in 30 minutes. Note that the optimal path is not a straight line in the Hill coordinate frame.

where

$$
A = \begin{bmatrix}
4 - 3\cos(\omega\Delta T) & 0 & \frac{\sin(\omega\Delta T)}{\omega} & \frac{2\{1-\cos(\omega\Delta T)\}}{\omega} \\
-6\{\omega\Delta T - \sin(\omega\Delta T)\} & 1 & \frac{-2\{1-\cos(\omega\Delta T)\}}{\omega} & \frac{4\sin(\omega\Delta T)}{\omega} - 3\Delta T \\
3\omega\sin(\omega\Delta T) & 0 & \cos(\omega\Delta T) & 2\sin(\omega\Delta T) \\
-6\omega\{1 - \cos(\omega\Delta T)\} & 0 & -2\sin(\omega\Delta T) & 4\cos(\omega\Delta T) - 3
\end{bmatrix}
$$

$$
B = \begin{bmatrix}
\frac{\sin(\omega\Delta T)}{\omega} & \frac{2\{1-\cos(\omega\Delta T)\}}{\omega} \\
\frac{-2\{1-\cos(\omega\Delta T)\}}{\omega} & \frac{4\sin(\omega\Delta T)}{\omega} - 3\Delta T \\
\cos(\omega\Delta T) & 2\sin(\omega\Delta T) \\
-2\sin(\omega\Delta T) & 4\cos(\omega\Delta T) - 3
\end{bmatrix}
$$

We use the ISS's orbital angular speed, $\omega = 0.001164$ rad/sec, at which the station goes around the Earth in 90 minutes. We choose the interval $\Delta T = 120$ seconds. The number of time steps $N$ is set to 40. Hence, the entire plan is 4800 seconds (1 hour and 20 minutes). In the discretization, we assumed impulse inputs as follows:

$$
\begin{bmatrix} F_x \\ F_y \end{bmatrix} = \sum_{k=0}^{N-1} \delta(t - \Delta T \cdot k)\boldsymbol{u}_k,
$$

where $\delta(\cdot)$ is the Dirac delta function. Such an assumption is justified because the thrusters of the Reaction Control System (RCS) of the spacecraft, which are used for the final approach maneuver, operate for a very short duration ($0.01 - 5.0$ seconds) at each burn (Wertz & Wiley J. Larson, 1999).

We consider stochastic uncertainty $w$, added to the discrete-time dynamic equation:

$$
\boldsymbol{x}_{k+1} = A\boldsymbol{x}_k + B\boldsymbol{u}_k + w.
$$

Such an assumption of additive uncertainty is commonly used in past research on autonomous rendezvous and formation flight in space (Shields, Sirlin, & Wette, 2002; Smith & Hadaegh, 2007;

Campbell & Udrea, 2002). We assume that $w$ has a zero-mean Gaussian distribution, with the following covariance matrix:

$$\mathbf{\Sigma}_w = \begin{pmatrix} 10^{-6} & 0 & 0 & 0 \\ 0 & 10^{-6} & 0 & 0 \\ 0 & 0 & 0 & 0 \\ 0 & 0 & 0 & 0 \end{pmatrix}.$$

### 7.5.3 OBJECTIVE FUNCTION

We employ an objective function $J$ that requires for the p-Sulu Planner to minimize the fuel consumption. It follows from the Tsiolkovsky rocket equation that the fuel consumption of spacecraft is proportional to the total change in velocity, called Delta-V or $\Delta V$ (Wertz & Wiley J. Larson, 1999). The total fuel consumption is the summation of the fuel consumption of reaction jets in $x$ and $y$ directions for all time steps. Hence our objective function is described as follows:

$$
\begin{aligned}
J(\boldsymbol{u}_{0:N}) &= \Delta V_x + \Delta V_y \\
&= \int_0^{(N-1)\Delta T} |F_x| + |F_y| dt \\
&= \sum_{k=0}^{k=N-1} \left| \int_0^{(N-1)\Delta T} \delta(t - \Delta T \cdot k) \boldsymbol{u}_{x,k} dt \right| + \left| \int_0^{(N-1)\Delta T} \delta(t - \Delta T \cdot k) \boldsymbol{u}_{y,k} dt \right| \\
&= \sum_{k=0}^{k=N-1} |\boldsymbol{u}_{x,k}| + |\boldsymbol{u}_{y,k}|.
\end{aligned}
$$

### 7.5.4 SIMULATION RESULT

Figure 27 shows the planning result of the p-Sulu Planner. We compare the result with Sulu, as well as a nominal planning approach, in which we assume that HTV moves from AI to RI using a two-impulse transition (called "CW guidance law") (Matsumoto, Dubowsky, Jacobsen, & Ohkami, 2003; Vallado, 2001). From RI to CB, it follows a predetermined path that goes through the center of the Safe Zone, as shown in Figure 27-(b), with a constant speed.

As shown in Figure 27, the optimal paths generated by the p-Sulu Planner and Sulu are not straight. Such curved paths exploit the Coriolis effect to minimize fuel consumption.

Table 5 compares the performance of the three planning approaches. The two rows regarding the probabilities of failure correspond to the two chance constraints specified in the CCQSP, shown in Figure 25. The probabilities are evaluated by Monte Carlo simulation with one million samples.

As expected, the probabilities of failure of the path generated by the p-Sulu Planner are less than the risk bounds specified by the CCQSP, shown in Figure 25. On the other hand, once again, Sulu's path results in almost 100% probability of failure. This is because Sulu minimizes the fuel consumption without considering uncertainty. The resulting path pushes against the boundaries of feasible regions, as is evident in Figure 27-(c). Also note that, although the p-Sulu Planner significantly reduces the probability of constraint violation compared with Sulu, its cost (Delta V) is higher than Sulu only by 0.2%. The p-Sulu Planner results in a significantly smaller cost (Delta V) than the nominal planning approach. The 1.42 m/sec reduction in Delta V is equivalent to an 11.9 kg saving of fuel, assuming the $16,500$ kg mass of the vehicle and the 200 sec specific impulse

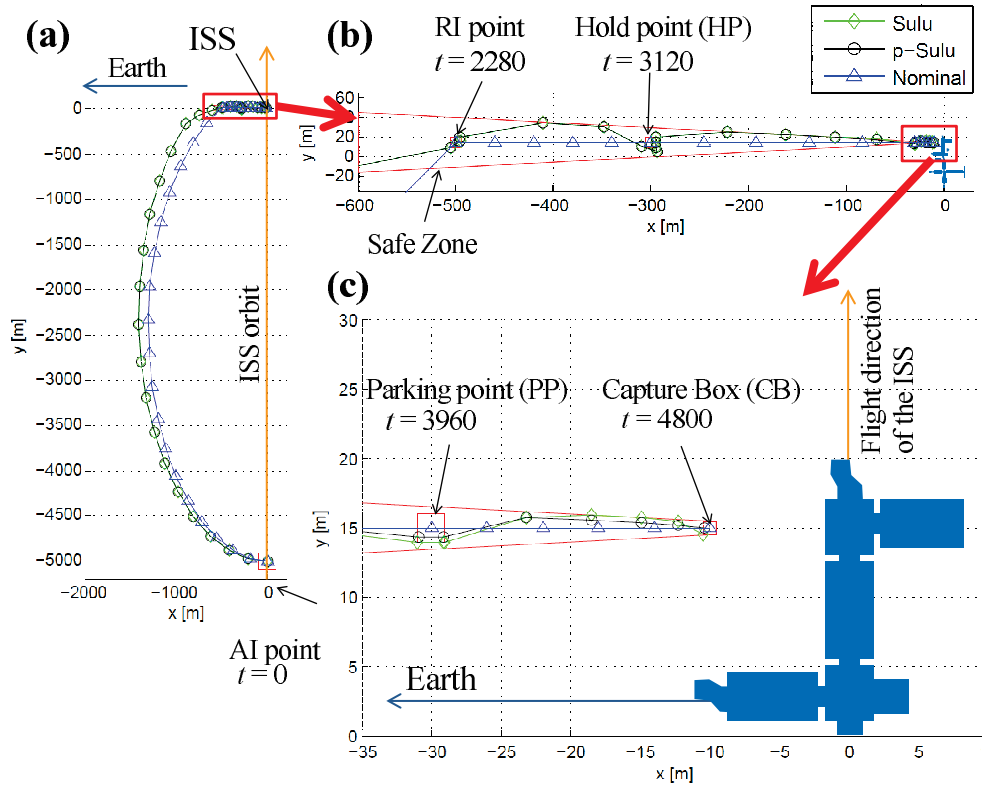

AI: Approach Initiation, RI: R-bar Initiation, YA: Yaw-around

Figure 27: Planning results of Sulu, the p-Sulu Planner, and a nominal planning approach. The input CCQSP is shown in Figure 25.

($I_{SP}$) of the thrusters. Although the p-Sulu Planner takes longer to compute the plan than the other two approaches, the 11.4 second computation time is negligible compared with the 1 hour and 20 minute plan duration.

|  | Sulu | **the p-Sulu Planner** | Nominal |
|---|---|---|---|
| Computation time [sec] | 3.9 | 11.4 | 0.09 |
| Probability of failure $P_{fail}$ (Navigation) | 0.92 | 0.0024 | $< 10^{-6}$ |
| Probability of failure $P_{fail}$ (Goals) | 1.0 | 0.0029 | $< 10^{-6}$ |
| Cost function value (Delta V) $J^\star$ [m/sec] | 7.30 | 7.32 | 8.73 |

Table 5: Performance comparison of Sulu, the p-Sulu Planner, and the nominal approach on the HTV rendezvous scenario.

## 8. Conclusions

This article introduced a model-based planner, the p-Sulu Planner, which operates within user-specified risk bounds. The p-Sulu Planner optimizes a continuous control sequence and a discrete schedule, given as input a continuous stochastic plant model, an objective function, and a newly developed plan representation, a chance-constrained qualitative state plan (CCQSP). A CCQSP involves time-evolved goals, simple temporal constraints, and chance constraints, which specify the user's acceptable levels of risk on subsets of the plan.

Our approach to developing the p-Sulu Planner was two-fold. In the first step, we developed an efficient algorithm, called non-convex iterative risk allocation (NIRA), that can plan in a non-convex state space but for a *fixed* schedule. We solved the problem based on the key concept of risk allocation and risk selection, which achieves tractability by allocating the specified risk to individual constraints and by mapping the result into an equivalent disjunctive convex program. The NIRA algorithm employs a branch-and-bound algorithm to solve the disjunctive convex program. Its subproblems are fixed-schedule CCQSP problems with a *convex* state space, which can be solved by our previously developed algorithms (Blackmore & Ono, 2009). We developed a novel relaxation method called fixed risk relaxation (FRR), which provides the tightest linear relaxation of the nonlinear constraints in the convex subproblems.

In the second step, we developed the p-Sulu Planner, which can solve a CCQSP planning problem with a *flexible* schedule. The scheduling problem was formulated as a combinatorial constrained optimization problem (COP), which is again solved by a branch-and-bound algorithm. Each subproblem of the branch-and-bound search is a CCQSP planning problem with a *fixed* schedule, which is solved by NIRA. The domain of the feasible schedule is pruned by running a shortest-path algorithm on the d-graph representation of the given temporal constraints. The lower bounds of the optimal objective value of the subproblems are obtained by solving fixed-schedule CCQSP planning problems where a subset of the state constraints are imposed. We proposed an efficient variable ordering that prioritizes convex subproblems over non-convex ones. We demonstrated the p-Sulu Planner on various examples, from a personal aerial transportation system to autonomous space rendezvous, and showed that it can efficiently solve CCQSP planning problems with small suboptimality, compared to past algorithms.

### Acknowledgments

This paper is based upon work supported in part by the Boeing Company under Grant No. MIT-BA-GTA-1 and by the National Science Foundation under Grant No. IIS-1017992. Any opinions, findings, and conclusions or recommendations expressed in this publication are those of the authors and do not necessarily reflect the view of the sponsoring agencies. We would like to thank Michael Kerstetter, Scott Smith, Ronald Provine, and Hui Li at Boeing Company for their support. Thanks also to Robert Irwin for advice on the draft.